\PassOptionsToPackage{%
  pagebackref=true,%
  breaklinks=true,%
  letterpaper=true,%
  colorlinks=true,%
  citecolor=blue,%
  linkcolor=blue,%
  bookmarks=false,%
  allcolors=cvprblue%
}{hyperref}
\documentclass[10pt,twocolumn,letterpaper]{article}

\usepackage[pagenumbers]{cvpr} % To force page numbers, e.g. for an arXiv version
\usepackage{graphicx}
\usepackage{amsmath}
\usepackage{amssymb}
\usepackage{booktabs}

\usepackage{float}

\usepackage{stfloats} % added by author2

\usepackage{dsfont}
\usepackage{subcaption}
\usepackage[linesnumbered,ruled,vlined]{algorithm2e}
\usepackage{caption}
\usepackage{nicefrac}
\usepackage{mathrsfs}
\usepackage{xcolor}
\usepackage{enumitem}
\usepackage{colortbl}
\usepackage{multirow}
\usepackage{marvosym}
\usepackage[export]{adjustbox}
\usepackage{amsfonts}
\usepackage{pifont}

\definecolor{sh_gray}{rgb}{0.84,0.84,0.84}
\definecolor{sh_gray2}{rgb}{1,0.89,0.75}
\definecolor{color3}{rgb}{0.95,0.95,0.95}
\definecolor{color4}{rgb}{0.94,0.94,1}
\definecolor{color5}{rgb}{1,0.96,0.88}
\definecolor{cvprblue}{rgb}{0.21,0.49,0.74}
\usepackage[pagebackref,breaklinks,colorlinks,allcolors=cvprblue]{hyperref}

\DeclareMathOperator*{\argmin}{arg\,min}
\title{Distribution-aware Dataset Distillation for Efficient Image Restoration}

\author{%
  Zhuroan Zheng\footnotemark[2]\\
  Sun Yat‑sen University\\
  \and
  Xin Su\footnotemark[2]\\
  Fuzhou University\\
  \and
  Chen Wu\\
  University of Science and Technology of China\\
  \and 
  Xiuyi Jia\footnotemark[1]\\
  Nanjing University of Science and Technology
}

\begin{document}
\maketitle

\footnotetext[2]{Equal contribution.}
\footnotetext[1]{Corresponding author: \texttt{jiaxy@njust.edu.cn}.}

\begin{abstract}
With the exponential increase in image data, training an image restoration model is laborious. 
Dataset distillation is a potential solution to this problem, yet current distillation techniques are a blank canvas in the field of image restoration.
To fill this gap, we propose the Distribution-aware Dataset Distillation method (TripleD), a new framework that extends the principles of dataset distillation to image restoration.
Specifically, TripleD uses a pre-trained vision Transformer to extract features from images for complexity evaluation, and the subset (the number of samples is much smaller than the original training set) is selected based on complexity.
The selected subset is then fed through a lightweight CNN that fine-tunes the image distribution to align with the distribution of the original dataset at the feature level.
%
%To efficiently condense knowledge, the training is divided into two stages. Early stages focus on simpler, low-complexity samples to build foundational knowledge, while later stages select more complex and uncertain samples as the model matures.
%
Our method achieves promising performance on multiple image restoration tasks, including multi-task image restoration, all-in-one image restoration, and ultra-high-definition image restoration tasks.
Note that we can train a state-of-the-art image restoration model on an ultra-high-definition (4K resolution) dataset using only one consumer-grade GPU in less than 8 hours (\textbf{500}\% savings in computing resources and immeasurable training time).
%
%Samples with higher scores, indicating they offer more learning value, are prioritized during training. 
%
%This dynamic selection ensures that the model engages with different parts of the dataset across training stages, maintaining diverse exposure while reducing redundant computation.
%
%To efficiently condense knowledge, the training is divided into two stages. Early stages focus on simpler, low-complexity samples to build foundational knowledge, while later stages select more complex and uncertain samples as the model matures. 
%
%By dynamically selecting samples, the dataset constructed with limited resources (only 2\% of the original dataset sample number) achieved a performance close to 90\% of the full original dataset. 
%

%
%It is worth noting that we only used a single RTX 3090 GPU to condense the ultra-high-definition image dataset in a limited amount of time (it only takes 20h to condense 8K pairs of images).

\end{abstract}    
\section{Introduction}
\begin{figure}[!t]
	\centering
	\includegraphics[width=0.5\textwidth]{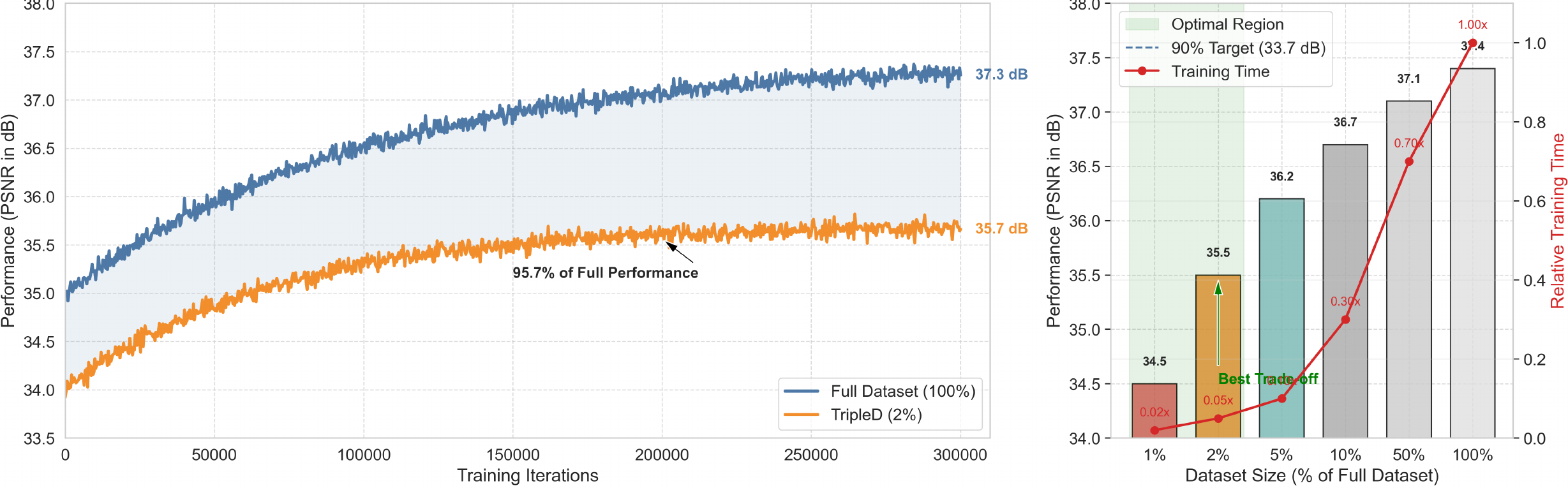}
	\caption{\textbf{Performance and analysis of our TripleD method for image restoration.} First, the figure on the left shows that using only \textbf{2}\% of the original dataset allows the model to perform close to training on the full set. The figure on the right demonstrates an ablation where the performance of the image restoration model rises steadily as the proportion of synthetic datasets rises, but the improvement is very limited.}
 \vspace{-4mm}
	\label{fig1}
\end{figure}

\par Image restoration provides high-quality information for downstream visual tasks such as denoising \cite{chen2017trainable}, deblurring \cite{kupyn2018deblurgan}, and deraining \cite{zhang2019image}. Recent advancements have employed deep learning techniques to boost restoration quality, achieving state-of-the-art results across various benchmarks \cite{yin2020image, zhang2021image}. 
Although deep learning algorithms have shown promising results, they are highly dependent on the quantity of the datasets.
Indeed, it has become a trend to create large-scale benchmarks, which unfortunately also brings expensive training costs.

To alleviate this problem, dataset distillation \cite{wang2021dataset} becomes a potential option.
This technique aims to synthesize a representative subset (the amount of data is much smaller than the original dataset) as a training set for the model to achieve similar performance when trained on a large-scale dataset.
However, this technology is still a blank in the field of image restoration. Since the task of image restoration is a refined regression operation, there is no margin error like the classification task (for classification tasks, a probability of 0.5 to 1 indicates that the label exists).

\par In this paper, we design a distribution-aware dataset distillation (TripleD) algorithm, which can efficiently synthesise a small dataset from a large-scale image restoration dataset.
As illustrated in Figure \ref{fig1},  TripleD mehrod, which has two properties: i) A high-quality synthetic dataset can be distilled for efficient training of the image recovery model using only 2\% of the data volume of the training set (PSNR reaches the upper limit of around 95\%). ii) Although the model performs better as the number of synthetic datasets increases, the improvement is limited.
TripleD achieves up to 90$\sim$95\% of the performance of full-dataset training while utilizing only 1$\sim$5\% of the data. Additionally, TripleD significantly reduces the training resource requirements, enabling models such as Restormer to be trained efficiently on a single RTX 3090 GPU. 
Specifically, TripleD uses features extracted from downsampled images (the input images are all downsampled to a resolution of $128 \times 128$) using a Vision Transformer (ViT) \cite{dosovitskiy2021image} to represent the complexity of images.
Here, this pre-trained Transformer is trained on the ImageNet dataset, and the label is the entropy value of the image. it is worth noting that relying only on the entropy value of the image to represent the image complexity is subject to error, and we manually conduct a significant amount of revision annotation.
We use the complexity of images for key purposes:
With the help of the GPU shader, Vision Transformer can be run faster on downsampled input images without losing the semantics of the original resolution image.
Relying only on traditional calculations to obtain the entropy value of an image is not only time-consuming and laborious but also subject to errors.
Indeed, we also considered measures such as cosine distance, standard deviation, KL divergence, and entropy, but they performed poorly in evaluating the complexity of images.
%
%After obtaining the complexity of the image, we implement a progressive curriculum strategy (as shown in Figure \ref{fig1}b) where the selection ratio dynamically increases from 30\% to 70\% across training stages. This allows us to select the most suitable sample clusters for distillation, starting with simpler samples and gradually incorporating more complex ones as training progresses.
%
For the generation of synthetic samples, we used a diffusion model, which produces better and more varied quality than GAN~\cite{dietz2025study}.
Finally, for distribution matching between the synthetic dataset and the selected dataset, we used a joint regularization i.e., KL scatter and L2.
Our contributions are as follows:
\begin{itemize}
	\item To the best of our knowledge, we are the first to design a dataset distillation framework (TripleD) for image restoration tasks, which alleviates the cost of training image restoration models on large-scale data sets.
	\item We develop an efficient dynamic distillation method that progressively increases the complexity of the sample bar to select high-quality sample clusters for distillation to obtain sub-datasets (2\% of full dataset).
	\item Our method can also achieve dataset distillation on a single GPU in a limited amount of time when executing ultra-high-definition datasets. Extensive experimental results show that our method can help the model maintain 90\% of its performance at a lower cost.
\end{itemize}

\section{Related Work}
\noindent \textbf{Image Restoration.}
Image restoration aims to recover clean images from degraded inputs, addressing tasks such as denoising~\cite{zhang2017beyond}, deblurring~\cite{nah2017deep}, and deraining~\cite{zhang2018density}.
Recently, since Transformer has a strong ability to capture global features, models based on it have shown very promising results, such as Restormer \cite{zamir2022restormer}, SwinIR~\cite{liang2021swinir} and IPT \cite{chen2021pre}, have achieved state-of-the-art results. 
Unfortunately, these methods are too computationally expensive for deployment on mobile or embedded devices.
So far, RAMiT \cite{nguyen2023ramit}, and FreqFormer \cite{freqformer2024}, propose more efficient attention mechanisms to reduce complexity while maintaining performance. 
%
%Furthermore, methods like Histoformer \cite{sun2024restoring} tackle adverse weather degradations by introducing novel approaches such as histogram self-attention. 
%
%
Although the inference efficiency of models has been significantly improved, the exponential increase in data sets has not been addressed in terms of training efficiency.
Dataset distillation provides a potential solution to this problem.

\noindent \textbf{Subset Selection.}
Subset selection is a sampling technique for picking representative samples from a dataset, maintaining model performance with sampled data.
Active learning~\cite{settles2009active, sener2018active}, continuous learning~\cite{parisi2019continual, loshchilov2016sgdr}, and incremental learning~\cite{rebuffi2017icarl} all reduce training costs by selecting the samples with the greatest information content, such as the sample variance or entropy.
For example, Settles \cite{settles2010active} demonstrated how active learning can effectively prioritize data points, minimizing labeling efforts while maintaining high model performance. 
The Herding method in \cite{belouadah2020scail} iteratively minimizes the mean embedding discrepancy between the selected subset and the whole dataset in the feature space.
Indeed, it is particularly important for image restoration tasks, which are usually high-cost due to its dense prediction.
Recent advances in dynamic data selection for low-level vision tasks further highlight the need for adaptive sampling strategies that evolve during training \cite{nguyen2023ramit, yang2023dynamic, wang2022adaptive, liu2021low}. In this paper, we extend these principles by leveraging Vision-Language Models (VLMs) to compute selection scores, dynamically selecting samples based on complexity. This enables more efficient and targeted training, particularly in large-scale image restoration tasks.

\noindent \textbf{Dataset Distillation.}
While dataset distillation has been successful in reducing training load for classification tasks \cite{wang2021dataset, zhao2023datasetdistillation, cazenavette2022dataset}, its application to image restoration remains challenging. 
The challenge is the continuous nature of outputs in image restoration, which requires precise pixel-level predictions, making it difficult to select representative samples. 
For classification tasks, there is some tolerance for errors. For example, a label between 0 and 0.5 may be classified as 0.
In addition, image restoration involves various types of degradation, such as noise, blur, and rain, each of which introduces different complexities~\cite{zhang2017beyond,zhang2018density}.
Traditional sample selection and data synthesis methods \cite{zhao2021dataset} struggle to capture the full range of these degradations, leading to suboptimal training efficiency. 
Recent dynamic sample selection approaches \cite{nguyen2022dataset, yang2023dynamic} have begun to address these limitations, but they are still underdeveloped for tasks like image restoration, which require nuanced data selection and adaptation.
In this paper, a Dynamic Data Distillation algorithm is developed to solve the problems of accurate pixel prediction and the variability of degradation types.

\begin{figure*}[!t]
    \centering
    \includegraphics[width=0.95\linewidth]{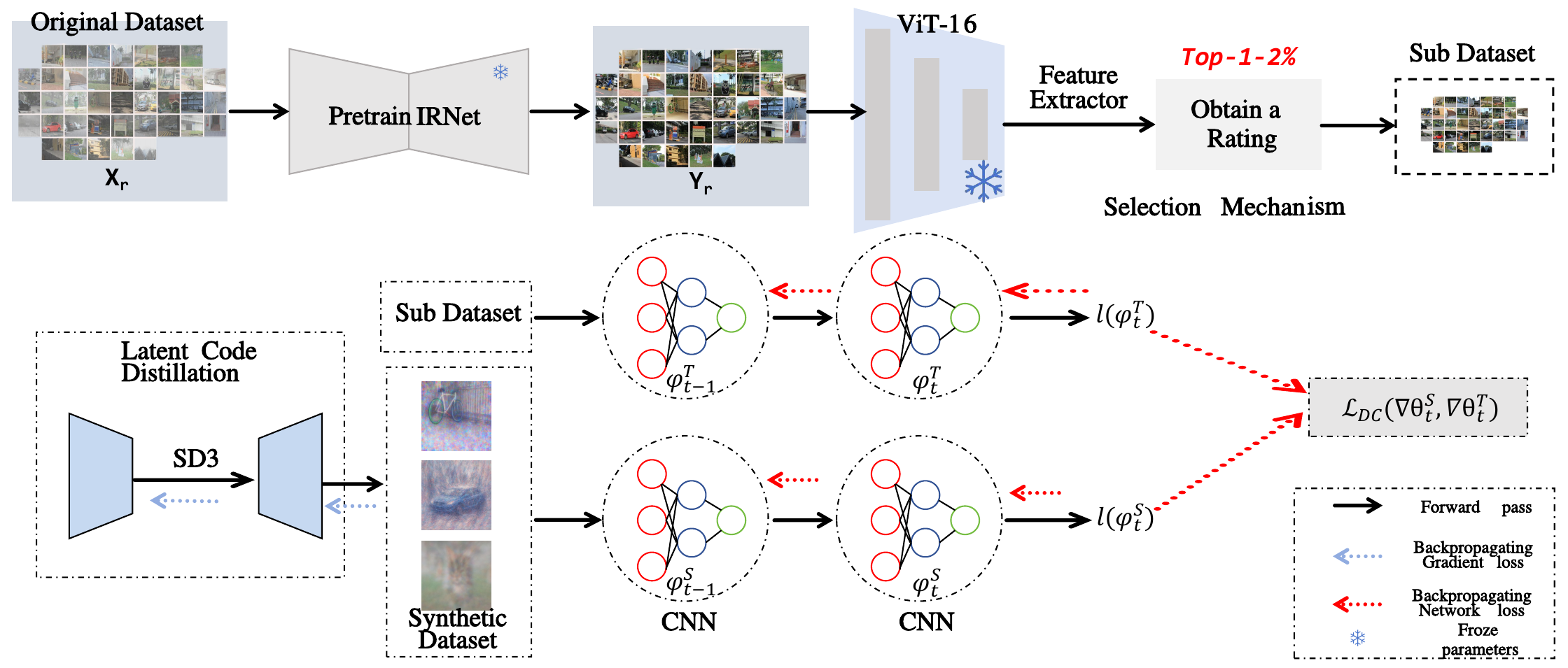}
    \caption{\textbf{Overview of the TripleD  Pipeline.} A pretrained IRNet backbone (SwinIR) first processes large-scale real–degraded pairs ($X_{\text{r}} \rightarrow Y_{\text{r}}$), and a ViT-based entropy scorer selects the top 1-2\% most informative samples as the real sub-dataset. Meanwhile, a diffusion-based latent distillation module (SD3) synthesizes compact training pairs whose latents match those of the selected real samples. Finally, both the distilled real and synthetic subsets are used to train the restoration network under task and distillation losses, greatly reducing training data while preserving performance.}
    \label{fig: framework}
\end{figure*}

\section{Method}
In this section, we introduce the Distribution-aware Dataset Distillation (TripleD) method, which includes preliminary knowledge, image complexity calculation,  and fine-tuning of data distribution  (see Figure~\ref{fig: framework}). 
\subsection{Preliminary knowledge}
\noindent \textbf{Image Restoration.}
A trained image restoration (IR) model $\zeta_\theta: \mathbb{R}^{\text{H} \times \text{W} \times \text{C}} \rightarrow \mathbb{R}^{\text{H} \times \text{W} \times \text{C}}$ is optimized to approximate the inverse of a degradation process that maps a clear image (HQ) $\mathbf{y} \in \mathbb{R}^{\text{H} \times \text{W} \times \text{C}}$ to its corresponding Low-Quality (LQ) counterpart $\mathbf{x} \in \mathbb{R}^{\text{H} \times \text{W} \times \text{C}}$.
The model parameters $\theta$ are optimized using a dataset $\mathcal{T} = (X_r, Y_r)$, a collection of LQ-HQ image pairs. 
The objective function for training the IR model is
\begin{equation}
    \theta^{*} = \argmin_\theta \mathbb{E}_{\mathbf{x}_i \in X_r, \mathbf{y}_i \in Y_r}\lVert \psi_\theta (\mathbf{x}_i) - \mathbf{y}_i\rVert^2.
\end{equation}

\noindent \textbf{Dataset Distillation.}
The goal of dataset distillation is to compress a raw dataset $\mathcal{T} = (X_r, Y_r)$, where $X_r \in \mathbb{R}^{\text{N} \times \text{H} \times \text{W} \times \text{C}}$ and N is the number of samples, into a significantly smaller synthetic dataset $\mathcal{S}=(X_s, Y_s)$ with $X_s \in \mathbb{R}^{\text{M} \times \text{H} \times \text{W} \times \text{C}}$ and $\text{M} \ll \text{N}$. 
%However, the goal is not only to reduce the dataset size but also to retain the essential training quality of $\mathcal{T}$ \cite{cazenavette2023generalizing}.
In this setting, M is defined as $\text{M}=\mathcal{C} \cdot \text{IPC}$, $\mathcal{C}$, where $\mathcal{C}$ is the number of classes and IPC represents the Images Per Class (IPC). 
The overall optimization for dataset distillation is formulated as
\begin{equation}\small
    \text{\small $\mathcal{S}^* = \mathop{\arg\min}\limits_{\mathcal{S}}\mathcal{L}(\mathcal{S}, \mathcal{T})$}, \label{eq:def}
\end{equation}
where $\mathcal{L}$ is a predefined objective for dataset distillation.
It computes the loss on real data ($\ell^\mathcal{T}$) and the corresponding synthetic data ($\ell^\mathcal{S}$), then minimizes the discrepancy between the gradients of both networks:
\begin{equation}
    \mathcal{L}_{DC} =1- \frac{\nabla_\theta\ell^\mathcal{S}(\theta) \cdot \nabla_\theta\ell^\mathcal{T}(\theta)}{\left\|\nabla_\theta\ell^\mathcal{S}(\theta)\right\|\left\|\nabla_\theta\ell^\mathcal{T}(\theta)\right\|}.
\end{equation}

\subsection{Selection Subset}
Given a dataset $\mathbf{X}$, it extracts features from a pre-trained encoder (ViT-16, 16 denotes the number of layers of the model) to evaluate the complexity of the image distribution.
The output of ViT comes with a Sigmoid function to squeeze the values to 0-1, and the corresponding entropy labels of the modified ImageNet dataset are also squeezed to 0-1 with the help of Max-Min normalization.
Specifically, the input image is downsampled to a resolution of $128 \times 128$ by using bilinear downsampling.
Here, the image is downsampled to improve the efficiency of the data distillation algorithm.
Finally, we select a subset of the images in this image restoration dataset that are ranked in the top 1\% or 2\% of entropy values.

\begin{figure}[h]
\begin{center}
\scalebox{0.850}{
\begin{tabular}[b]{c@{ } c@{ } c@{ }}
    \includegraphics[width=3.1cm,height=3.3cm, valign=t]{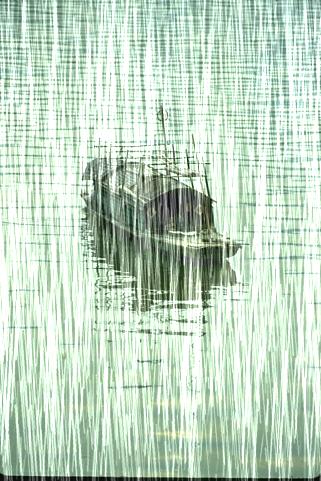}&
  	\includegraphics[width=3.1cm,height=3.3cm, valign=t]{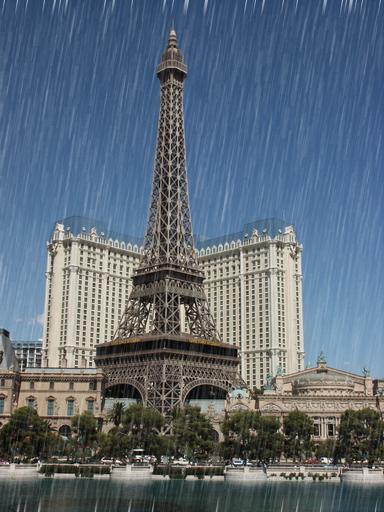}&   
    \includegraphics[width=3.1cm,height=3.3cm, valign=t]{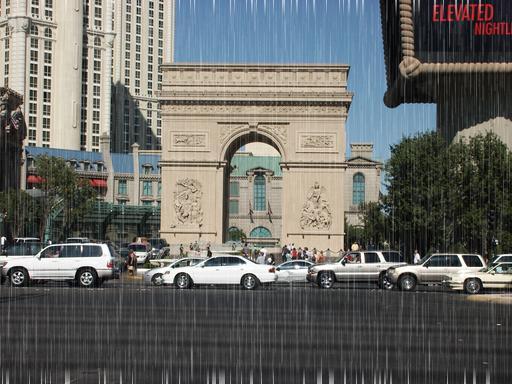}\\
    \small~(a) Low complexity &\small~(b) Medium complexity & \small~(c) High complexity \\
    0.18 & 0.38  & 0.57
    \\
    %\multicolumn{3}{c}{\includegraphics[width=9.3cm,valign=t]{figs/combined_boxplot.jpg}}\\
    %\multicolumn{3}{c}{\small~(d) Combined Box Plot} \\
\end{tabular}}
\end{center}
\vspace{-4mm}
    \caption{This figure shows the complexity of the image is evaluated by using the VIT that is pre-trained.}
    \label{fig:combined_plot}
\vspace{-4mm}
\end{figure}
To better illustrate the effectiveness of this method, we visualize some degraded images to explain.
%As illustrated in Figure~\ref{fig:combined_plot}, the box plot in (d) highlights the variance differences across the three images. 
Figure~\ref{fig:combined_plot}(a), featuring uniform rain streaks, exhibits the lowest variance, indicating the simplest complexity. 
In contrast, Figure~\ref{fig:combined_plot}(b) shows moderate complexity, with a broader range of variance, representing a mix of simple and more varied features. 
Lastly, Figure~\ref{fig:combined_plot}(c), which includes diverse objects and textures, demonstrates the highest variance, reflecting the greatest complexity. 
It is important to note that we observe a phenomenon where the strength of the noise (rain and fog) affects the complexity of the image, and therefore, the subset selection is mainly considered on clean images.
Finally, we obtain a subset dataset $\mathbf{B}$.
\subsection{Latent Distillation for IR}
In addition to pixel-level distillation, we also consider the application of Generative Latent Distillation (GLaD)~\cite{cazenavette2023generalizing} in the distillation of latent datasets. Instead of optimizing synthetic images directly in pixel space, GLaD operates in the latent space of the generative model (the GAN in the model, which we replace with a diffusion model (SD3)). This approach enables distillation at a higher level of abstraction, capturing complex image features while reducing the need for direct pixel-by-pixel optimization. Specifically, GLaD learns latent codes that reconstruct high-fidelity synthetic images as they pass through the generator. The method can distill large-scale image structures while preserving fine-grained details.

\subsection{Data Fine-tuning via CNN}
We built an 8-layer convolutional CNN network with $3 \times 3$ kernels and a step size of 1, which is used to fine-tune the distribution of the dataset $\mathbf{B}$ (generate a latent feature space, where the features of the subset are $\mathbf{F}_a$ and the features of the synthetic data $\mathbf{B}_s$ are $\mathbf{F}_b$). 
The fine-tuned dataset $\mathbf{B}_s$ is used to train the image restoration model, and the CNN is updated with the help of the loss of the image restoration model.
A simple L2 loss and KL are employed to provide pixel-level supervision:
\begin{equation}
    \mathcal{L}_{pix} = \| \mathbf{I}_{\mathbf{F}_a} - \mathbf{I}_{\mathbf{F}_b} \|_2 +  \mathcal{L}_{DC}(\mathbf{F}_a, \mathbf{F}_b),
\end{equation}
where $\| \cdot \|_2$ denotes $\text{L}_2$ norm, $\mathcal{L}_{DC}$ denotes KL dispersion. 

\section{Experiments}
In this section, we conduct extensive experiments to demonstrate the effectiveness of TripleD for image restoration tasks.
Our experiment mainly contains three groups: one is multi-task image restoration, one is all-in-one image restoration, and one is ultra-high-definition (UHD) low-light image restoration.
In addition, we evaluate the effectiveness of the various components of our method on the task of multi-task image restoration.
\subsection{Multi-task Image Restoration}
We conduct experiments across several standard low-level vision tasks, including single-image deraining, motion deblurring, defocus deblurring, and image denoising. 
All comparison methods were retrained on a small-scale dataset distilled from the Restormer \cite{zamir2022restormer}. 
%All experiments follow the architecture of Restormer \cite{zamir2022restormer}, a state-of-the-art image restoration net.
%
\vspace{-5mm}
\paragraph{Datasets.} We employ the Rain100L~\cite{yang2017deep}, Rain100H~\cite{yang2017deep}, GoPro~\cite{nah2017deep}, RealBlur-J~\cite{rim2020real}, RealBlur-R~\cite{rim2020real}, DPDD~\cite{abuolaim2020defocus}, SIDD~\cite{abdelhamed2018high}, and DND~\cite{plotz2017benchmarking} datasets for benchmarking. These datasets provide a wide variety of image degradations, allowing us to evaluate the generalizability of TripleD across different tasks.
\vspace{-5mm}
\paragraph{Training Setting.} All models are trained using the AdamW optimizer, with an initial learning rate of $3 \times 10^{-4}$, and a cosine annealing learning rate scheduler is applied. Training is conducted on patches of size $384\times384$, with a batch size of $16$. In cases where the datasets are relatively small, we employ the TripleD strategy, dynamically selecting 2\% of the data in each epoch. We conduct an experiments on a single NVIDIA RTX 3090 GPU for both TripleD and baseline single-GPU training, following the original setup of the Restormer. To comprehensively assess the effectiveness of our strategy, we use two widely adopted metrics: Peak Signal-to-Noise Ratio (PSNR)~\cite{huynh2008scope} and Structural Similarity Index Measure (SSIM)~\cite{wang2004image}. 

\subsection{Baseline Comparisons}
\begin{figure*}[!t]
\begin{center}
\scalebox{0.9}{
    \begin{tabular}{@{}ccccccc@{}}
        % 第一行图像
        \includegraphics[width=2.5cm]{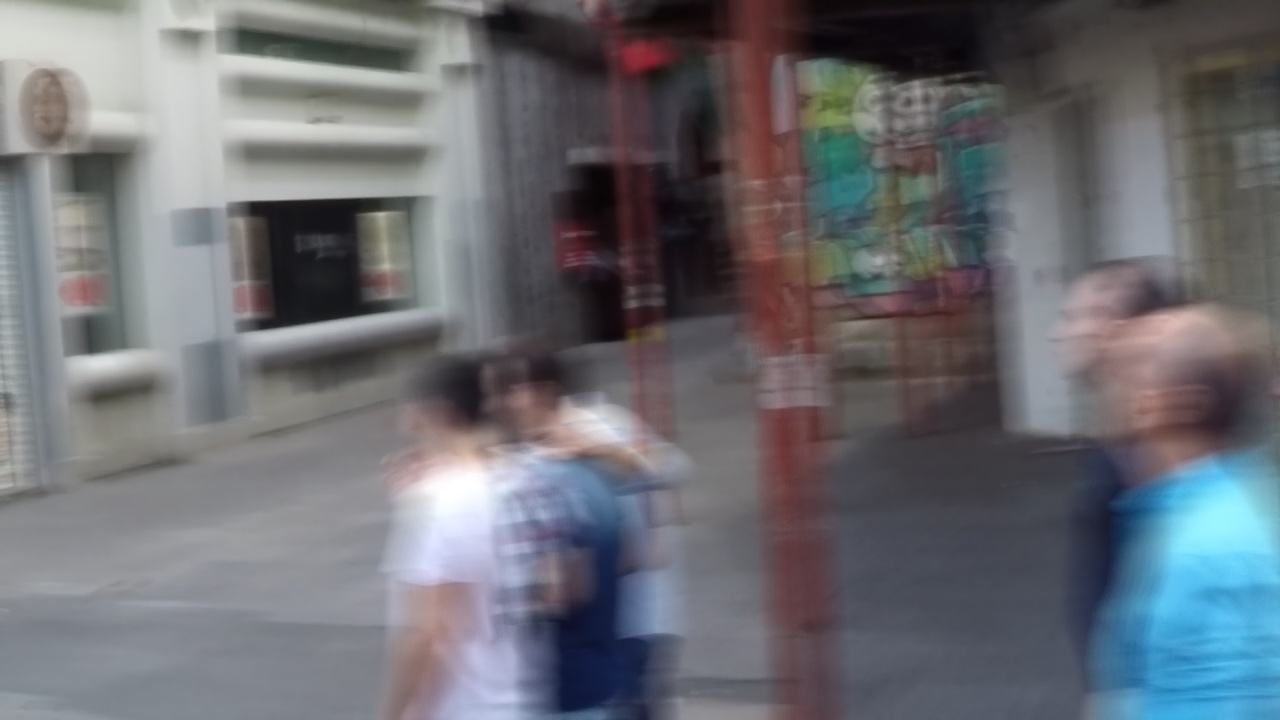} \hspace{-4mm} &
        \includegraphics[width=2.5cm]{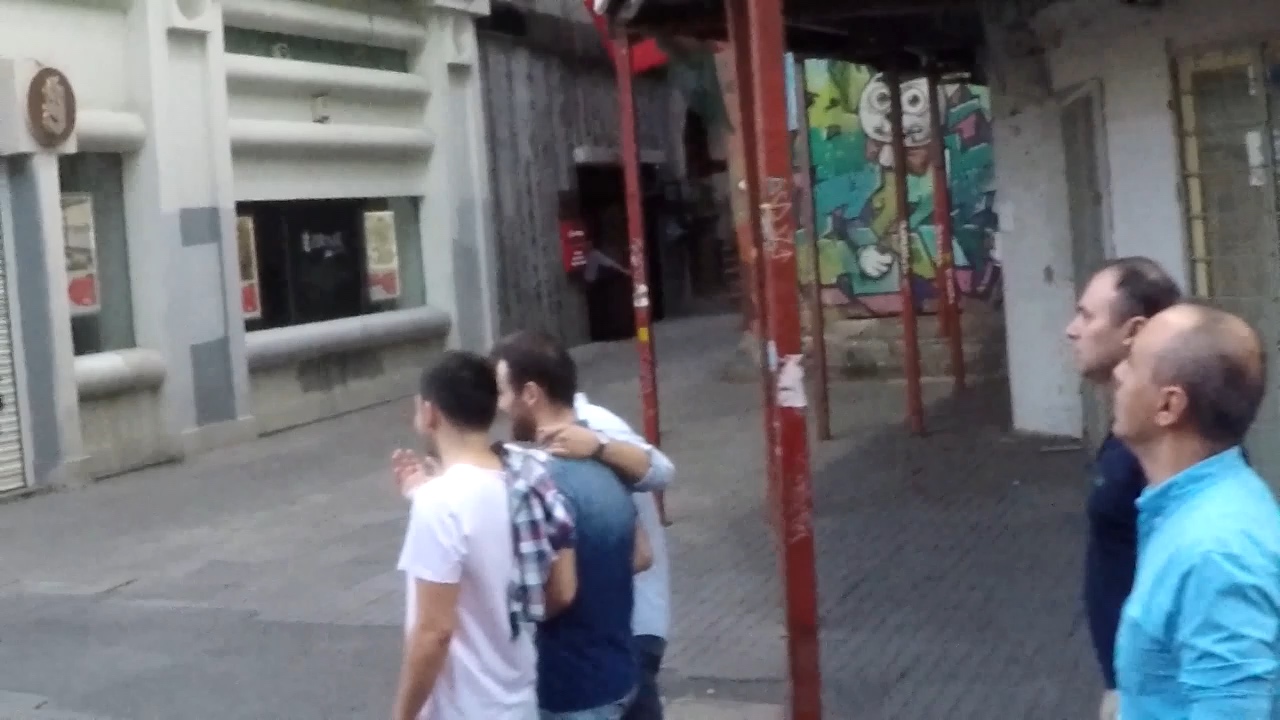} \hspace{-4mm} &
        \includegraphics[width=2.5cm]{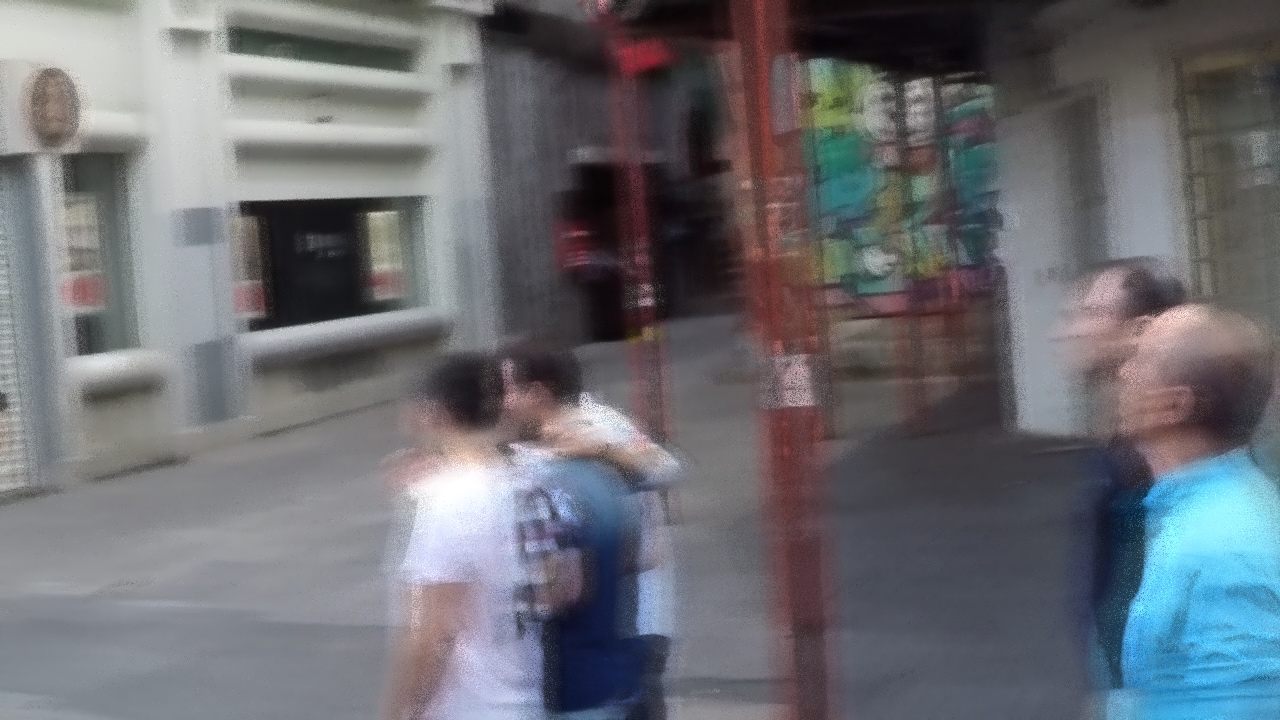} \hspace{-4mm} &
        \includegraphics[width=2.5cm]{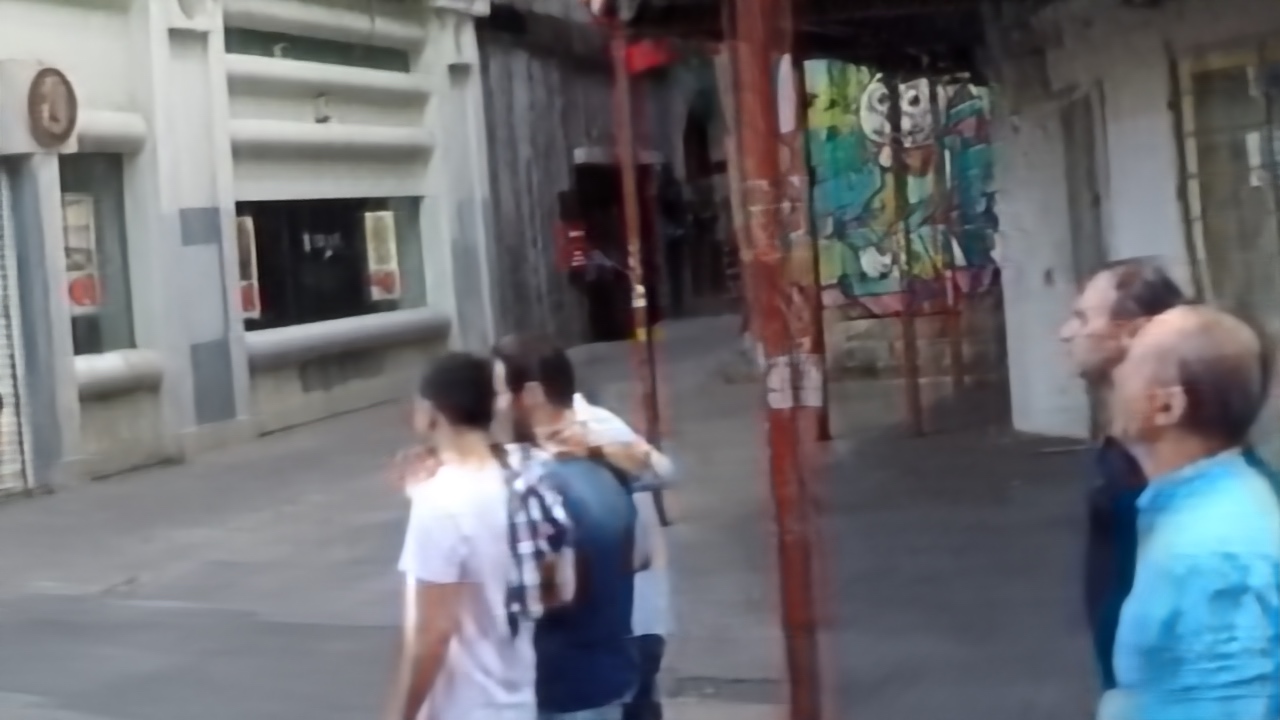} \hspace{-4mm} &
        \includegraphics[width=2.5cm]{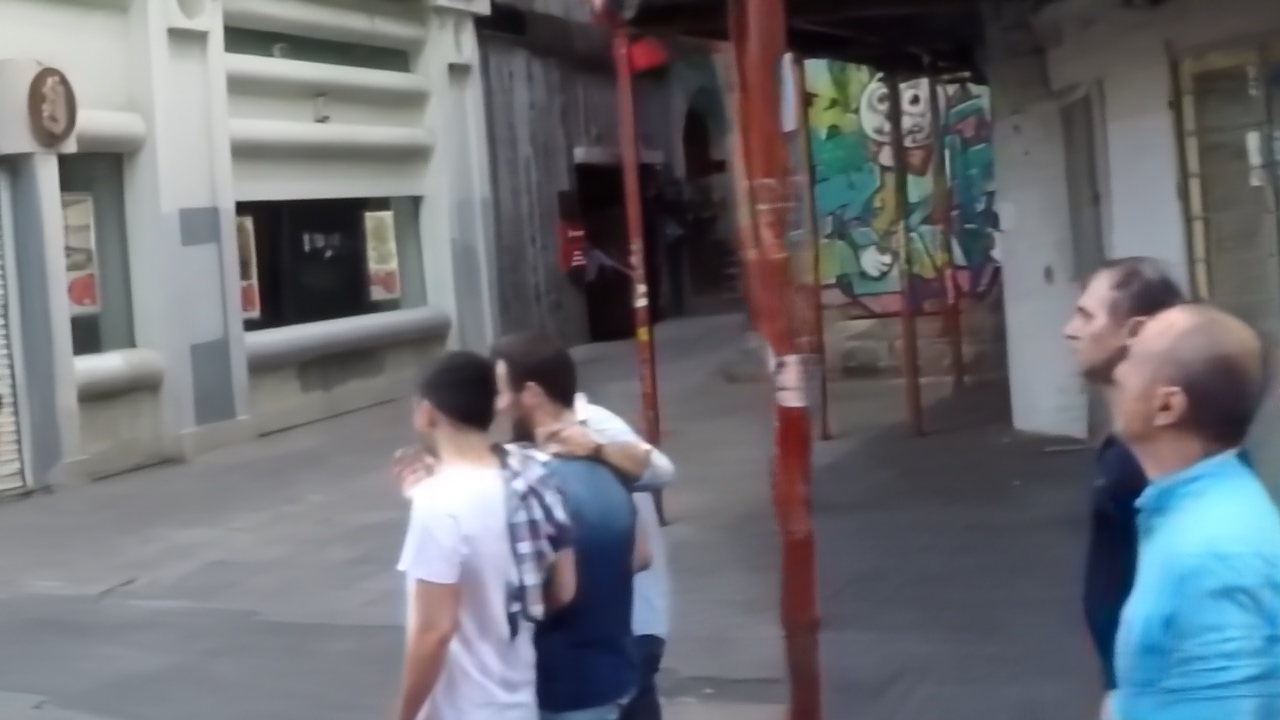} \hspace{-4mm} &
        \includegraphics[width=2.5cm]{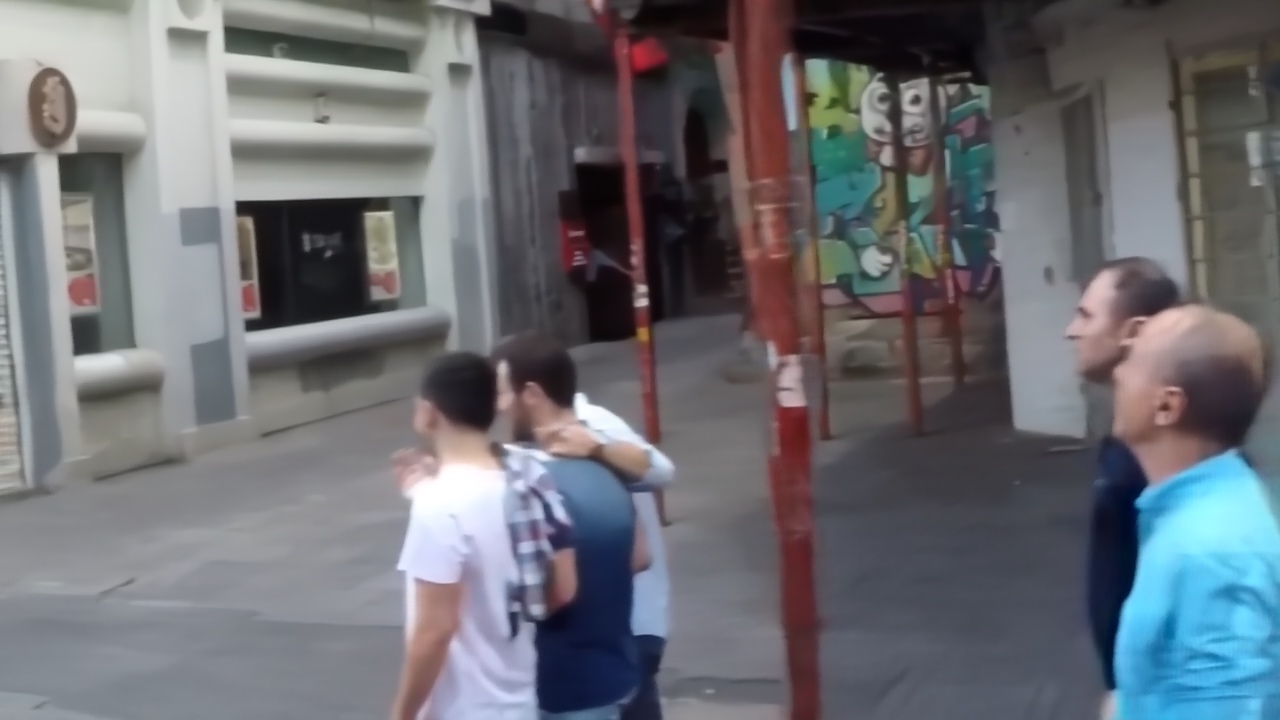} \hspace{-4mm} &
        \includegraphics[width=2.5cm]{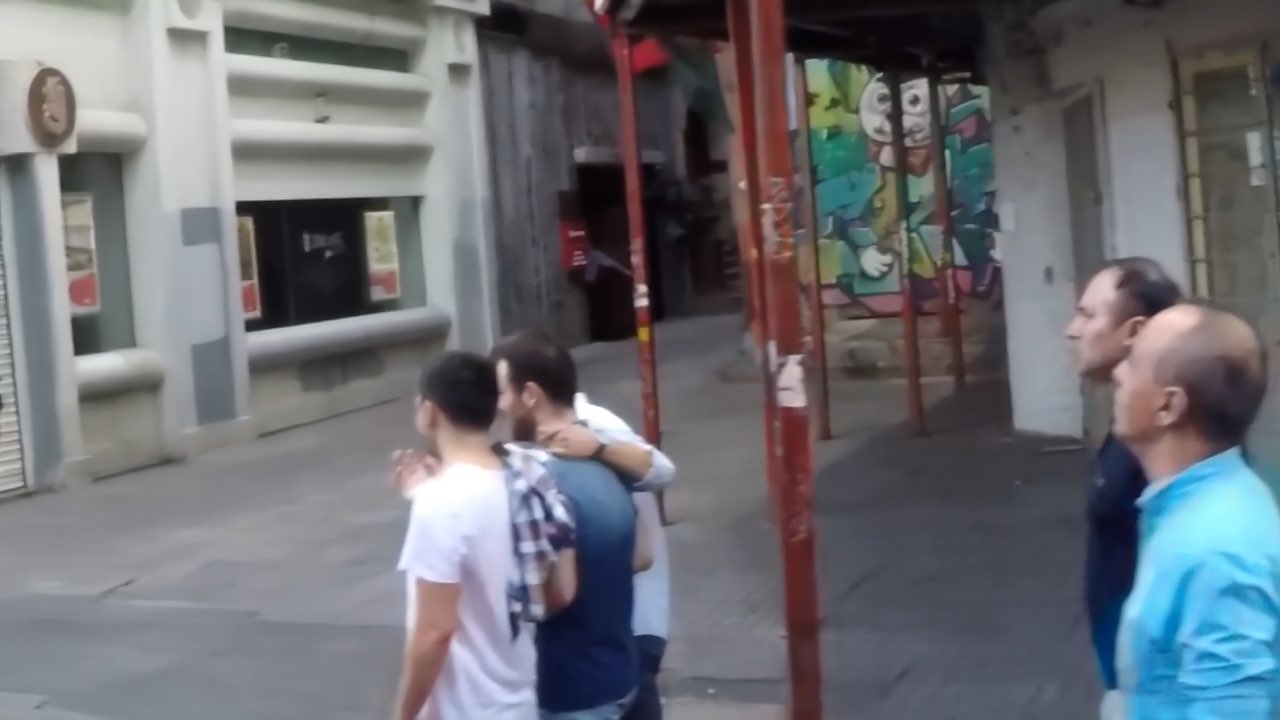} \\
        (a) Input & (b) GT & (c) DBGAN & (d) SRN & (e) HINet & (f) MPRNet & (g) Restormer \\
        
        \includegraphics[width=2.5cm]{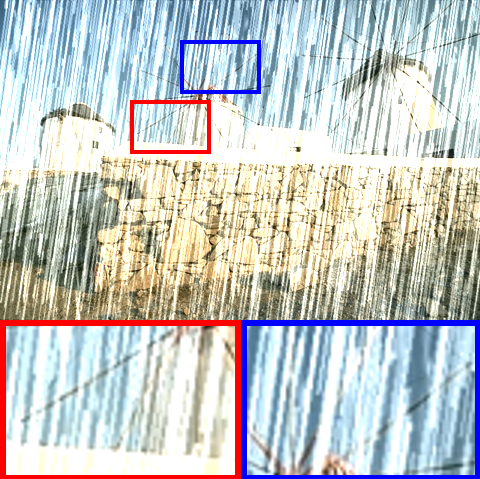} \hspace{-4mm} &
        \includegraphics[width=2.5cm]{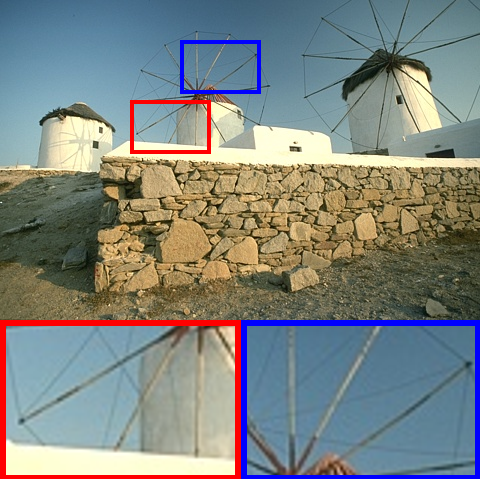} \hspace{-4mm} &
        \includegraphics[width=2.5cm]{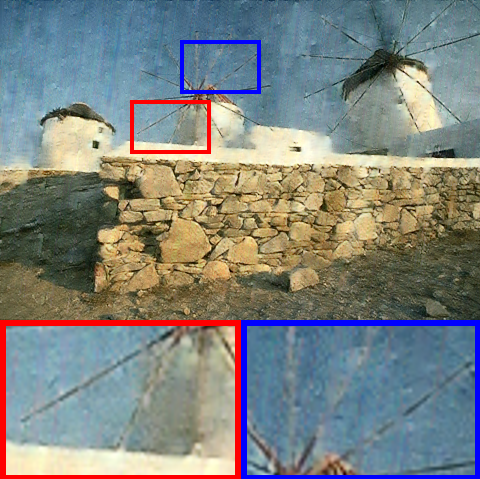} \hspace{-4mm} &
        \includegraphics[width=2.5cm]{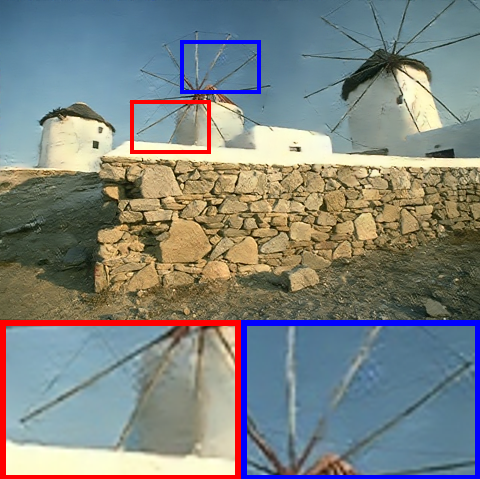} \hspace{-4mm} &
        \includegraphics[width=2.5cm]{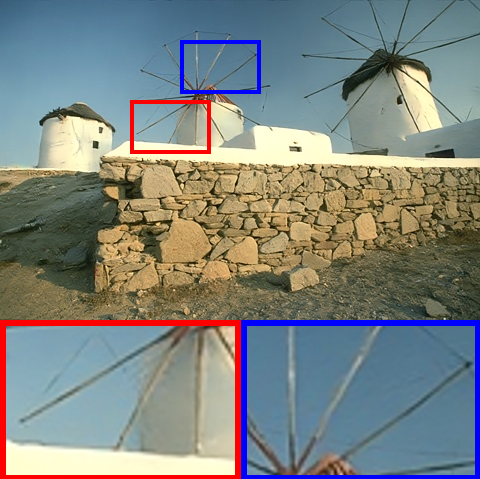} \hspace{-4mm} &
        \includegraphics[width=2.5cm]{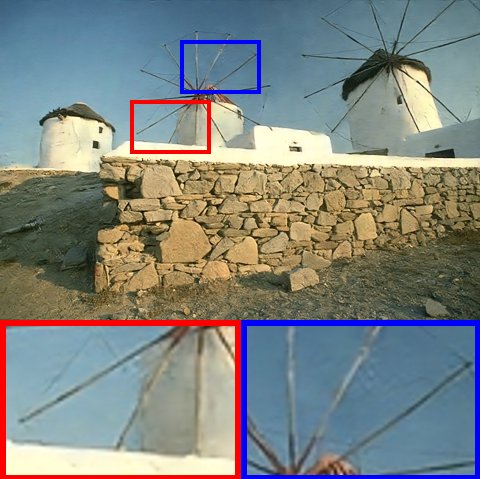} \hspace{-4mm} &
        \includegraphics[width=2.5cm]{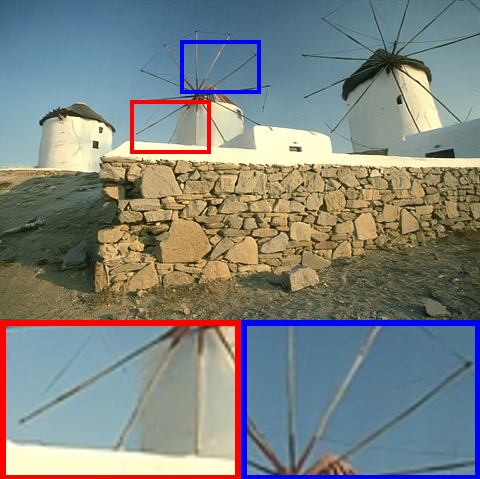} \\
        (a) Input & (b) GT & (c) DDN & (d) PReNet & (e) DRSformer & (f) SPDNet & (g) Restormer \\

        \includegraphics[width=2.5cm]{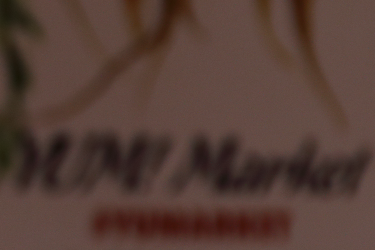} \hspace{-4mm} &
        \includegraphics[width=2.5cm]{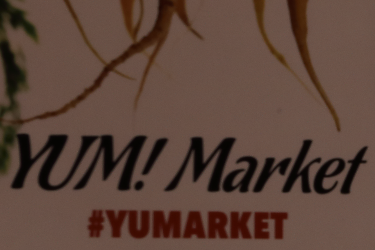} \hspace{-4mm} &
        \includegraphics[width=2.5cm]{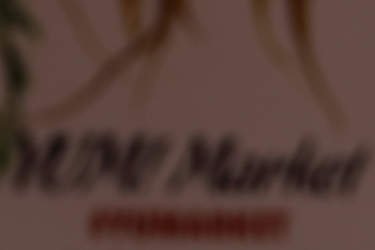} \hspace{-4mm} &
        \includegraphics[width=2.5cm]{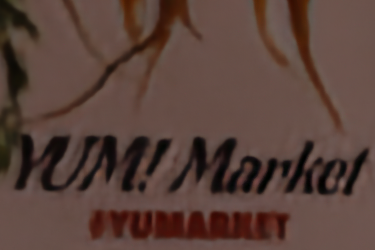} \hspace{-4mm} &
        \includegraphics[width=2.5cm]{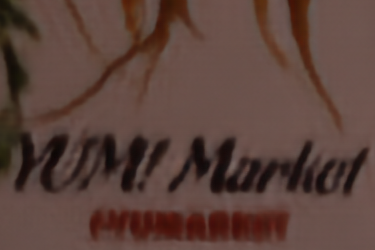} \hspace{-4mm} &
        \includegraphics[width=2.5cm]{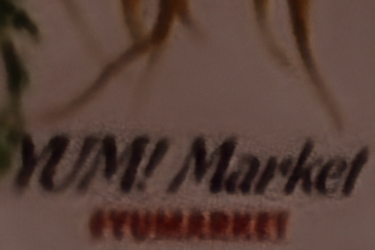} \hspace{-4mm} &
        \includegraphics[width=2.5cm]{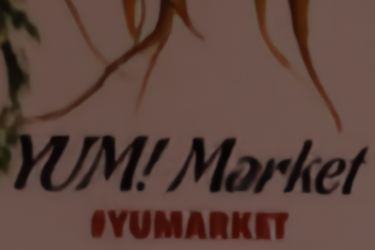} \\
        (a) Input & (b) GT & (c) SFNet & (d) IFAN & (e) DMENet & (f) MPRNet & (g) Restormer \\
    \end{tabular}}
\end{center}
\vspace{-6mm}
\caption{Visual comparisons with state-of-the-art methods on the GoPro, Rain100 and DPDD datasets. All comparison methods are retrained on the Restormer synthesised dataset.}
\vspace{-4mm}
\label{fig:4rain_removal_comparison}
\end{figure*}

We summarize the quantitative performance of our TripleD method against the full-dataset training and single-GPU baselines in Table~\ref{tab:quant_results} and Figure~\ref{fig:4rain_removal_comparison}. Across various datasets and tasks, TripleD consistently achieves close to 90$\sim$95\% of the performance of full-dataset training, despite using only 2\% of the data. Notably, TripleD significantly reduces memory and computational costs, allowing it to run efficiently on a single GPU.
In the supplementary material, we show some real-world scenarios for our algorithm.

%%%%%%%%%%%%%%%%%%%%%%%%%%%%%%

%%%%%%%%%%%%%%%%%%%%%%%%%%%%%%%%
\begin{table*}[htpb]
\caption{Performance comparison between Restormer (multi-GPU), our TripleD approach (Restormer is retrained on our synthetic dataset ), and Restormer (single-GPU, large-resolution images are processed in chunks) across different image restoration tasks. PSNR/SSIM values are reported for each method, and tasks that cannot be trained on a single GPU are marked with a "-".}
\vspace{-4mm}
\label{tab:quant_results}
\begin{center}
\scalebox{0.95}{
\begin{tabular}{c|c|ccc}
\toprule[0.15em]
\multicolumn{1}{c|}{\textbf{Dataset}} & \multicolumn{1}{c|}{\textbf{Task}} & \textbf{Restormer (Multi-GPU)} & \textbf{Ours (Single-GPU)} & \textbf{Restormer (Single-GPU)} \\ 
\midrule[0.15em]
\multicolumn{1}{l|}{Rain100L~\cite{yang2017deep}} 
                 & \multicolumn{1}{c|}{Deraining}    
                 & 36.52 / 0.925
                 & 32.08 / 0.910
                 & 33.22 / 0.921\\

\multicolumn{1}{l|}{Rain100H~\cite{yang2017deep}}
                 & \multicolumn{1}{c|}{Deraining}    
                 & 29.46 / 0.851
                 & 28.05 / 0.823
                 & 28.89 / 0.840\\ 

\multicolumn{1}{l|}{Test100~\cite{zhang2018image}}
                 & \multicolumn{1}{c|}{Deraining}    
                 & 30.12 / 0.867
                 & 28.25 / 0.830
                 & 29.01 / 0.849 \\ 

\multicolumn{1}{l|}{Test1200~\cite{zhang2018image}}
                 & \multicolumn{1}{c|}{Deraining}    
                 & 32.46 / 0.899
                 & 30.78 / 0.875
                 & 31.11 / 0.876 \\ \hline

\multicolumn{1}{l|}{GoPro~\cite{nah2017deep}} 
                 & \multicolumn{1}{c|}{Deblurring}    
                 & 30.25 / 0.935
                 & 29.12 / 0.889
                 & 26.50 / 0.880 \\ 

\multicolumn{1}{l|}{RealBlur-J~\cite{rim2020real}}
                 & \multicolumn{1}{c|}{Real-world Deblurring}    
                 & 33.12 / 0.952
                 & 31.67 / 0.938 
                 & 30.80 / 0.930 \\ 

\multicolumn{1}{l|}{RealBlur-R~\cite{rim2020real}}
                 & \multicolumn{1}{c|}{Real-world Deblurring}    
                 & 33.12 / 0.952
                 & 31.67 / 0.938
                 & 30.70 / 0.931 \\ 

\multicolumn{1}{l|}{DPDD~\cite{abuolaim2020defocus}}
                 & \multicolumn{1}{c|}{Defocus Deblurring}    
                 & 25.98 / 0.811
                 & 24.21 / 0.802
                 & 23.90 / 0.790 \\ 

\multicolumn{1}{l|}{SIDD~\cite{abdelhamed2018high}}
                 & \multicolumn{1}{c|}{Denoising}    
                 & 40.02 / 0.955
                 & 38.12 / 0.919 
                 & 37.50 / 0.910 \\ 

\multicolumn{1}{l|}{DND~\cite{plotz2017benchmarking}} 
                 & \multicolumn{1}{c|}{Denoising}    
                 & 40.03 / 0.956
                 & 38.56 / 0.932
                 & 35.99 / 0.925 \\ 
\bottomrule[0.15em]
\end{tabular}}
\end{center}\vspace{-4mm}
\end{table*}

%\begin{table*}[!thpb]
%\caption{Performance comparison between static and dynamic dataset selection across different image restoration tasks. PSNR and SSIM values are reported, along with the number of epochs needed to reach convergence. Dynamic selection consistently outperforms static selection, especially in challenging tasks such as deblurring.}
%\vspace{-4mm}
%\label{tab:sd-quant_results}
%\begin{center}
%\scalebox{0.92}{
%\begin{tabular}{c|c|ccc}
%\toprule[0.15em]
%\textbf{Dataset} & \textbf{Task} & \textbf{Static Selection} & \textbf{Dynamic Selection (TripleD)} & \textbf{Convergence Speed (Epochs)} \\ 
%\midrule[0.15em]
%GoPro~\cite{nah2017deep} & Deblurring & 25.80 / 0.870 & 27.12 / 0.889 & Static: 450, %Dynamic: 350 \\ 
%Rain100L~\cite{yang2017deep} & Deraining & 30.50 / 0.900 & 32.08 / 0.910 & Static: 400, Dynamic: 300 \\ 
%SIDD~\cite{abdelhamed2018high} & Denoising & 37.00 / 0.915 & 38.12 / 0.919 & Static: 600, Dynamic: 450 \\ 
%RealBlur-J~\cite{rim2020real} & Real-world Deblurring & 30.00 / 0.910 & 31.67 / 0.938 & Static: 500, Dynamic: 400 \\ 
%\bottomrule[0.15em]
%\end{tabular}}
%vspace{-6mm}
%\end{center}
%\end{table*}

\subsection{Experimental Analysis}
In this section, we show some quantitative and qualitative analyses to explain the effectiveness of our method.

\begin{figure}[ht]
  \centering
  \includegraphics[width=0.48\textwidth]{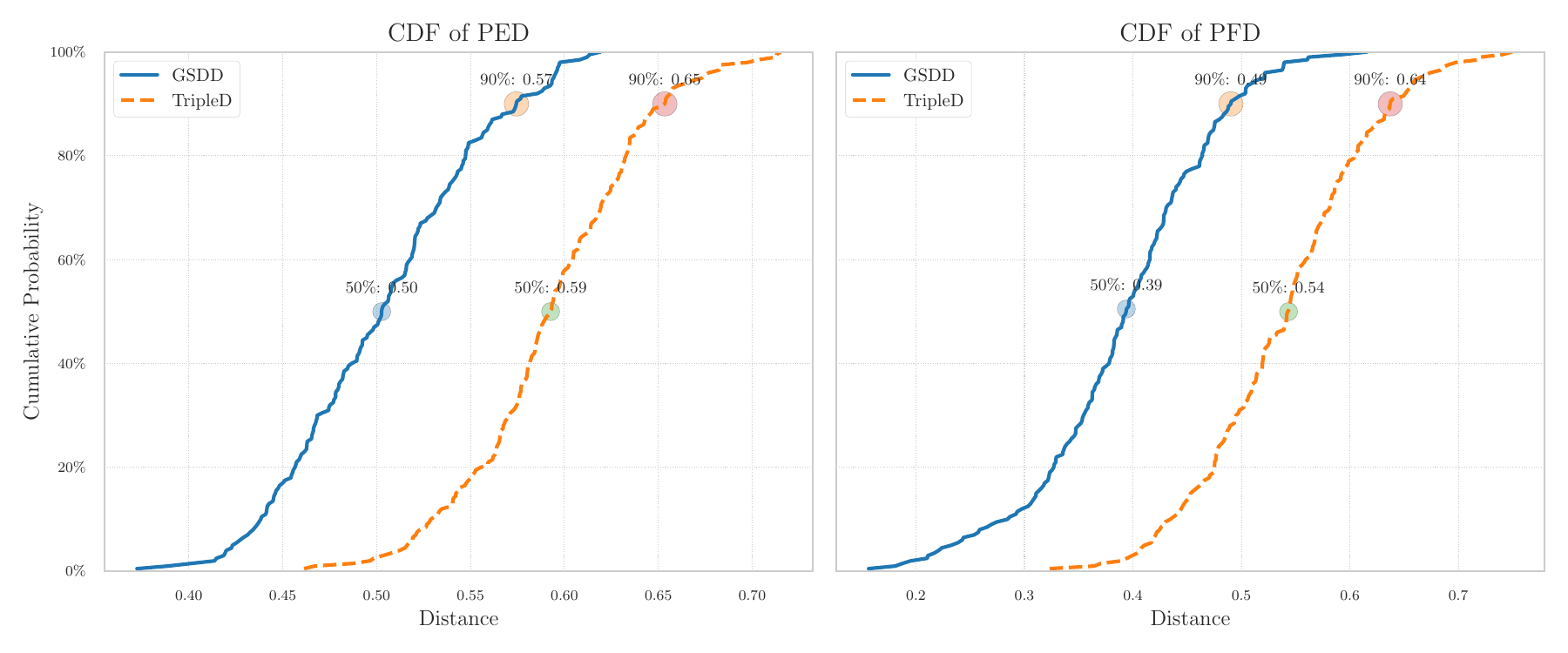}
  \caption{CDFs of pairwise distances between distilled‑sample features under GSDD vs.\ TripleD. The left half shows PED and the right half shows PFD. In both metrics, the TripleD curves are right‑shifted, indicating higher sample diversity.}
  \label{fig:feature_distance_cdf}
\end{figure}

\noindent \textbf{Diversity Analysis of Distilled Samples}
To quantitatively compare the diversity of distilled samples under GSDD and our TripleD strategy, we compute the pairwise Euclidean distances (PED) and pairwise feature distances (PFD) between feature embeddings (extracted by ResNet‑50) of 200 distilled samples, and plot their cumulative distribution functions (CDFs). As shown in Figure~\ref{fig:feature_distance_cdf}, TripleD exhibits a right‐shifted CDF in both metrics—indicating higher sample diversity, which helps the model generalize better under extreme pruning ratios.

%\noindent \textbf{Static vs Dynamic Dataset Selection.}
%
%We compare the performance of static and dynamic dataset selection strategies in the context of image restoration tasks. Static selection involves selecting a fixed subset of the training data at the start of the training process and using this subset throughout. 
%
%As shown in Table~\ref{tab:sd-quant_results}, our method achieves better performance than static data selection using a minimum number of iterations.
%
%In addition, our method converges faster than the static method for image restoration models with the same hardware configuration.
%

\noindent \textbf{Effectiveness of CNN.}
In the Rain100L deraining task, we experimented with different CNNs to evaluate their effectiveness. Specifically, we compare the performance of CNNs with different layers. Here, we do not consider the computationally expensive Transformer. 
Table~\ref{tab:cnn_comparison_rain100l} summarizes the results, where we report PSNR and SSIM for each CNN when applied in our framework.

\noindent \textbf{Effectiveness of Downsampling.}
In the Rain100L deraining task, we experimented with different downsampling resolutions ($64 \times 64$, $384 \times 384$, and full resolution) to evaluate their effectiveness. 
Table~\ref{tab:d_comparison_rain100l} summarizes the results, where we report PSNR and SSIM for each downsampling resolution when applied in our framework.
Figure~\ref{fig:distribution_alignment} summarizes the distribution alignment achieved by different CNN adjustments, which visually demonstrates the improvements as the adjusted dataset more closely aligns with the target distribution in our framework.

\noindent \textbf{Effectiveness of the proportion of synthetic dataset $p$.}
In the Rain100L deraining task, we experimented with different distillation hyperparameters $p$ ($1\%$, $5\%$, and $10\%$) to evaluate their effectiveness.
Table~\ref{tab:p_comparison_rain100l} summarizes the results, where we report PSNR and SSIM for $p$ when applied in our framework.

%\noindent \textbf{Effectiveness of feature extractor.}
%
%We evaluated the performance of various feature extractors (see Table~\cite{tab:feature_comparison_rain100l}), among which the ViT achieved the best results. They all used the same training strategy.
%
%Additionally, we also evaluated their inference speeds. ResNet-50 had the fastest inference speed, while Mamba was the slowest. Our method achieved a good balance between speed and accuracy.
%%In the Rain100L deraining task, we experimented with different group hyperparameter $\alpha$ and $\beta$ ($\{0.2, 0.3, 0.4\}$ (g1), $\{0.2, 0.5, 0.7\}$ (g2) and $\{0.5, 0.6, 0.7\}$ (g3)) to evaluate their effectiveness.
%able~\ref{tab:ab_comparison_rain100l} summarizes the results, where we report PSNR and SSIM for $\alpha$ and $\beta$ when applied in our framework.

\begin{table}[!t]
\caption{Performance of CNN. Only CNNs with 8 layers have good results, and it is not the case that the more convolutional layers the better.}
\vspace{-4mm}
\label{tab:cnn_comparison_rain100l}
\begin{center}
\scalebox{0.95}{
\begin{tabular}{c|c|c}
\toprule[0.15em]
\textbf{Layers of CNN} & \textbf{PSNR (dB)} & \textbf{SSIM} \\ 
\midrule[0.15em]
0 & 30.33 & 0.897 \\ 
4 & 31.12 & 0.889 \\ 
16 & 32.05 & 0.908 \\
Ours & 32.08 & 0.910\\
\bottomrule[0.15em]
\end{tabular}}\vspace{-4mm}
\end{center}
\end{table}

\begin{table}[!t]
\caption{Performance of downsampling. Although full resolution performs better, it takes more than \textbf{50 $\times$} longer.}
\vspace{-4mm}
\label{tab:d_comparison_rain100l}
\begin{center}
\scalebox{0.95}{
\begin{tabular}{c|c|c}
\toprule[0.15em]
\textbf{Resolution of downsampling} & \textbf{PSNR (dB)} & \textbf{SSIM} \\ 
\midrule[0.15em]
$64 \times 64$ & 27.11 & 0.844 \\ 
Ours & 32.08 & 0.910 \\ 
$384 \times 384$ & 31.89 & 0.911 \\
Full resolution & 33.21 & 0.912\\
\bottomrule[0.15em]
\end{tabular}}\vspace{-4mm}
\end{center}
\end{table}

\begin{table}[!t]
\caption{Performance of different feature extractors for dynamic sample selection on Rain100L deraining task.}
\vspace{-4mm}
\label{tab:feature_comparison_rain100l}
\begin{center}
\scalebox{0.95}{
\begin{tabular}{c|c|c}
\toprule[0.15em]
\textbf{Feature Extractor} & \textbf{PSNR (dB)} & \textbf{SSIM} \\ 
\midrule[0.15em]
ResNet-50 & 30.50 & 0.900 \\ 
ViT & 32.08 & 0.910 \\ 
Mamba & 31.10 & 0.879 \\
MLP-Mixer & 32.05 & 0.901\\
\bottomrule[0.15em]
\end{tabular}}\vspace{-4mm}
\end{center}
\end{table}

\begin{figure}[!t]
\begin{center}
\scalebox{0.62}{
    \begin{tabular}{cccccc}
        % 第一行图像
        \includegraphics[width=1.8cm]{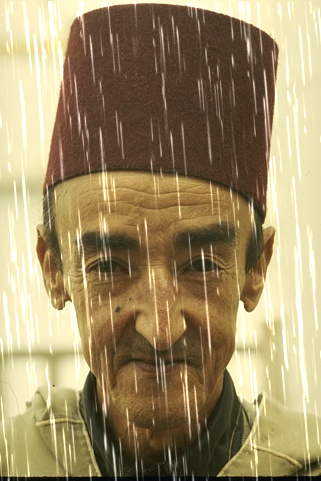} &
        \includegraphics[width=1.8cm]{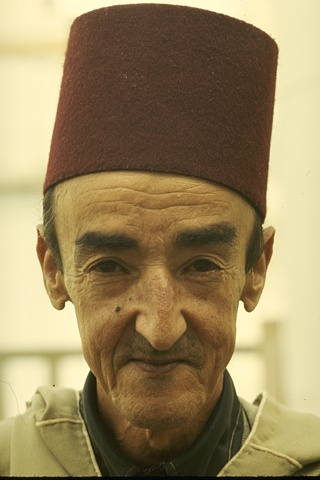} &
        \includegraphics[width=1.8cm]{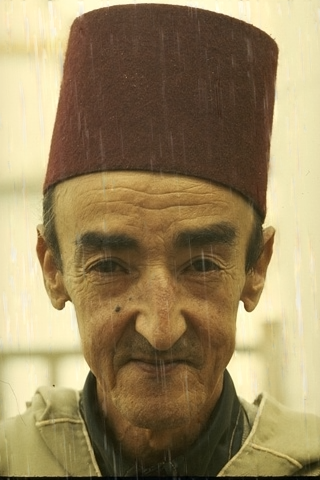} &
        \includegraphics[width=1.8cm]{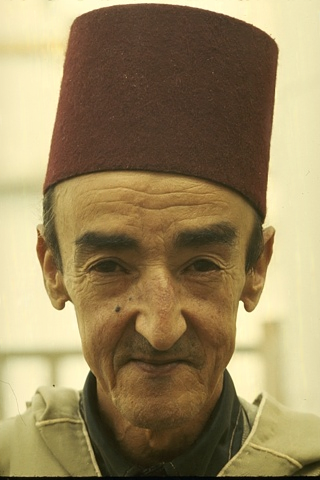}&
       \includegraphics[width=1.8cm]{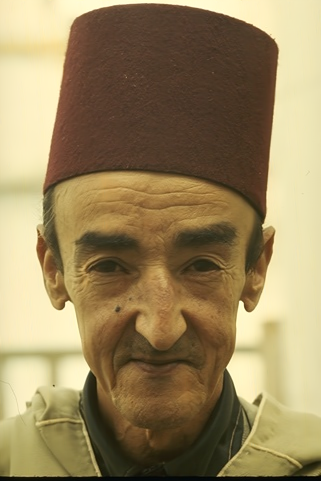} &
        \includegraphics[width=1.8cm]{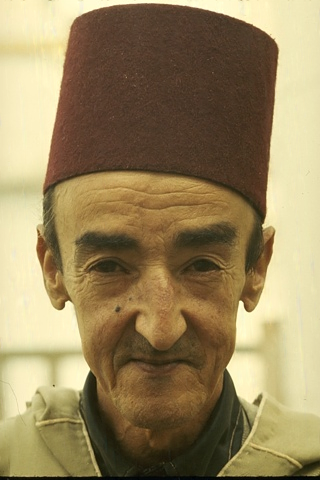}\\
        
        % 第二行图像
       \includegraphics[width=1.8cm]{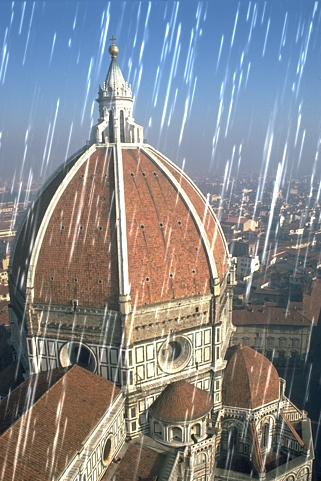} &
      \includegraphics[width=1.8cm]{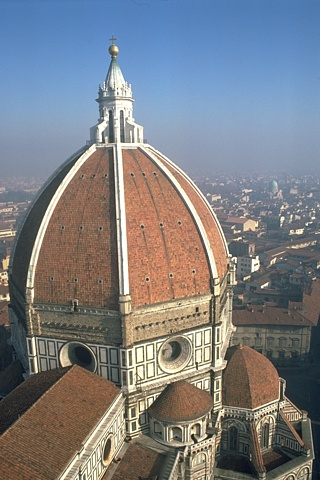} &
        \includegraphics[width=1.8cm]{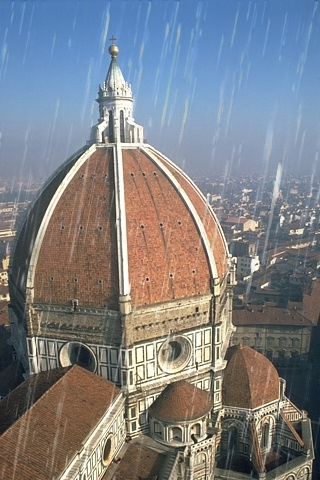} &
        \includegraphics[width=1.8cm]{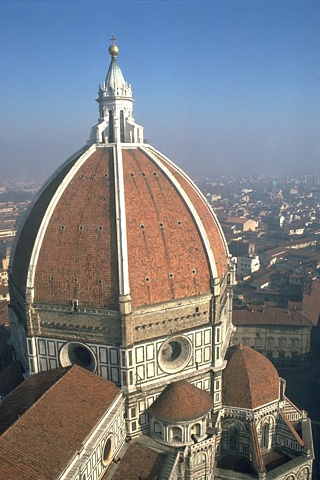} &
        \includegraphics[width=1.8cm]{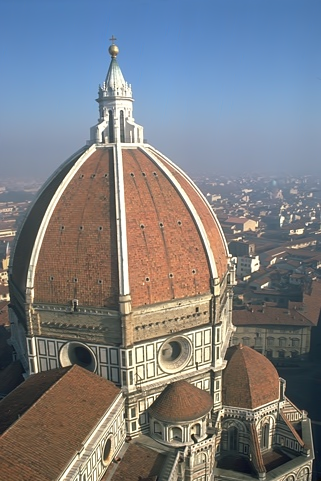}&
        \includegraphics[width=1.8cm]{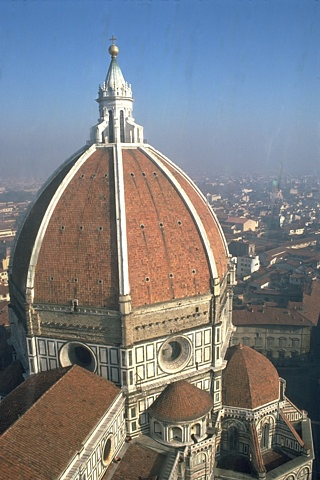} \\
        (a) Input & (b) GT& ResNet-50& (c) ViT & (d) Mamba &  (e) Ours\\
        PSNR (mean) & (b) +$\infty$ & 31.4dB& (c) 32.1dB & (d) 32.2dB &   32.8dB\\
    \end{tabular}}
    \end{center}
    \vspace{-4mm}
    \caption{Visual comparison of rain removal results for two samples from the Rain100L dataset using various models: Input, Ground Truth (GT), ResNet-50, ViT, CLIP, and ViT+CLIP.}
    \vspace{-4mm}
    \label{fig:clip-rain_removal_comparison}
\end{figure}

\begin{table}[!t]
\caption{Performance of $p$. To trade off accuracy and efficiency, we use 2\% of the data size.}
\vspace{-4mm}
\label{tab:p_comparison_rain100l}
\begin{center}
\scalebox{0.95}{
\begin{tabular}{c|c|c}
\toprule[0.15em]
\textbf{Distillation hyperparameter $p$} & \textbf{PSNR (dB)} & \textbf{SSIM} \\ 
\midrule[0.15em]
$1\%$ & 25.33 & 0.831 \\ 
Ours & 32.08 & 0.910 \\ 
$5\%$ & 32.09 & 0.912 \\
$10\%$ & 33.14 & 0.925 \\
\bottomrule[0.15em]
\end{tabular}}\vspace{-4mm}
\end{center}
\end{table}

%\begin{table}[!t]
%%\caption{Performance of $\alpha$ and $\beta$. To trade off accuracy and efficiency, we use $\{0.1, 0.5, 0.7\}$ of $\alpha$ and $\beta$.}
%%\vspace{-4mm}
%\label{tab:ab_comparison_rain100l}
%\begin{center}
%\scalebox{0.95}{
%\begin{tabular}{c|c|c}
%\toprule[0.15em]
%\textbf{$\alpha$ and $\beta$} & \textbf{PSNR (dB)} & \textbf{SSIM} \\ 
%\midrule[0.15em]
%g1 & 29.92 & 0.906 \\ 
%g2 & 30.88 & 0.901 \\ 
%g3 & 32.01 & 0.905 \\
%Ours & 32.08 & 0.910 \\
%\bottomrule[0.15em]
%\end{tabular}}\vspace{-4mm}
%\end{center}
%\end{table}

\subsection{More Ablation Experiments}
To further quantify the contribution of each module, we perform three additional ablation studies on the Rain100L deraining task. Table~\ref{tab:ablate-cnn}--Table~\ref{tab:ablate-gen} summarize the results.

\begin{table}[!t]
  \caption{Effect of removing the CNN fine-tuning module on Rain100L deraining.}
  \vspace{-2mm}
  \label{tab:ablate-cnn}
  \centering
  \scalebox{0.95}{
  \begin{tabular}{l|cc}
    \toprule[0.15em]
    \textbf{Configuration}                   & \textbf{PSNR (dB)} & \textbf{SSIM} \\
    \midrule[0.15em]
    Full TripleD (Ours)                      & 32.08              & 0.910         \\
    w/o CNN Fine-Tuning                  & 30.75              & 0.894         \\
    \bottomrule[0.15em]
  \end{tabular}}
  \vspace{-2mm}
\end{table}

\begin{table}[!t]
  \caption{Comparison of generative backbones for data distillation.}
  \vspace{-2mm}
  \label{tab:ablate-gen}
  \centering
  \scalebox{0.95}{
  \begin{tabular}{l|cc}
    \toprule[0.15em]
    \textbf{Backbone}                         & \textbf{PSNR (dB)} & \textbf{SSIM} \\
    \midrule[0.15em]
    SD3 Diffusion (Ours)                     & 32.08              & 0.910         \\
    StyleGAN2                                & 31.02              & 0.903         \\
    \bottomrule[0.15em]
  \end{tabular}}
  \vspace{-2mm}
\end{table}

%\noindent \textbf{Training Process Analysis.}
%
%To demonstrate the effectiveness of dynamic dataset selection, we evaluated both static and dynamic selection methods on the GoPro dataset. The convergence of the loss over 500 epochs is presented in Figure \ref{fig:loss_convergence}. The loss was tracked for both static and dynamic selection during training.
%
%As seen in Figure \ref{fig:loss_convergence}, dynamic selection significantly reduces the loss faster than static selection. Static selection, being limited to a fixed subset, converges more slowly, indicating its inefficiency in adapting to different phases of training.
%\begin{figure}[!t]
%    \centering
%    \includegraphics[width=8.33cm]{figs/loss_convergence_static_dynamic_romans_font.png}
%    \vspace{-4mm}
%    \caption{Loss convergence on the GoPro dataset for the image deblurring task. TripleD method results in faster loss reduction compared to the static selection method.}
%    \vspace{-4mm}
%    \label{fig:loss_convergence}
%\end{figure}

%\noindent \textbf{Quantitative Results.}
%To provide a clearer comparison of the effectiveness of static and dynamic dataset selection, we report the performance across several standard image restoration tasks in Table \ref{tab:quant_results}. Dynamic selection consistently achieves higher PSNR and SSIM scores compared to static selection, demonstrating better generalization and superior restoration quality. 
\begin{figure*}[t!]
\tabcolsep 0.8pt
\centering
		\begin{tabular}{cccccc}

        \includegraphics[height=0.13\linewidth,width = 0.16\textwidth]{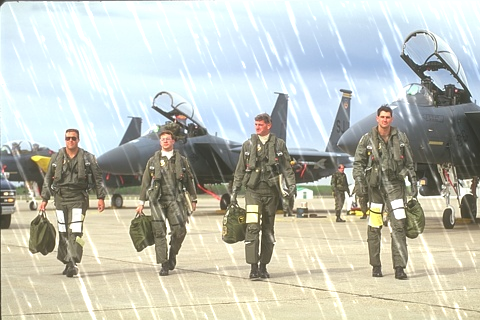}
        &\includegraphics[height=0.13\linewidth,width = 0.16\textwidth]{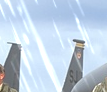}
        &\includegraphics[height=0.13\linewidth,width = 0.16\textwidth]{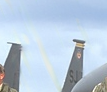} 
        &\includegraphics[height=0.13\linewidth,width = 0.16\textwidth]{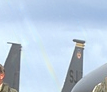} 
        &\includegraphics[height=0.13\linewidth,width = 0.16\textwidth]{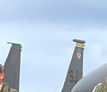} 
        &\includegraphics[height=0.13\linewidth,width = 0.16\textwidth]{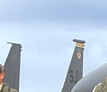} \\
        Degraded &27.49 dB&35.53 dB& 39.25 dB & 44.90 dB& PSNR\\ 
        \includegraphics[height=0.13\linewidth,width = 0.16\textwidth]{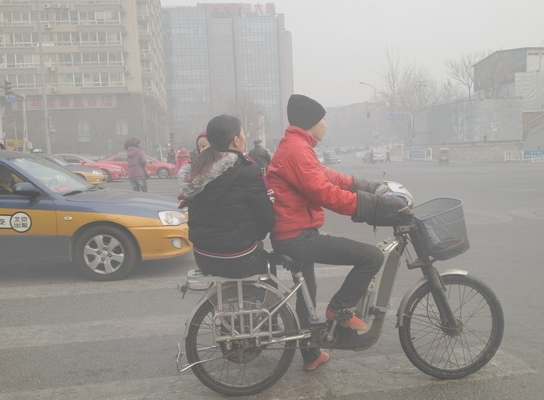}
        &\includegraphics[height=0.13\linewidth,width = 0.16\textwidth]{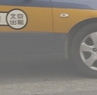}
        &\includegraphics[height=0.13\linewidth,width = 0.16\textwidth]{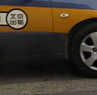}
        &\includegraphics[height=0.13\linewidth,width = 0.16\textwidth]{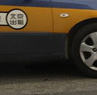} 
        &\includegraphics[height=0.13\linewidth,width = 0.16\textwidth]{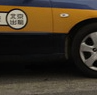} 
        &\includegraphics[height=0.13\linewidth,width = 0.16\textwidth]{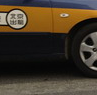} \\
        Degraded &10.82 dB&27.49 dB& 28.75 dB &31.68 dB& PSNR\\ 

        \includegraphics[height=0.13\linewidth,width = 0.16\textwidth]{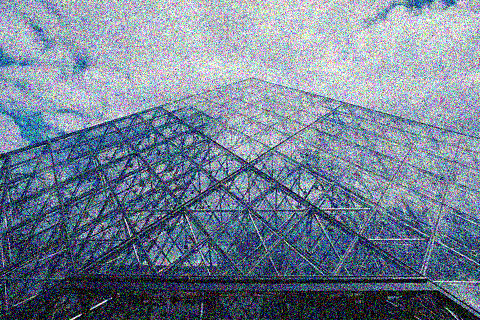}
        &\includegraphics[height=0.13\linewidth,width = 0.16\textwidth]{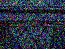}
        &\includegraphics[height=0.13\linewidth,width = 0.16\textwidth]{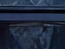}
        &\includegraphics[height=0.13\linewidth,width = 0.16\textwidth]{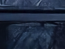} 
        &\includegraphics[height=0.13\linewidth,width = 0.16\textwidth]{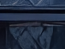} 
        &\includegraphics[height=0.13\linewidth,width = 0.16\textwidth]{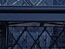} \\
        Degraded &15.63 dB&26.73 dB& 26.18 dB & 27.12 dB& PSNR\\

        Image & Input& AirNet & PromptIR   &  AdaIR & Reference\\
\end{tabular}
\vspace{-0.5em}
\caption{All-in-one model with TripleD. All comparison methods are retrained on the PromptIR synthesised dataset.}
\vspace{-4mm}
\label{fig:all-in-one}
\end{figure*}

\subsection{Training Optimization}

Due to the limited memory of a single NVIDIA RTX 3090 GPU (24GB), training large-scale models, such as Restormer, with full batch sizes is infeasible, especially when working with high-resolution images. To address this challenge, we employ gradient accumulation, which allows for the simulation of larger batch sizes without exceeding the memory limit by accumulating gradients over several mini-batches.

To demonstrate the efficacy of gradient accumulation, we conducted ablation studies across different accumulation steps (1, 4, and 8 steps) to evaluate its impact on memory usage, and restoration performance. Table~\ref{tab:grad_accum_comparison} presents the results of this experiment. We observe that while memory usage significantly decreases as the accumulation steps increase, the impact on PSNR and SSIM is minimal, demonstrating the robustness of this technique.

\begin{figure*}[!t]
    \centering
    \scalebox{0.56}{
    \begin{subfigure}[b]{0.17\textwidth}
        \includegraphics[width=\textwidth]{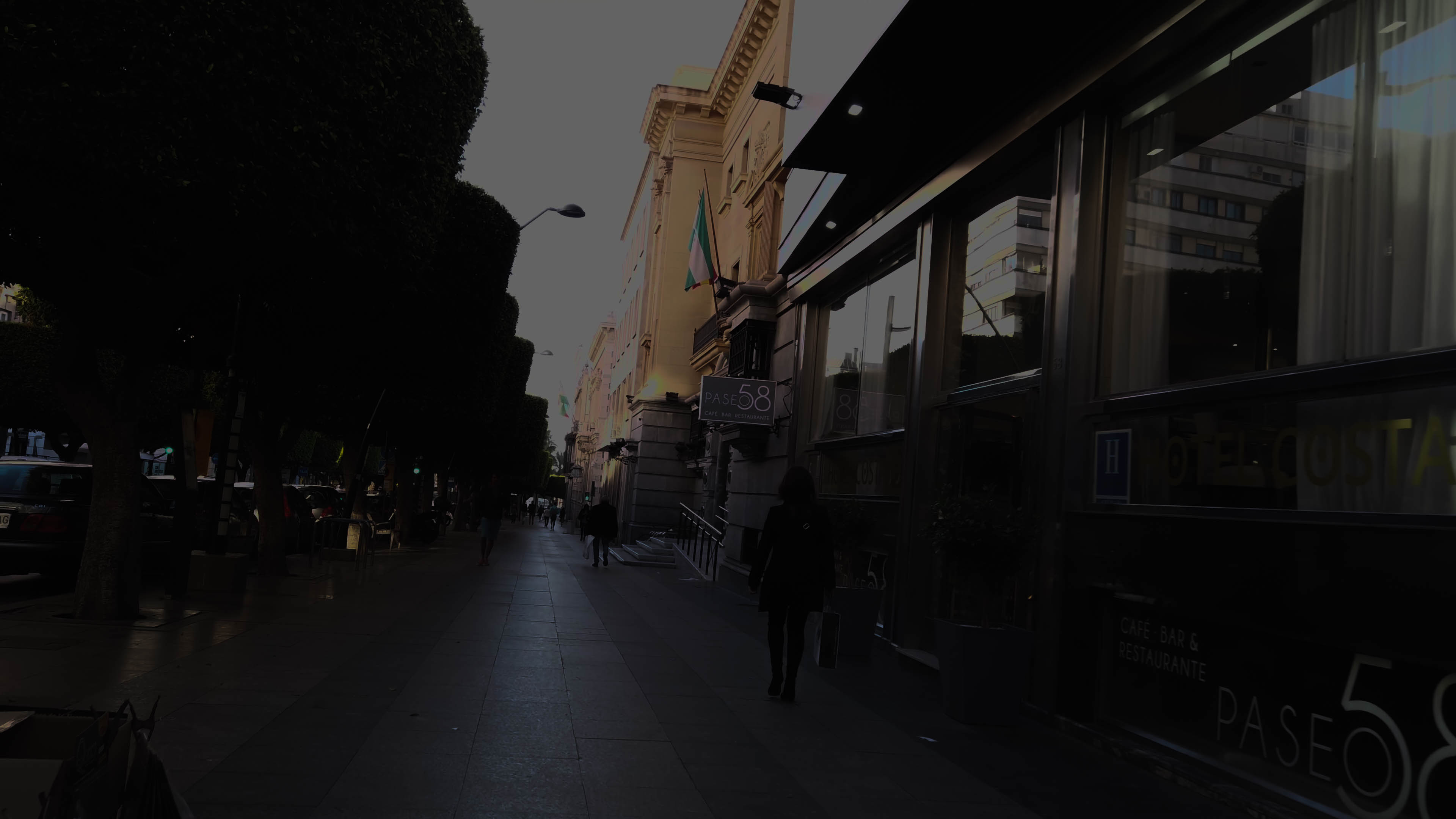}
        \caption{PSNR}
    \end{subfigure}
    \begin{subfigure}[b]{0.17\textwidth}
        \includegraphics[width=\textwidth]{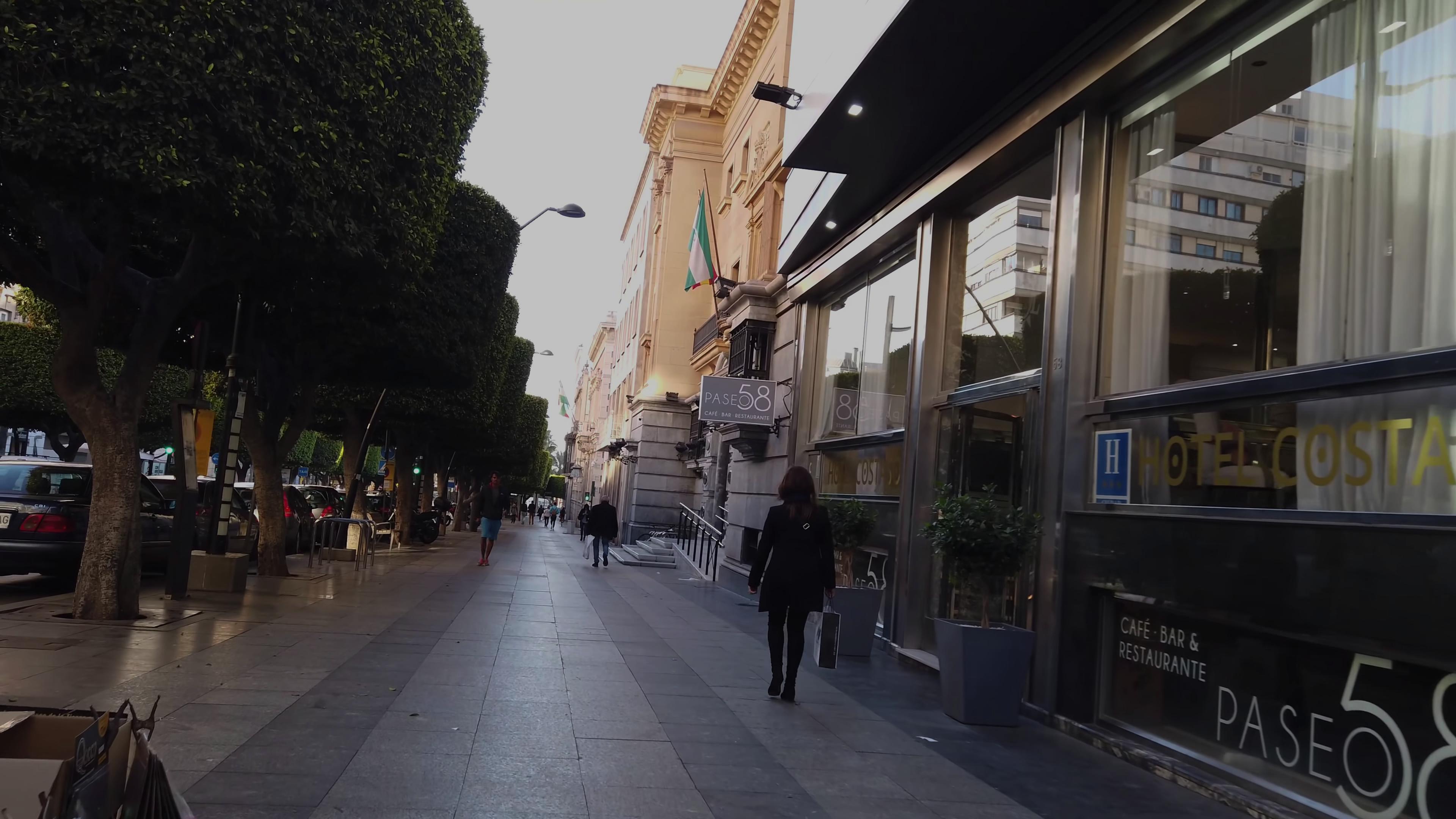}
        \caption{28.36}
    \end{subfigure}
    \begin{subfigure}[b]{0.17\textwidth}
        \includegraphics[width=\textwidth]{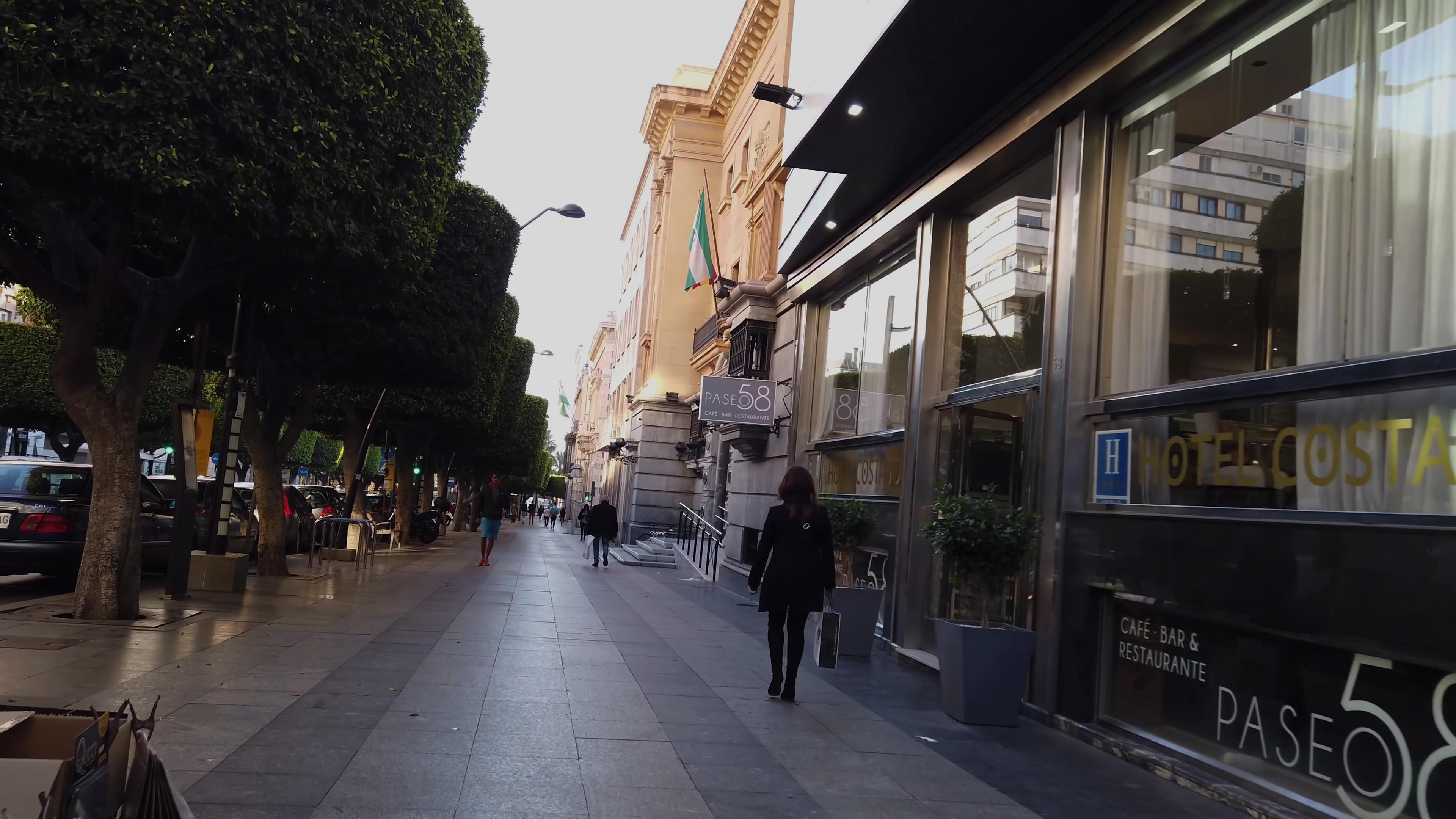}
        \caption{27.98}
    \end{subfigure}
    \begin{subfigure}[b]{0.17\textwidth}
        \includegraphics[width=\textwidth]{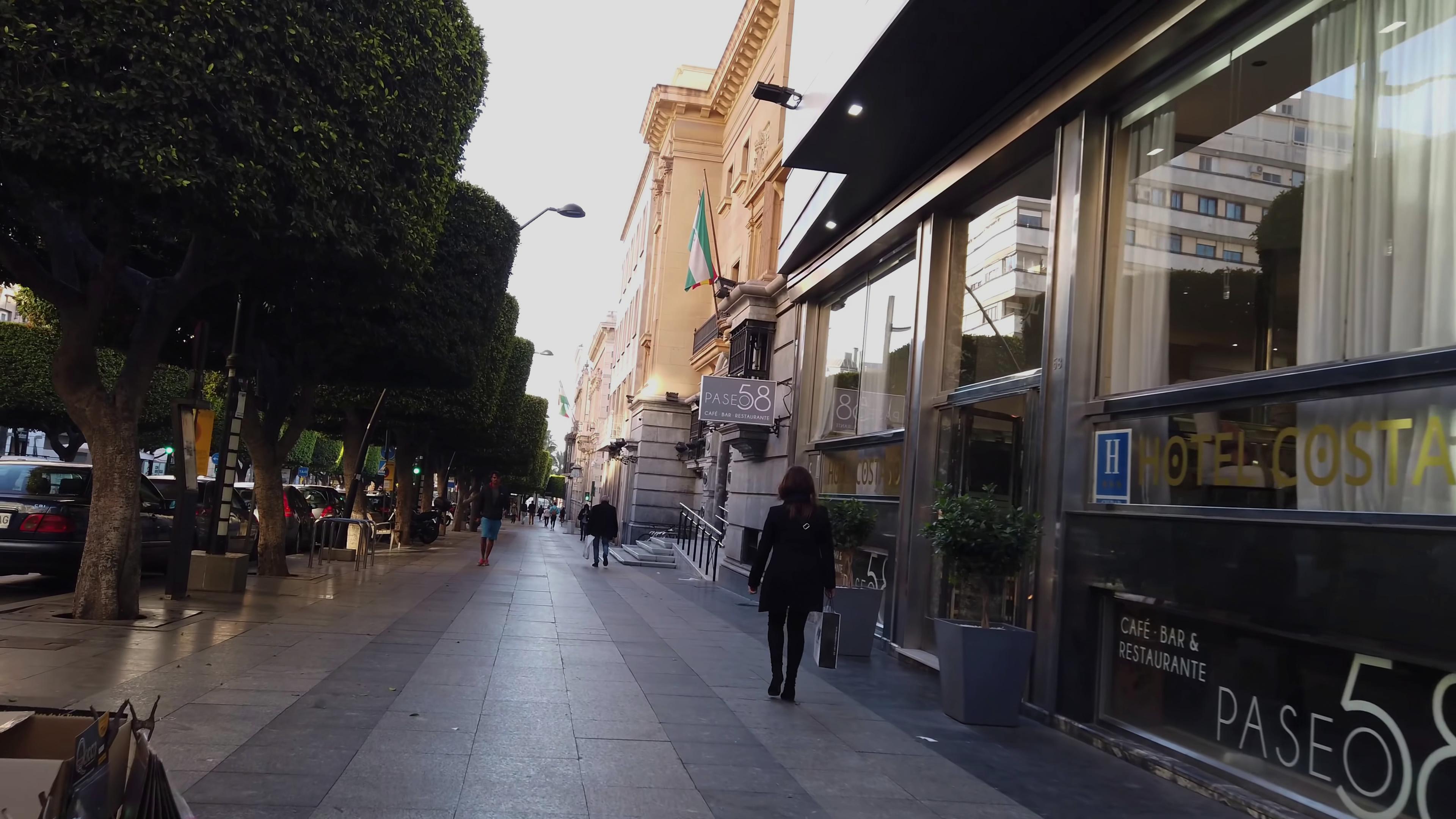}
        \caption{29.03}
    \end{subfigure}
    \begin{subfigure}[b]{0.17\textwidth}
        \includegraphics[width=\textwidth]{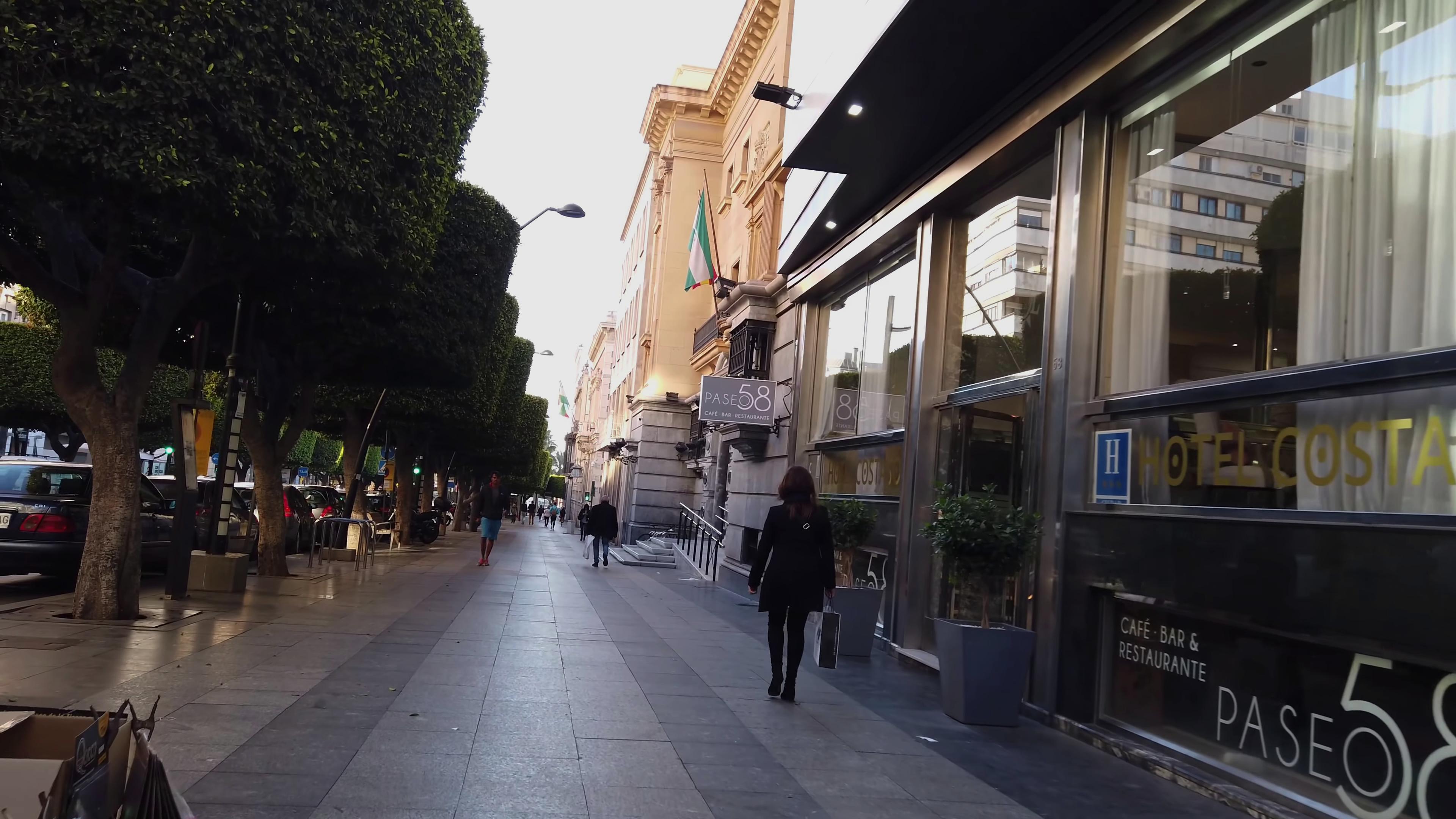}
        \caption{28.86}
    \end{subfigure}
    \begin{subfigure}[b]{0.17\textwidth}
        \includegraphics[width=\textwidth]{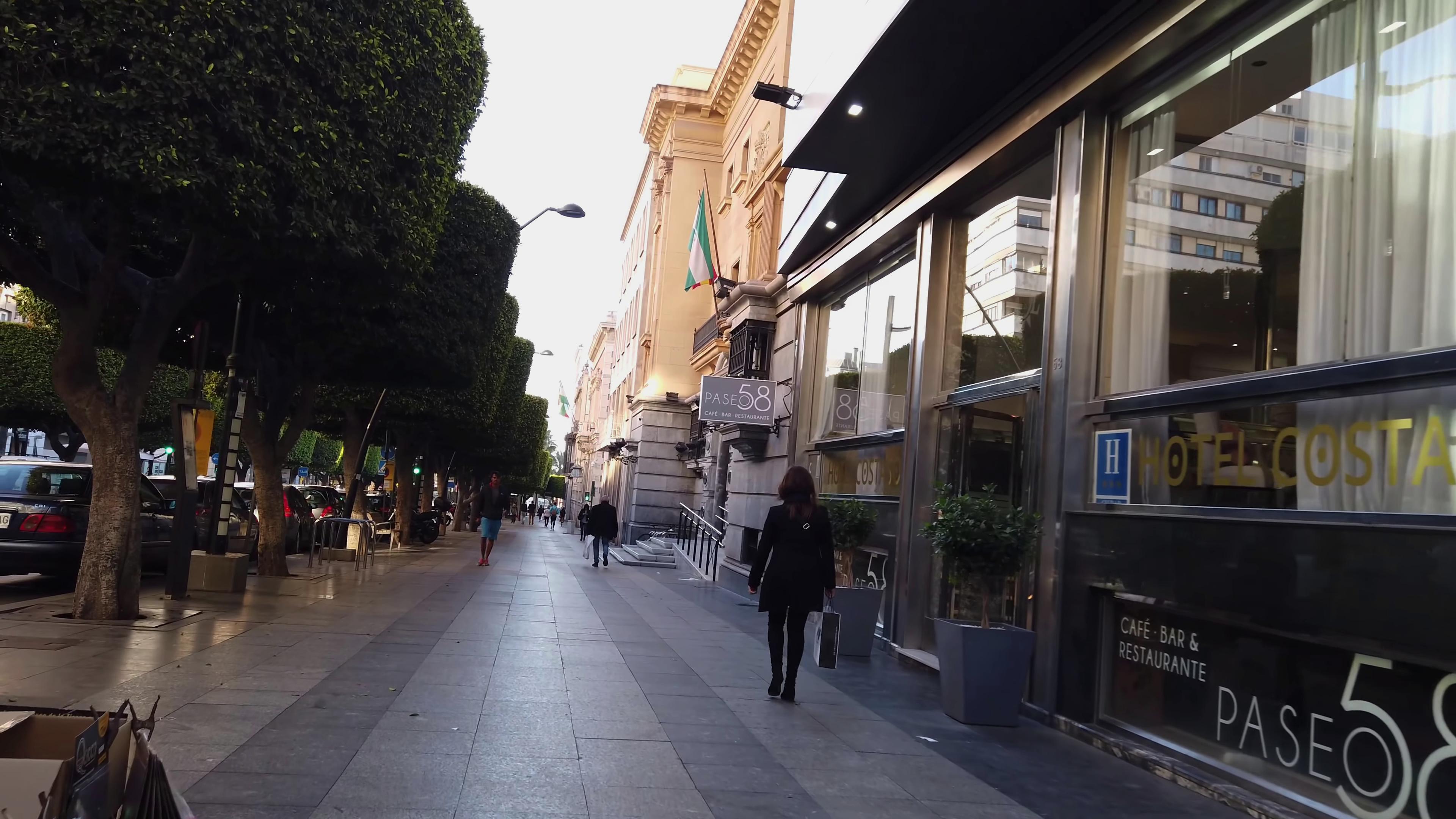}
        \caption{29.43}
    \end{subfigure}
    \begin{subfigure}[b]{0.17\textwidth}
        \includegraphics[width=\textwidth]{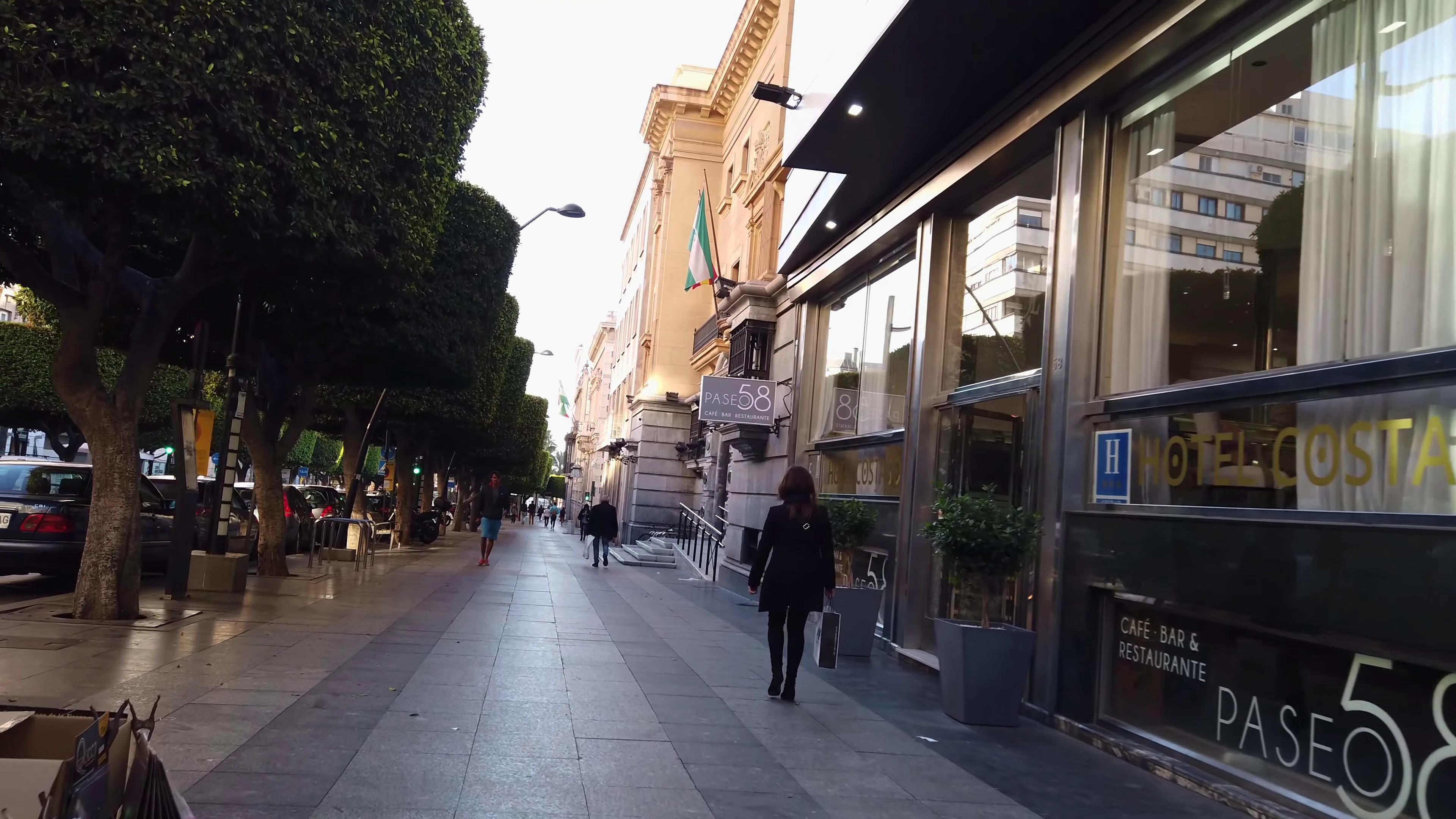}
        \caption{29.60}
    \end{subfigure}
    \begin{subfigure}[b]{0.17\textwidth}
        \includegraphics[width=\textwidth]{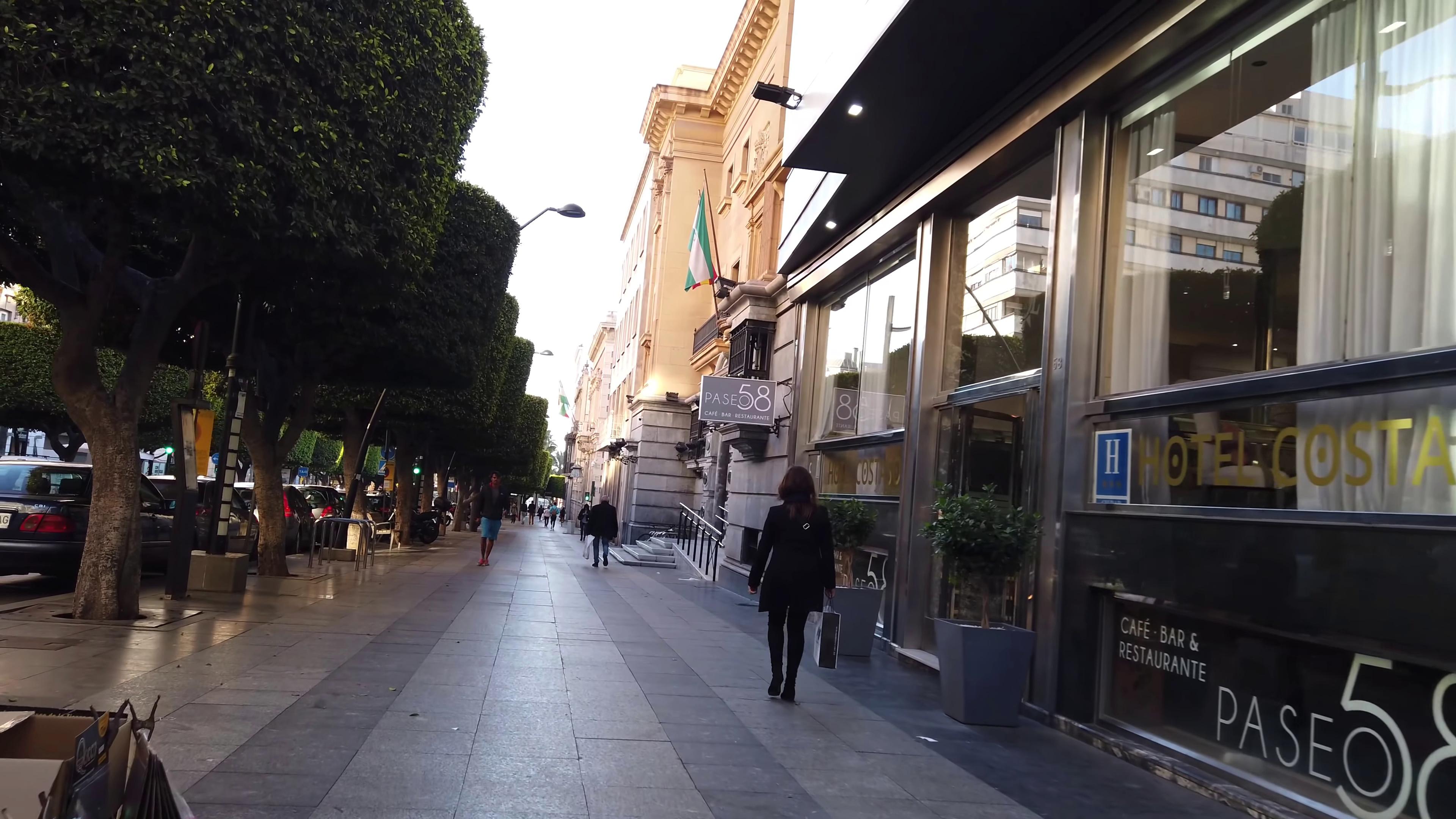}
        \caption{31.88}
    \end{subfigure}
    \begin{subfigure}[b]{0.17\textwidth}
        \includegraphics[width=\textwidth]{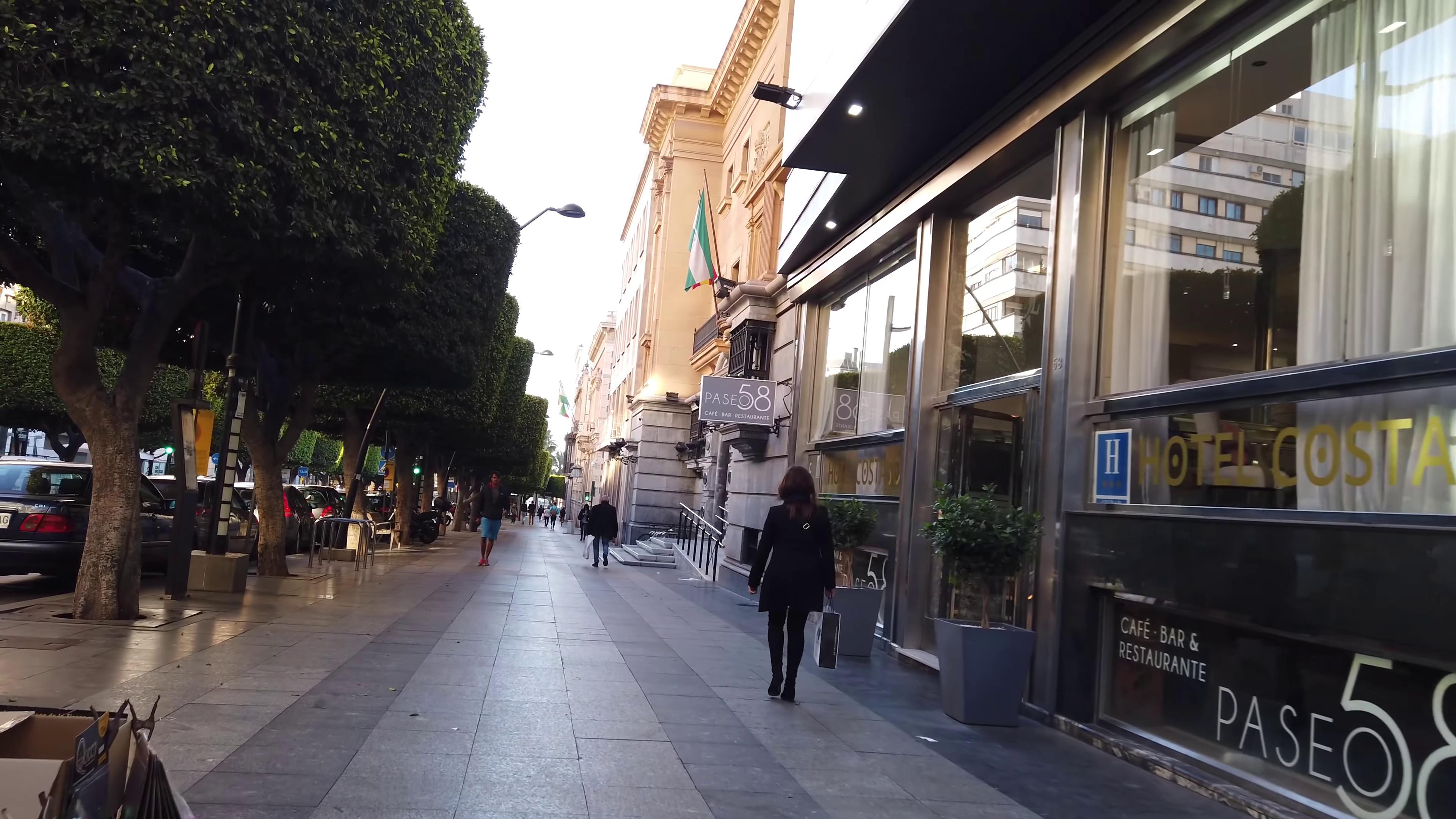}
        \caption{42.57}
    \end{subfigure}
    \begin{subfigure}[b]{0.17\textwidth}
        \includegraphics[width=\textwidth]{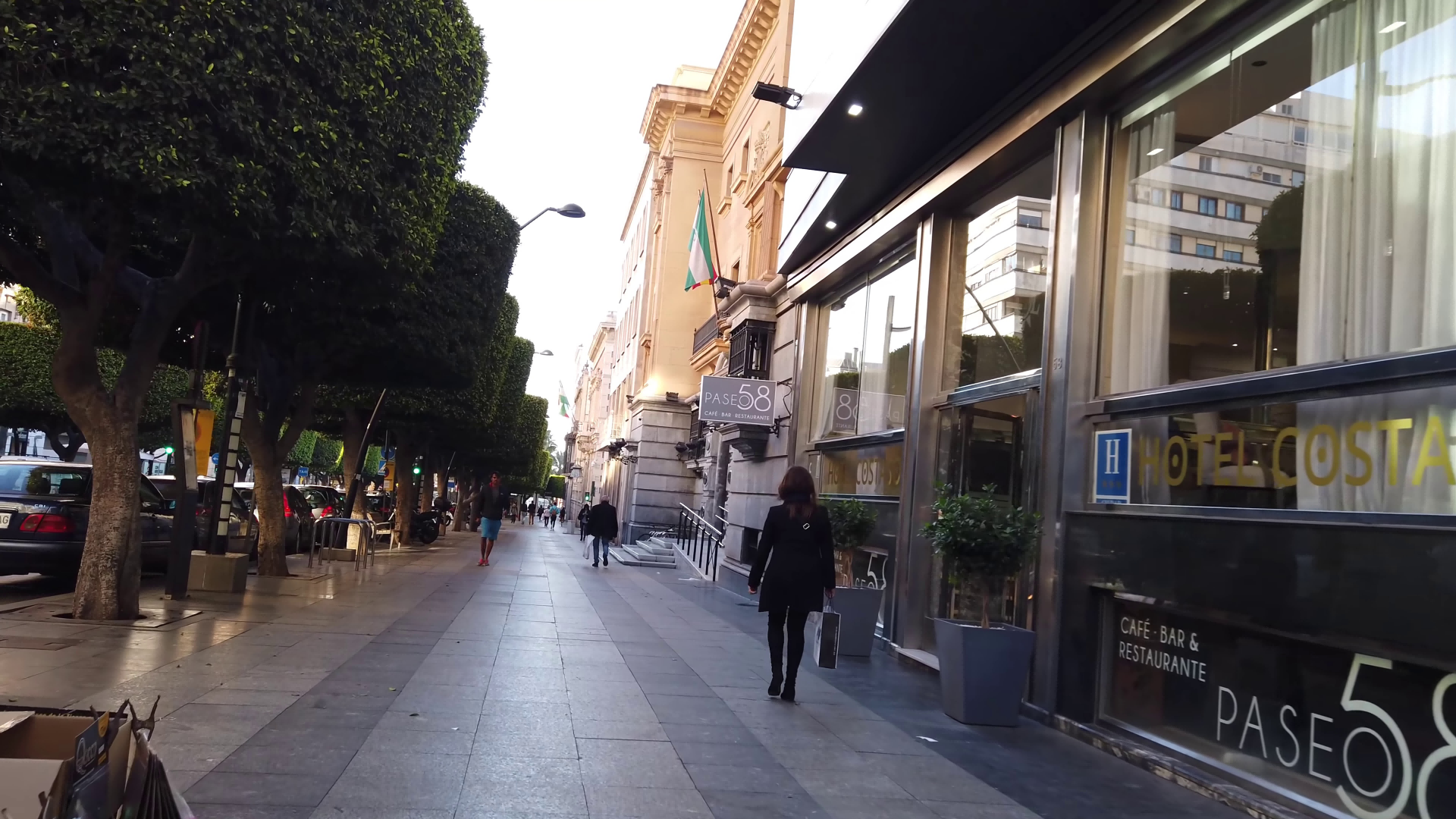}
        \caption{$+\infty$}
    \end{subfigure}}
\scalebox{0.56}{    
    \begin{subfigure}[b]{0.17\textwidth}
        \includegraphics[width=\textwidth]{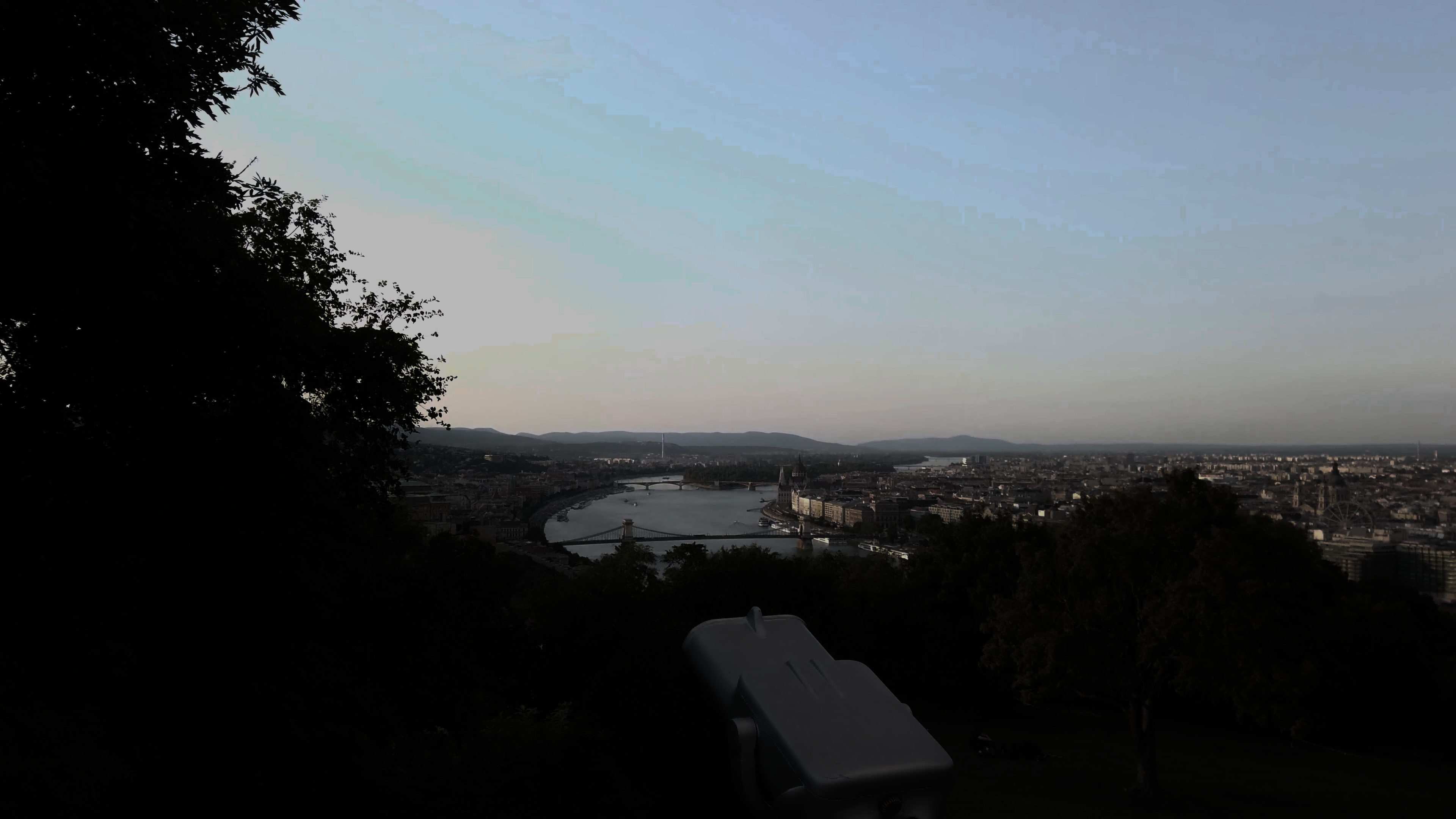}
        \caption{Input\\~}
    \end{subfigure}
    \begin{subfigure}[b]{0.17\textwidth}
        \includegraphics[width=\textwidth]{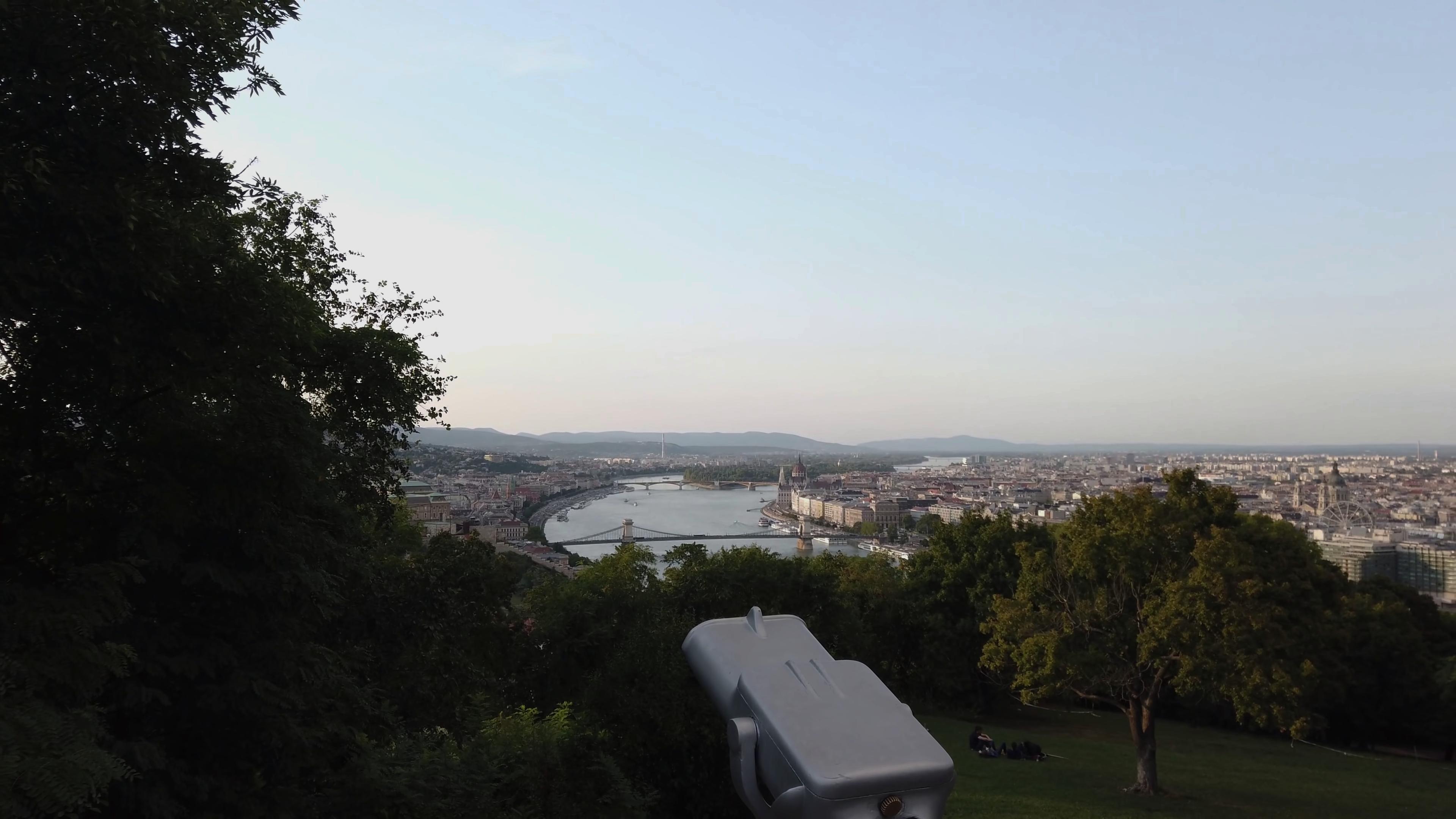}
        \caption{ZDCE++\\28.66}
    \end{subfigure}
    \begin{subfigure}[b]{0.17\textwidth}
        \includegraphics[width=\textwidth]{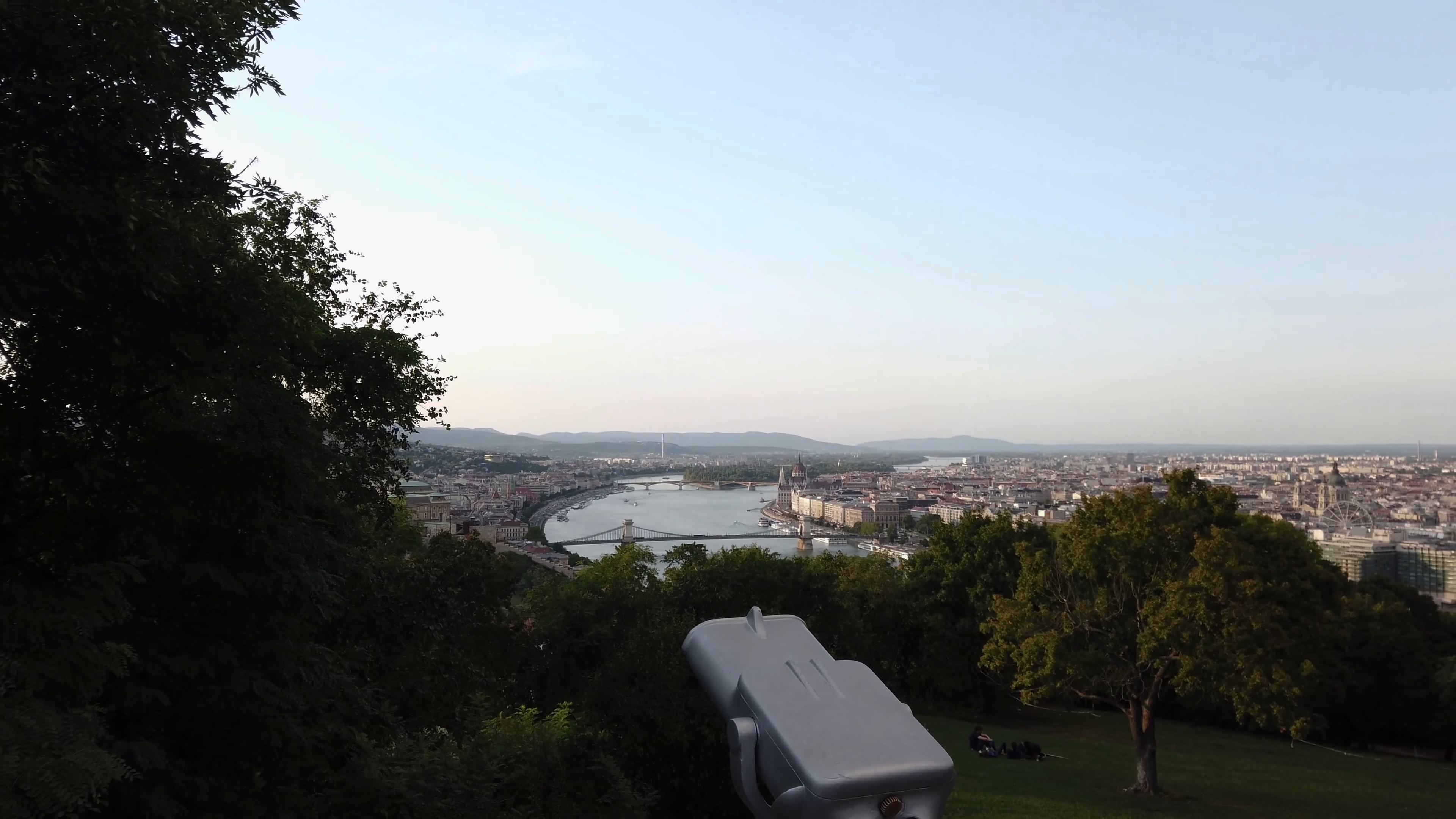}
        \caption{Uformer\\29.68}
    \end{subfigure}
    \begin{subfigure}[b]{0.17\textwidth}
        \includegraphics[width=\textwidth]{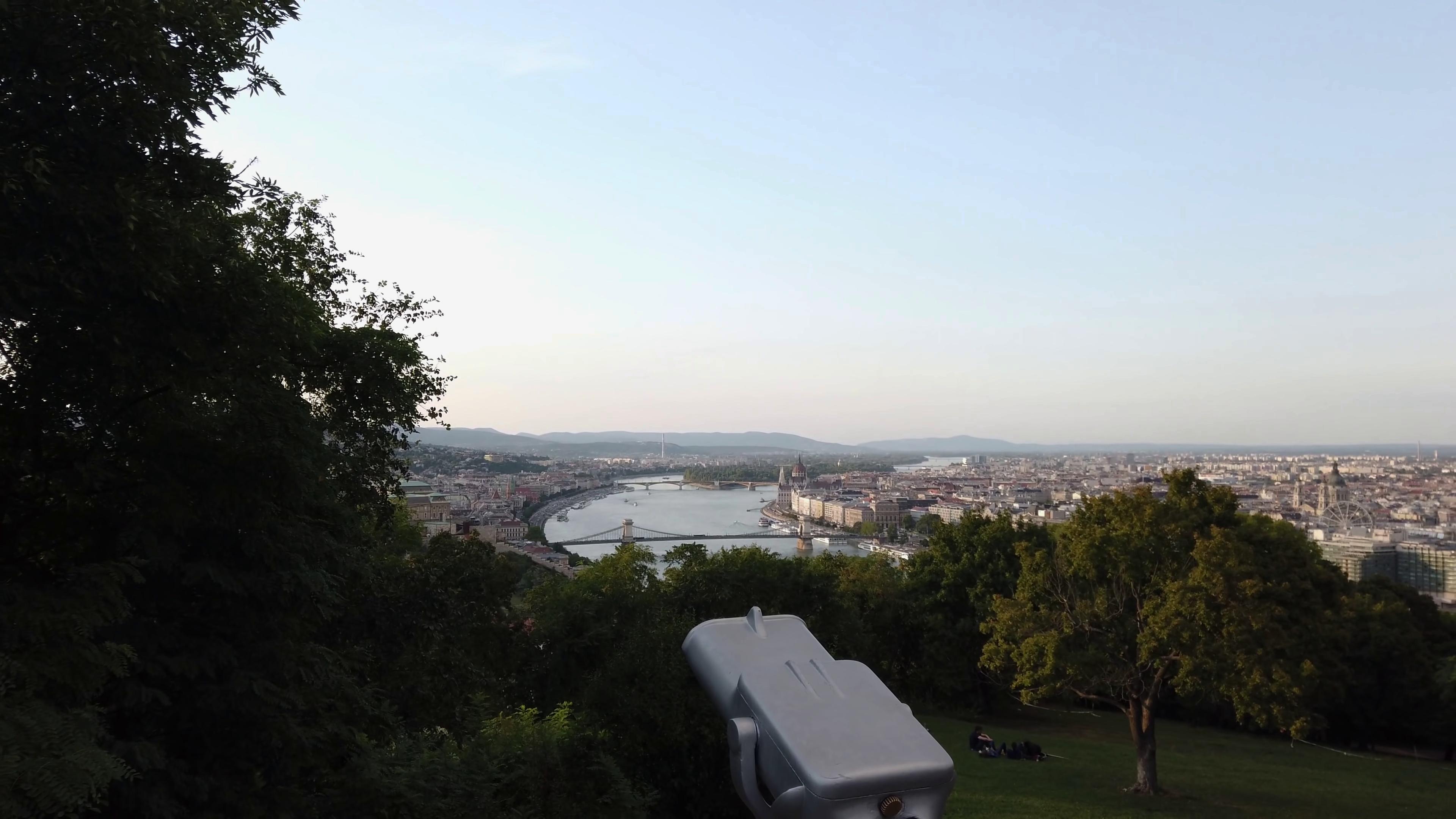}
        \caption{SCI\\29.40}
    \end{subfigure}
    \begin{subfigure}[b]{0.17\textwidth}
        \includegraphics[width=\textwidth]{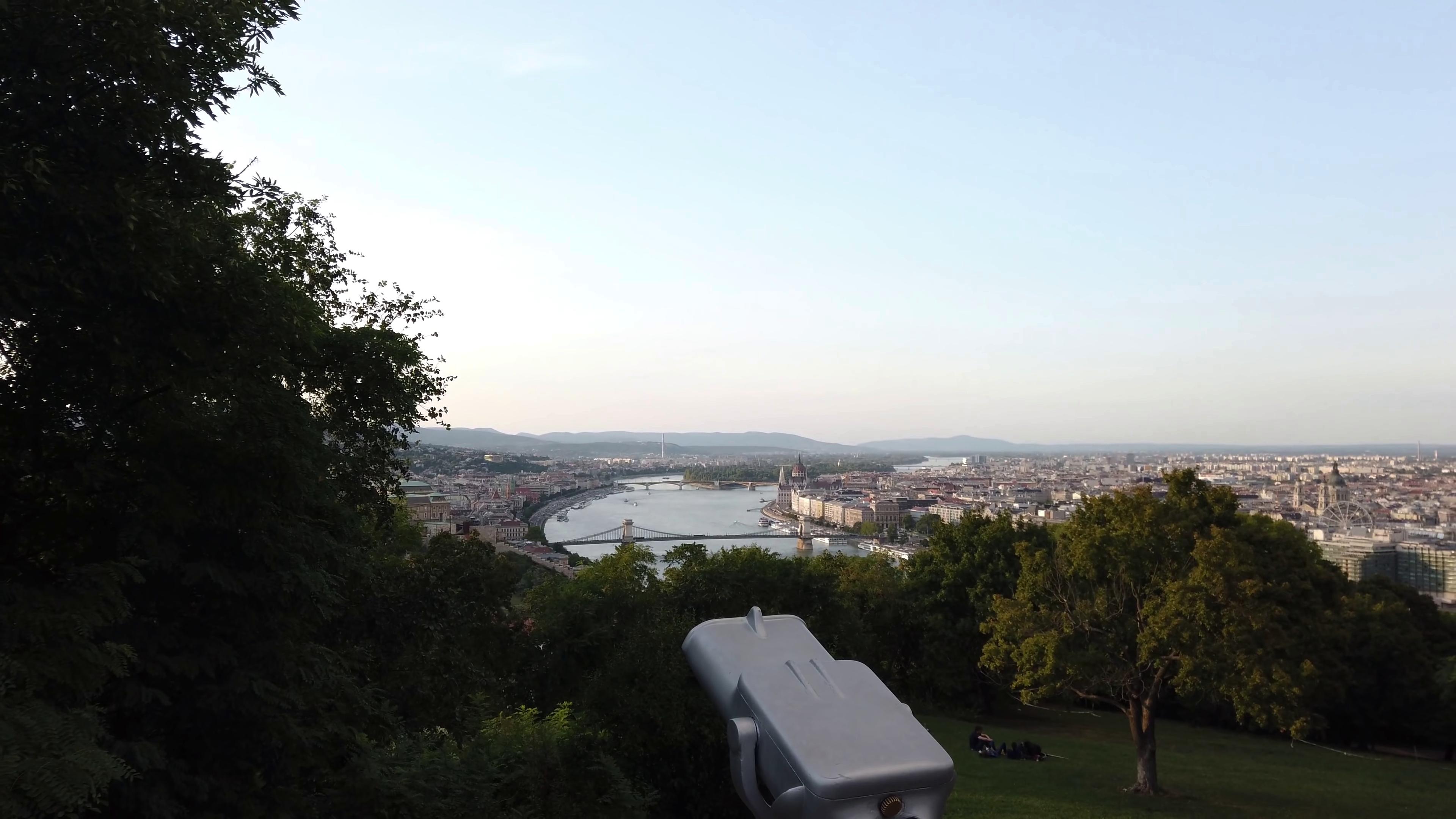}
        \caption{SNR-Aware\\32.27}
    \end{subfigure}
    \begin{subfigure}[b]{0.17\textwidth}
        \includegraphics[width=\textwidth]{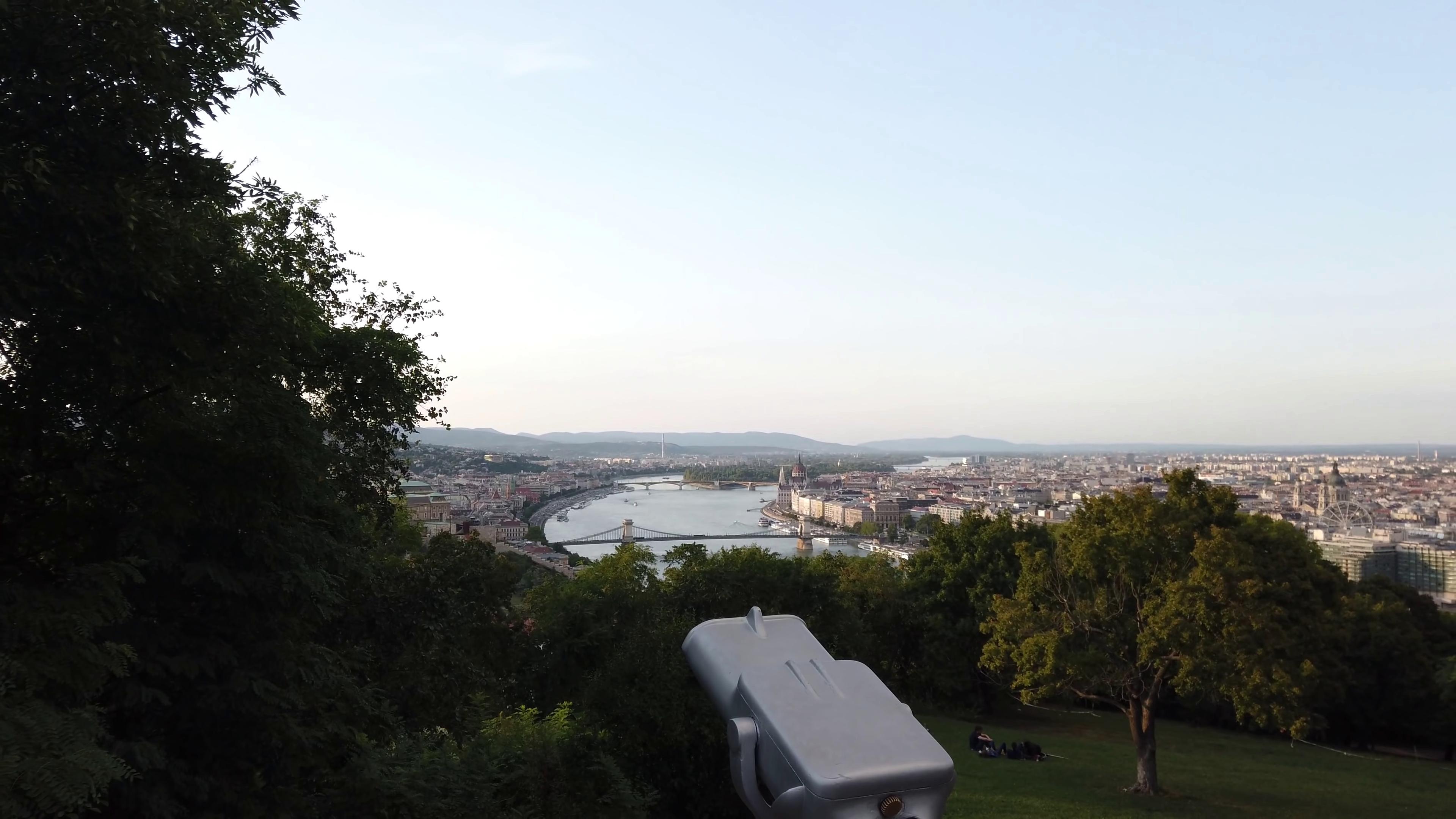}
        \caption{Restormer\\33.39}
    \end{subfigure}
    \begin{subfigure}[b]{0.17\textwidth}
        \includegraphics[width=\textwidth]{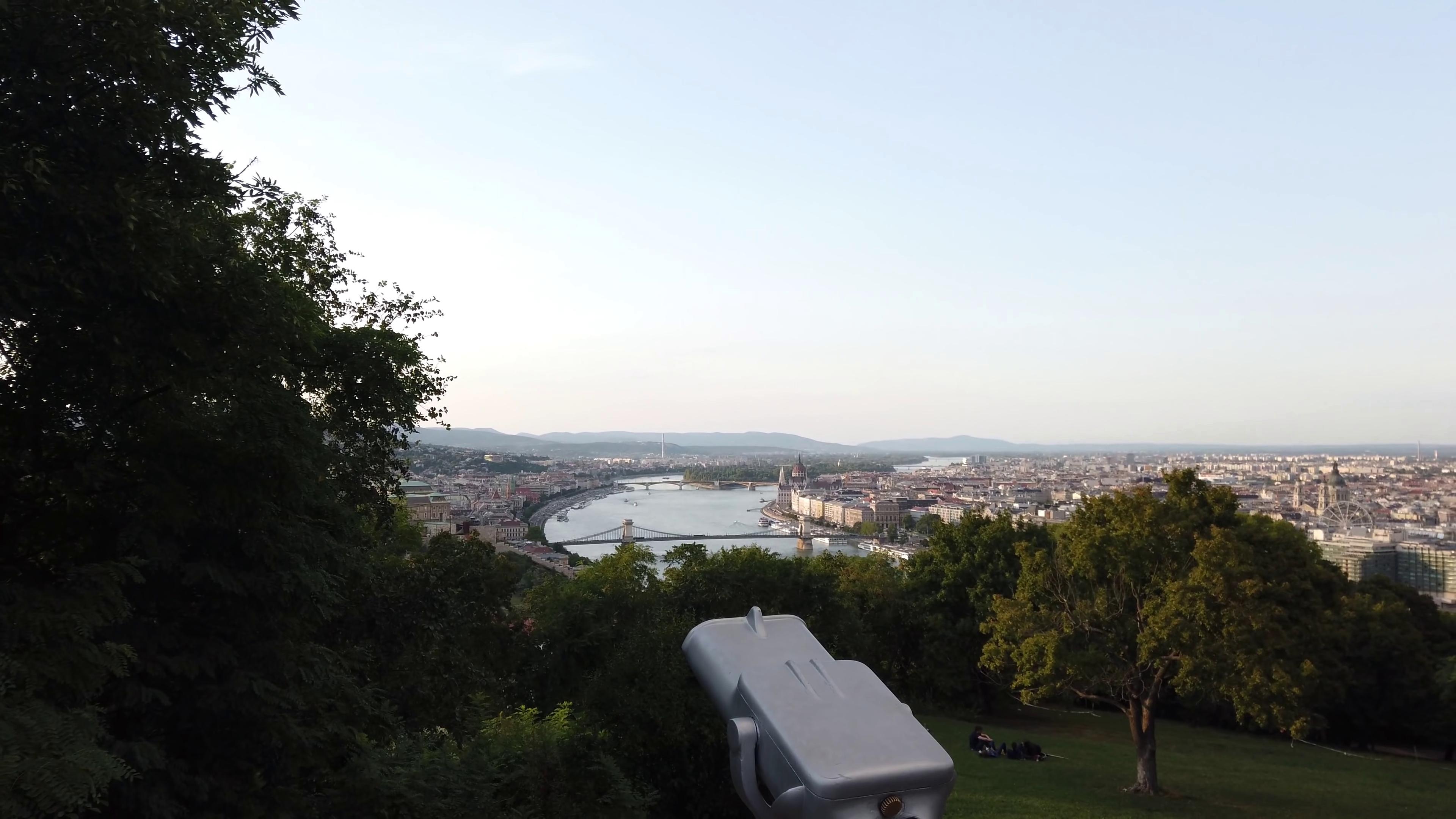}
        \caption{LLFormer\\33.99}
    \end{subfigure}
    \begin{subfigure}[b]{0.17\textwidth}
        \includegraphics[width=\textwidth]{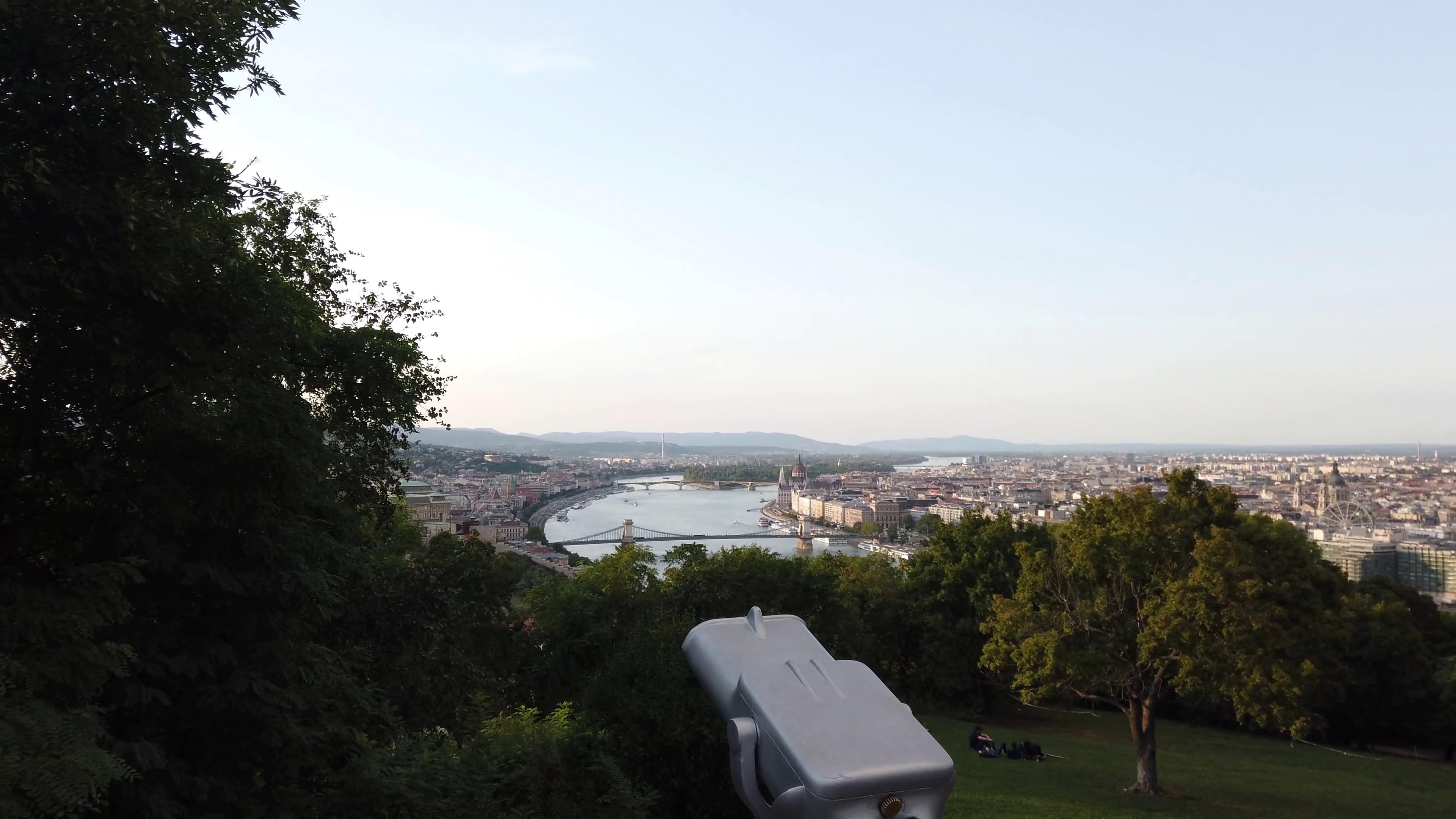}
        \caption{UHDFormer\\34.90}
    \end{subfigure}
    \begin{subfigure}[b]{0.17\textwidth}
        \includegraphics[width=\textwidth]{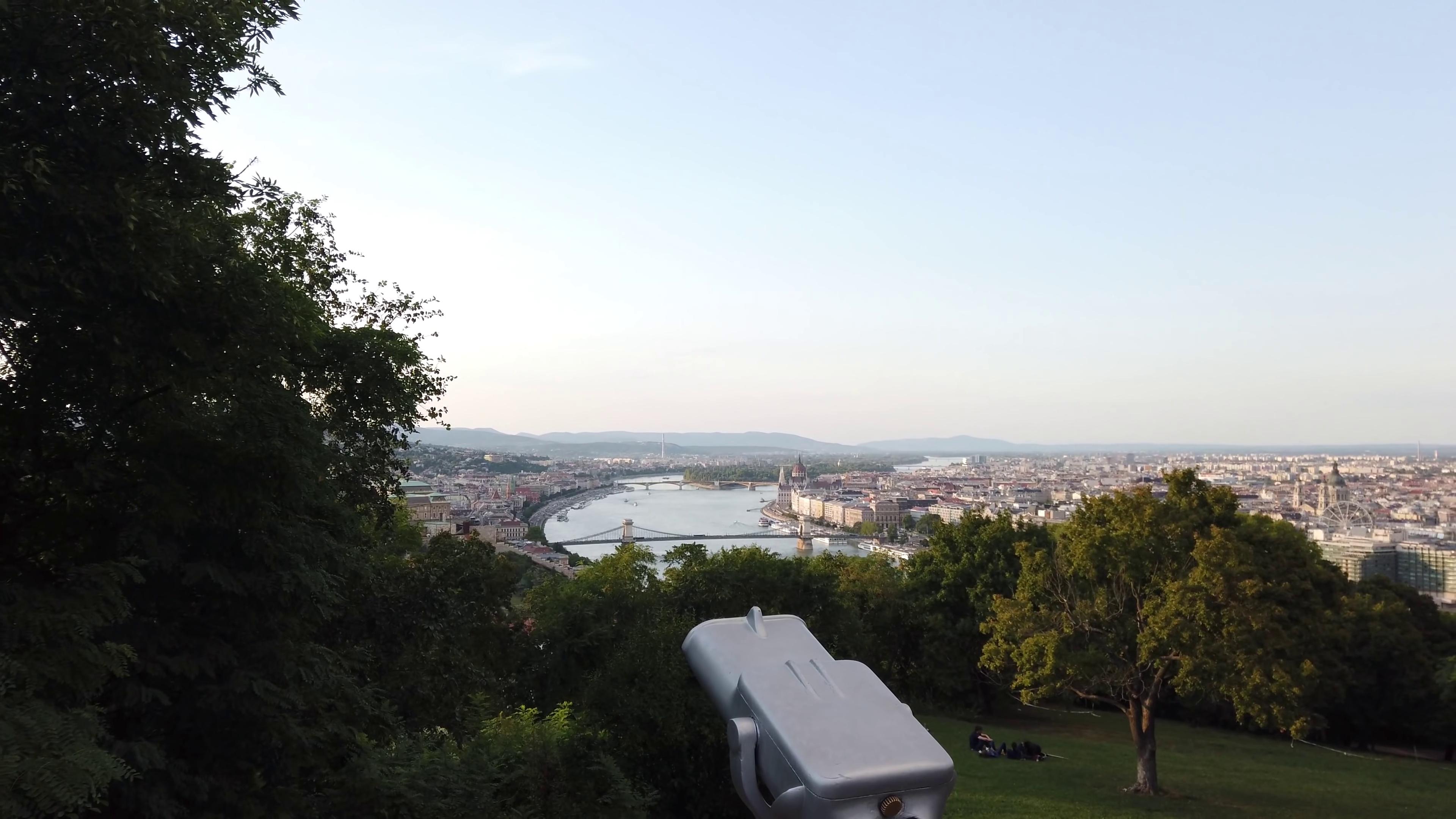}
        \caption{Ours\\43.80}
    \end{subfigure}
    \begin{subfigure}[b]{0.17\textwidth}
        \includegraphics[width=\textwidth]{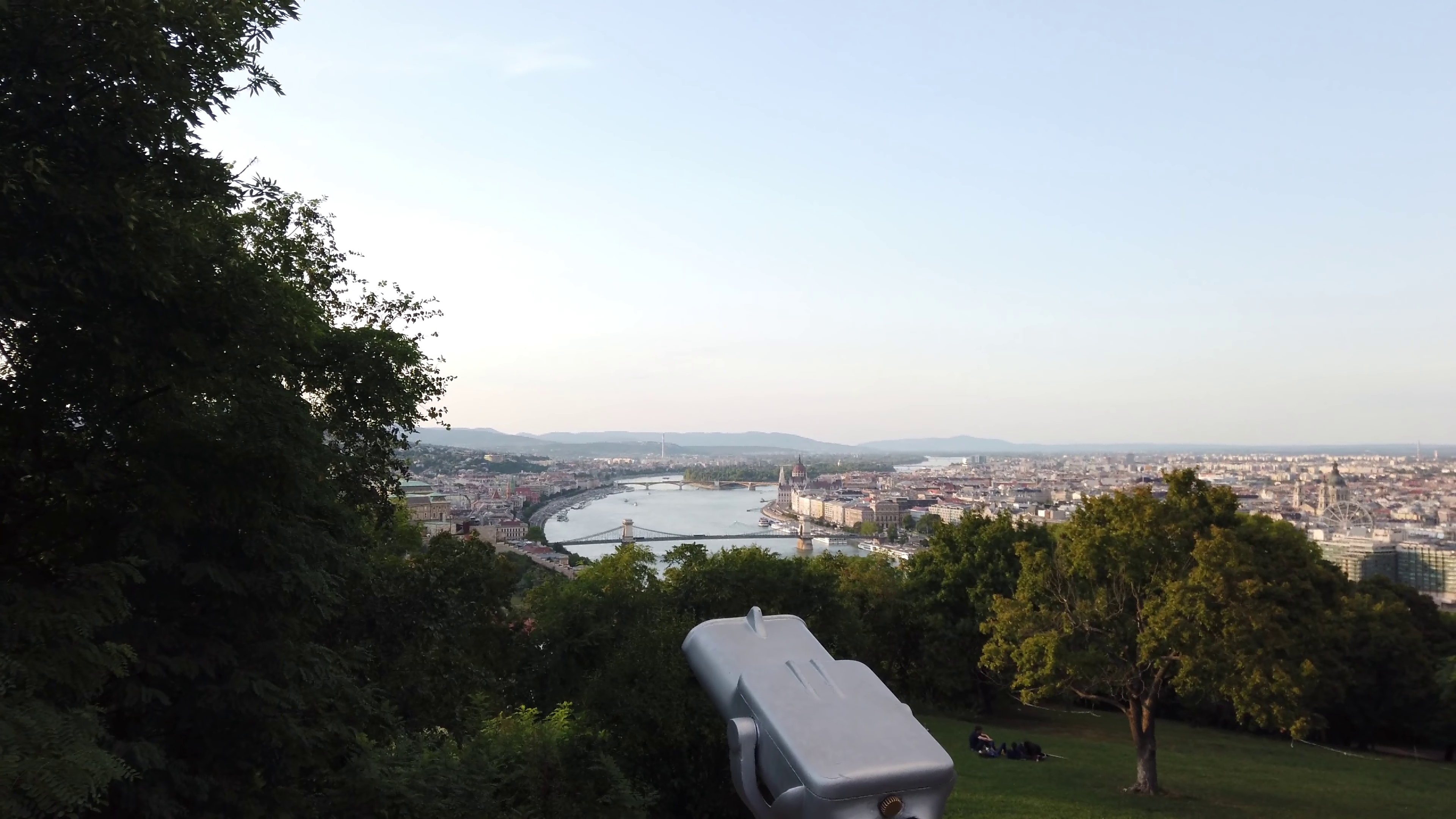}
        \caption{GT\\$+\infty$}
    \end{subfigure}}
    \vspace{-2mm}
    \caption{Comparison of methods on UHD-LOL4K dataset. All comparison methods are retrained on the UHDFormer synthesised dataset.}
    \vspace{-4mm}
    \label{4k}
\end{figure*}

\begin{figure*}[!t]
    \centering
    \scalebox{0.56}{
    % 第一行图片
    \begin{subfigure}[b]{0.17\textwidth}
        \includegraphics[width=\textwidth]{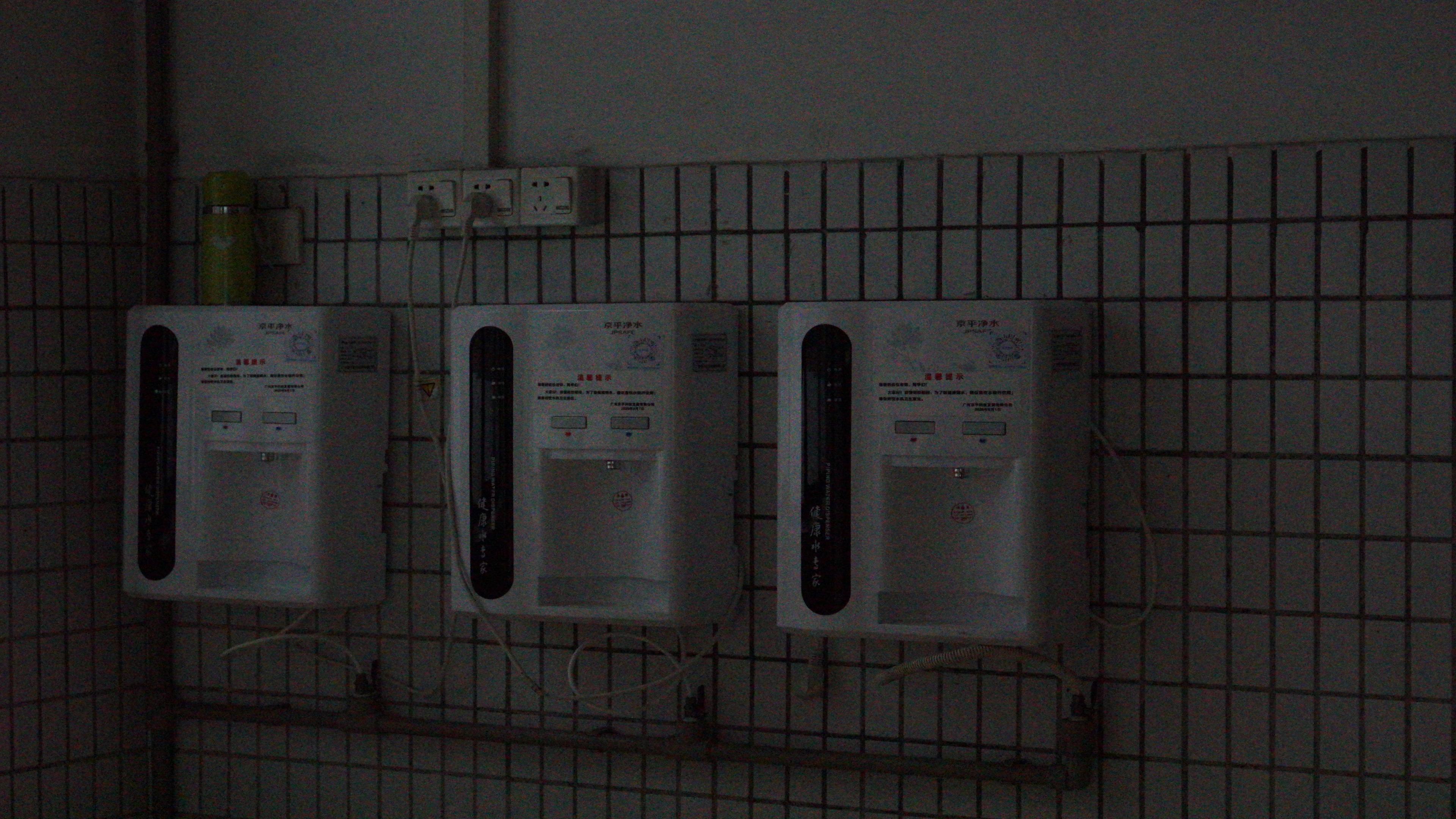}
        \caption*{PSNR}
    \end{subfigure}
    \begin{subfigure}[b]{0.17\textwidth}
        \includegraphics[width=\textwidth]{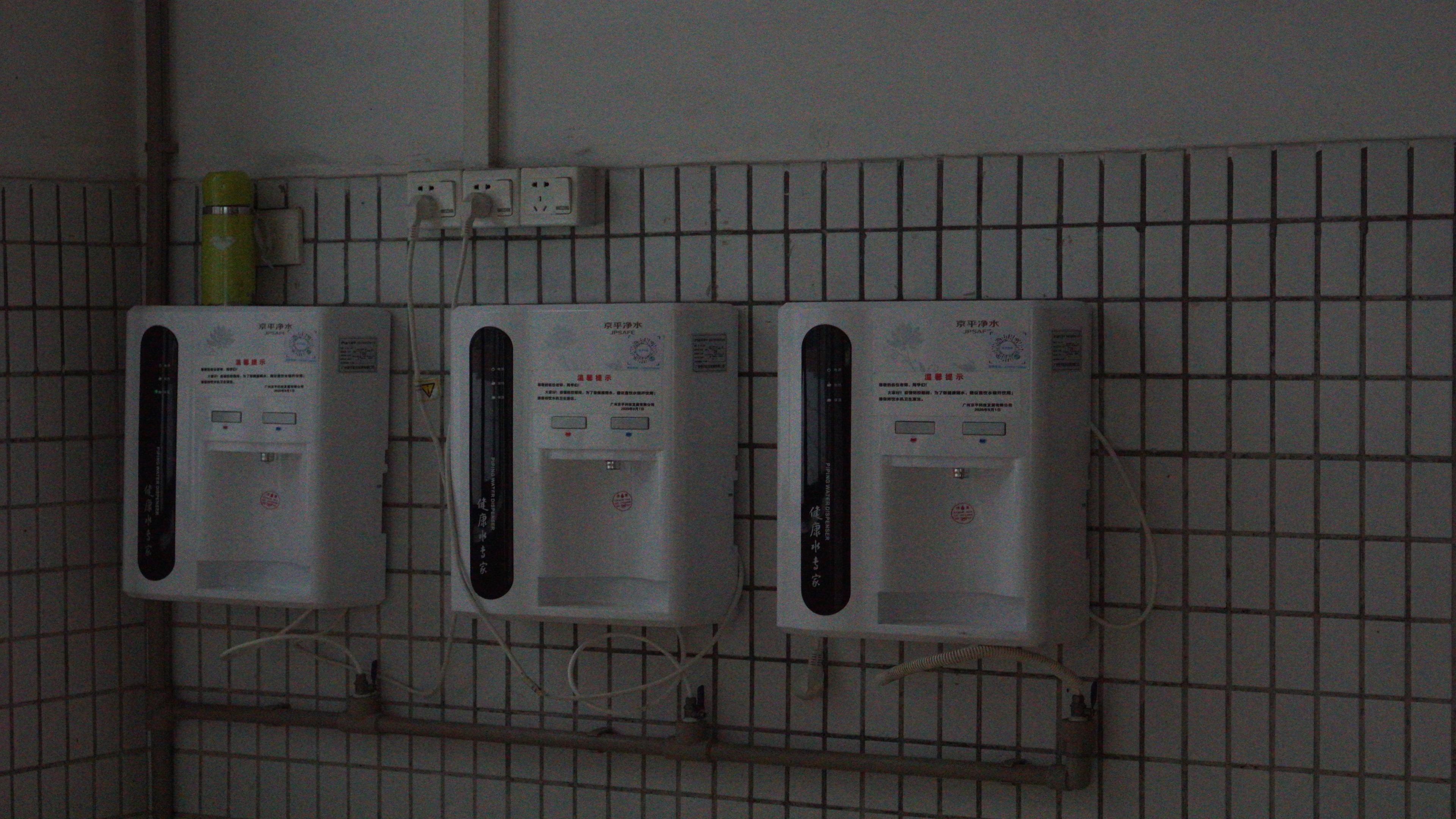}
        \caption*{20.11}
    \end{subfigure}
    \begin{subfigure}[b]{0.17\textwidth}
        \includegraphics[width=\textwidth]{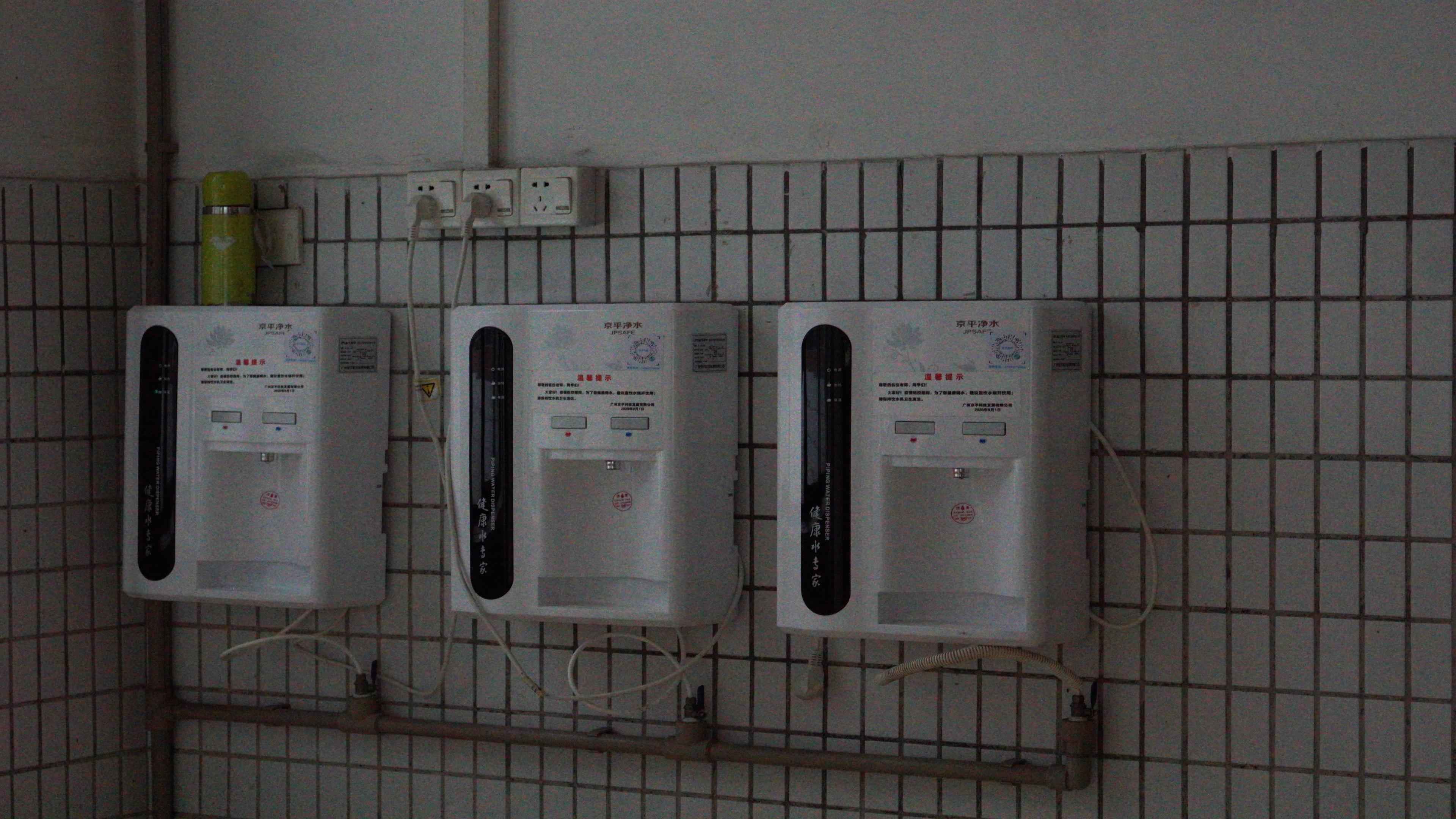}
       \caption*{21.28}
    \end{subfigure}
    \begin{subfigure}[b]{0.17\textwidth}
        \includegraphics[width=\textwidth]{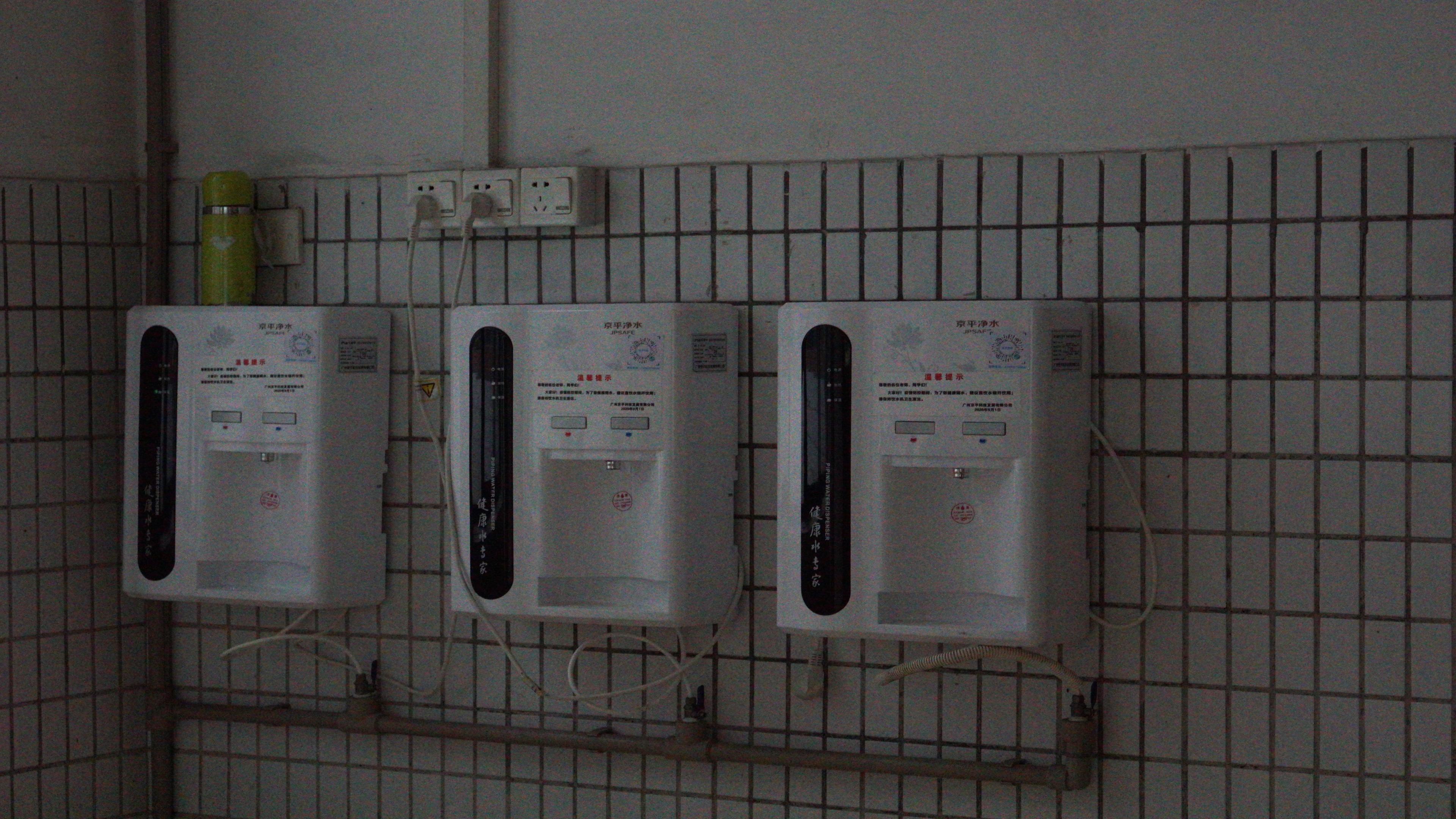}
       \caption*{21.02}
        % \caption{SCI}
    \end{subfigure}
    \begin{subfigure}[b]{0.17\textwidth}
        \includegraphics[width=\textwidth]{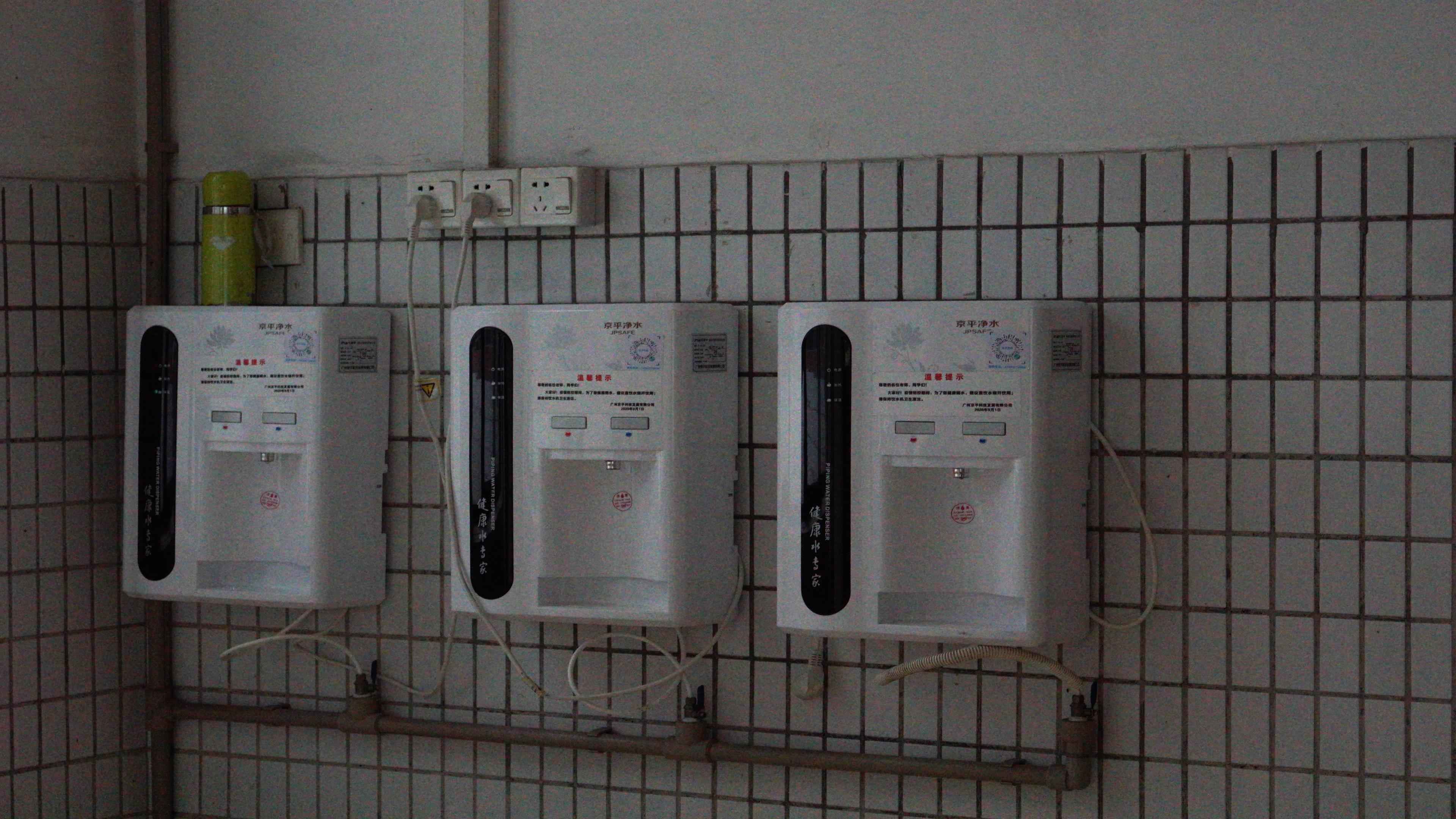}
        \caption*{21.22}
        % \caption{SNR-Aware}
    \end{subfigure}
    \begin{subfigure}[b]{0.17\textwidth}
        \includegraphics[width=\textwidth]{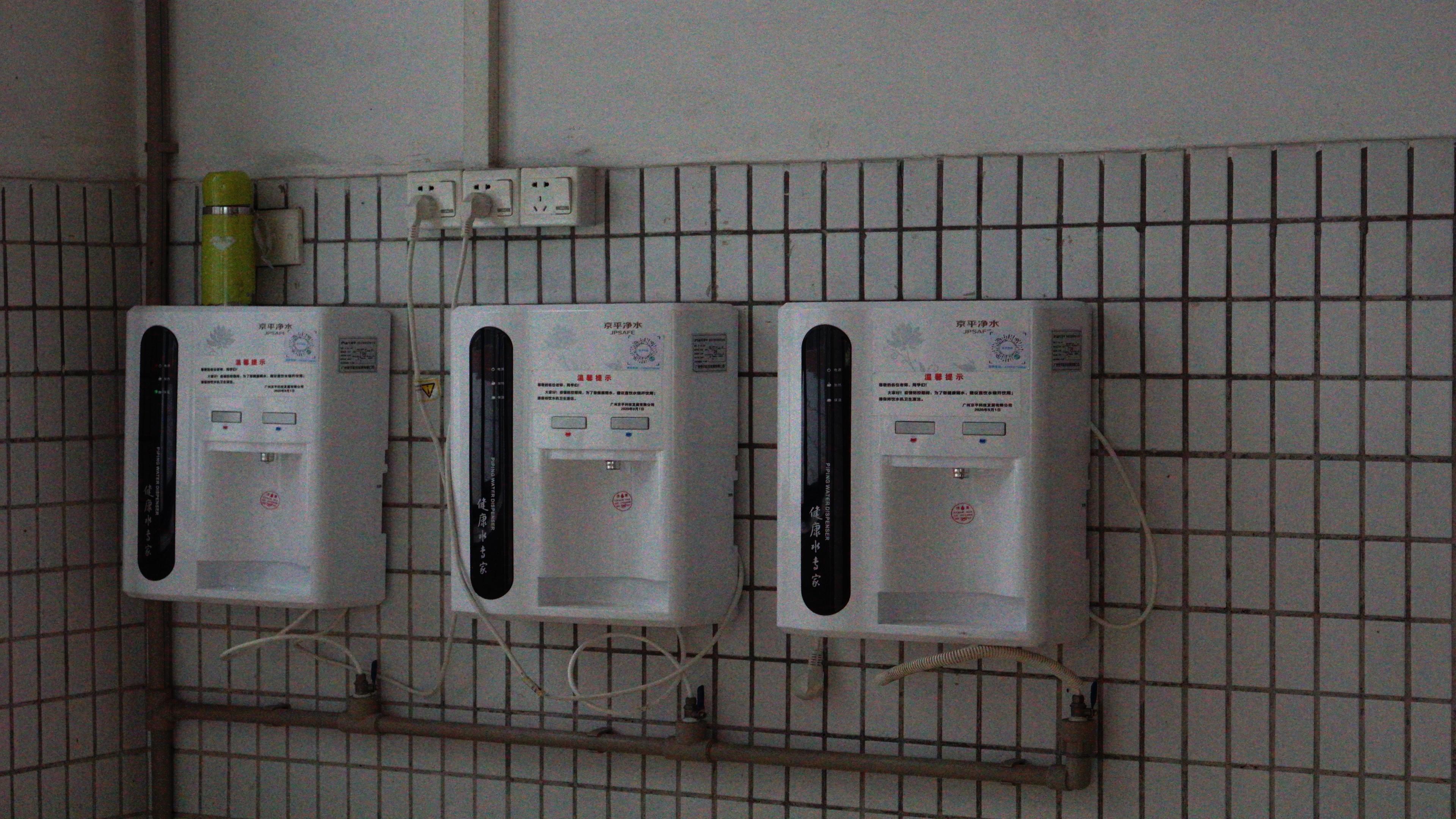}
        \caption*{22.07}
    \end{subfigure}
    \begin{subfigure}[b]{0.17\textwidth}
        \includegraphics[width=\textwidth]{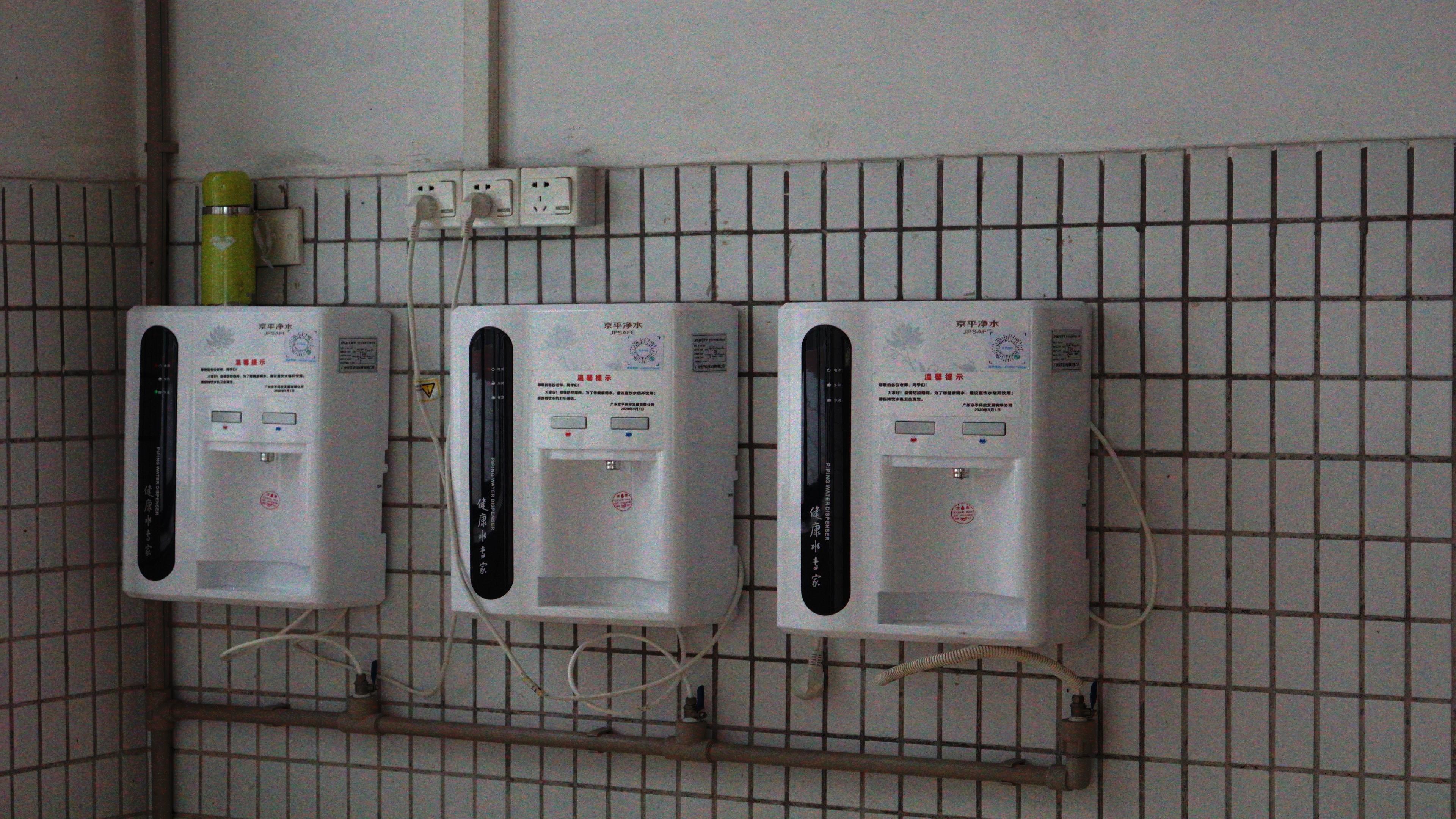}
        \caption*{23.01}
        % \caption{LLFormer}
    \end{subfigure}
    \begin{subfigure}[b]{0.17\textwidth}
        \includegraphics[width=\textwidth]{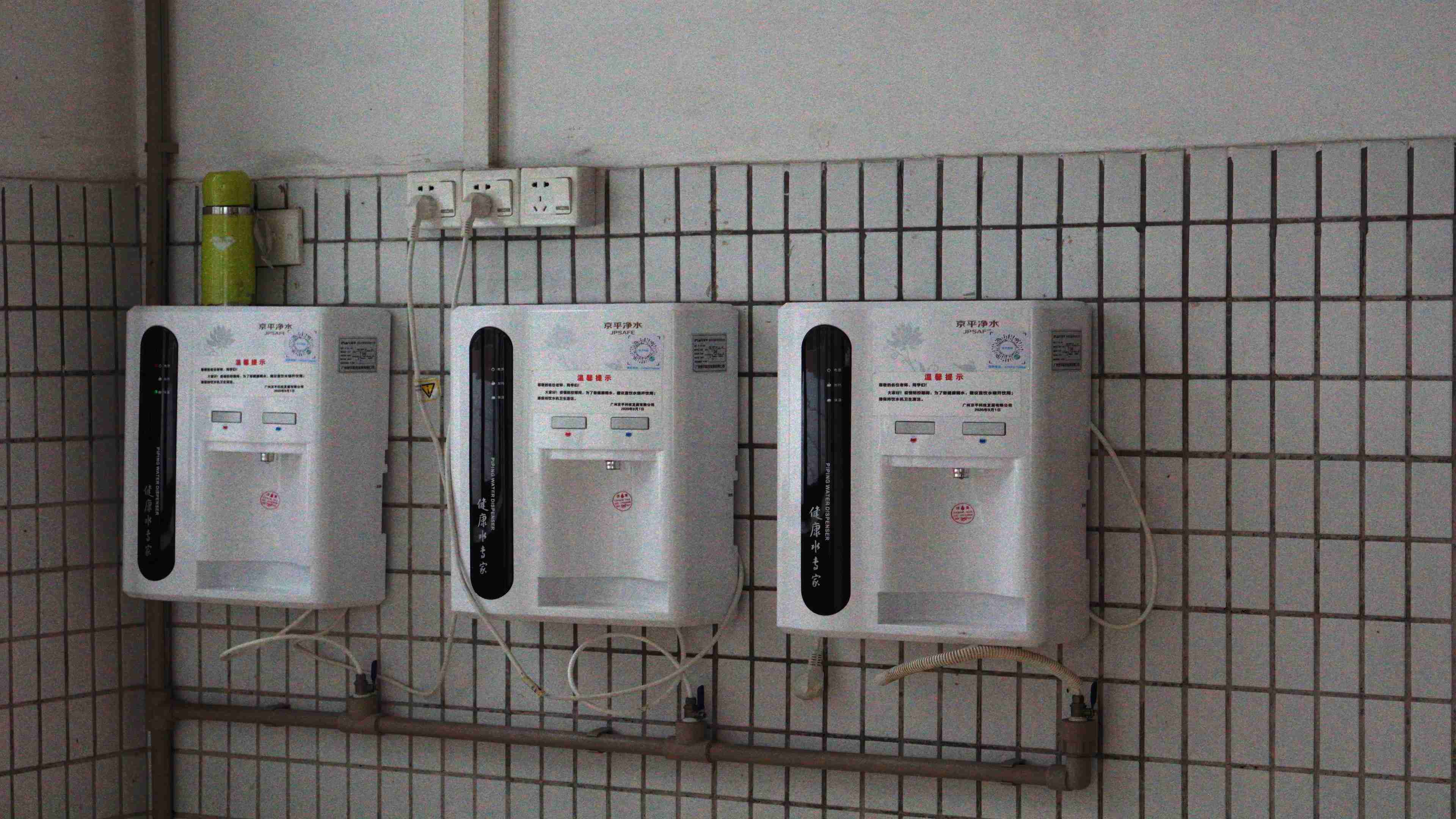}
        \caption*{24.11}
    \end{subfigure}
    
    \begin{subfigure}[b]{0.17\textwidth}
        \includegraphics[width=\textwidth]
        {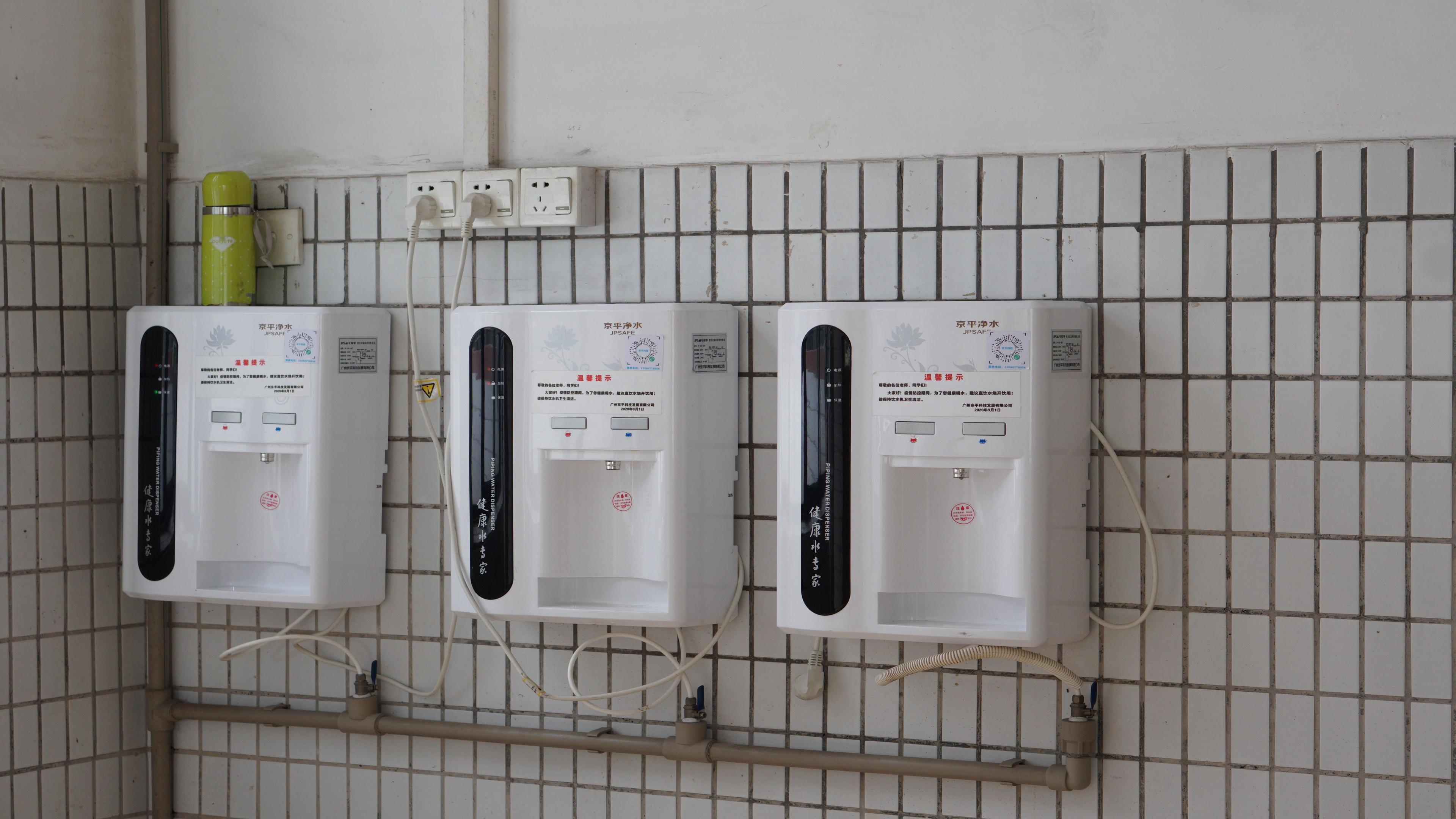}   
        \caption*{24.58}
    \end{subfigure}
    \begin{subfigure}[b]{0.17\textwidth}
        \includegraphics[width=\textwidth]
        {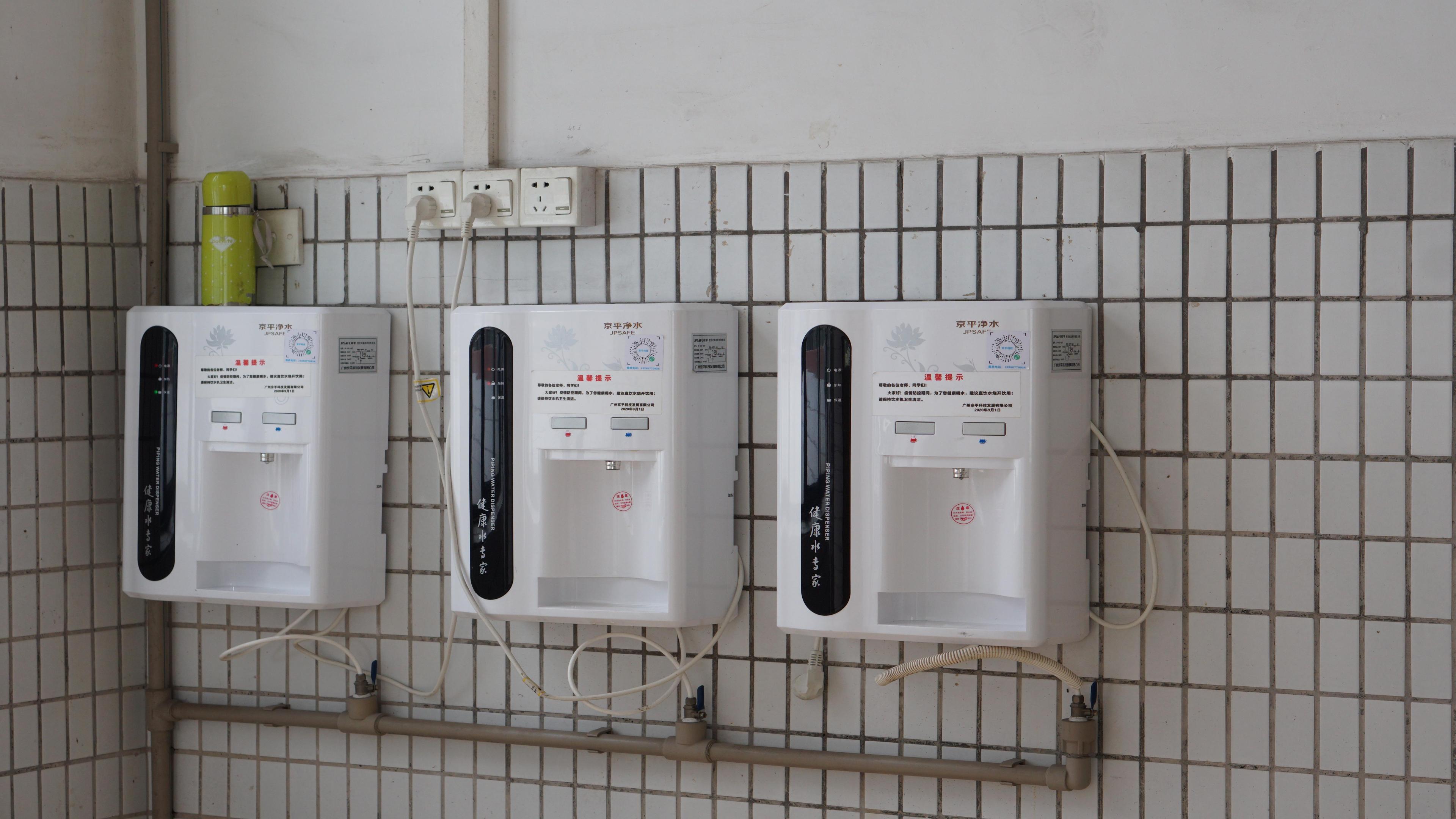}
        \caption*{$+\infty$}
        % \caption{GT}
    \end{subfigure}}
\scalebox{0.56}{
    % 第二行图片
    \begin{subfigure}[b]{0.17\textwidth}
        \includegraphics[width=\textwidth]
        {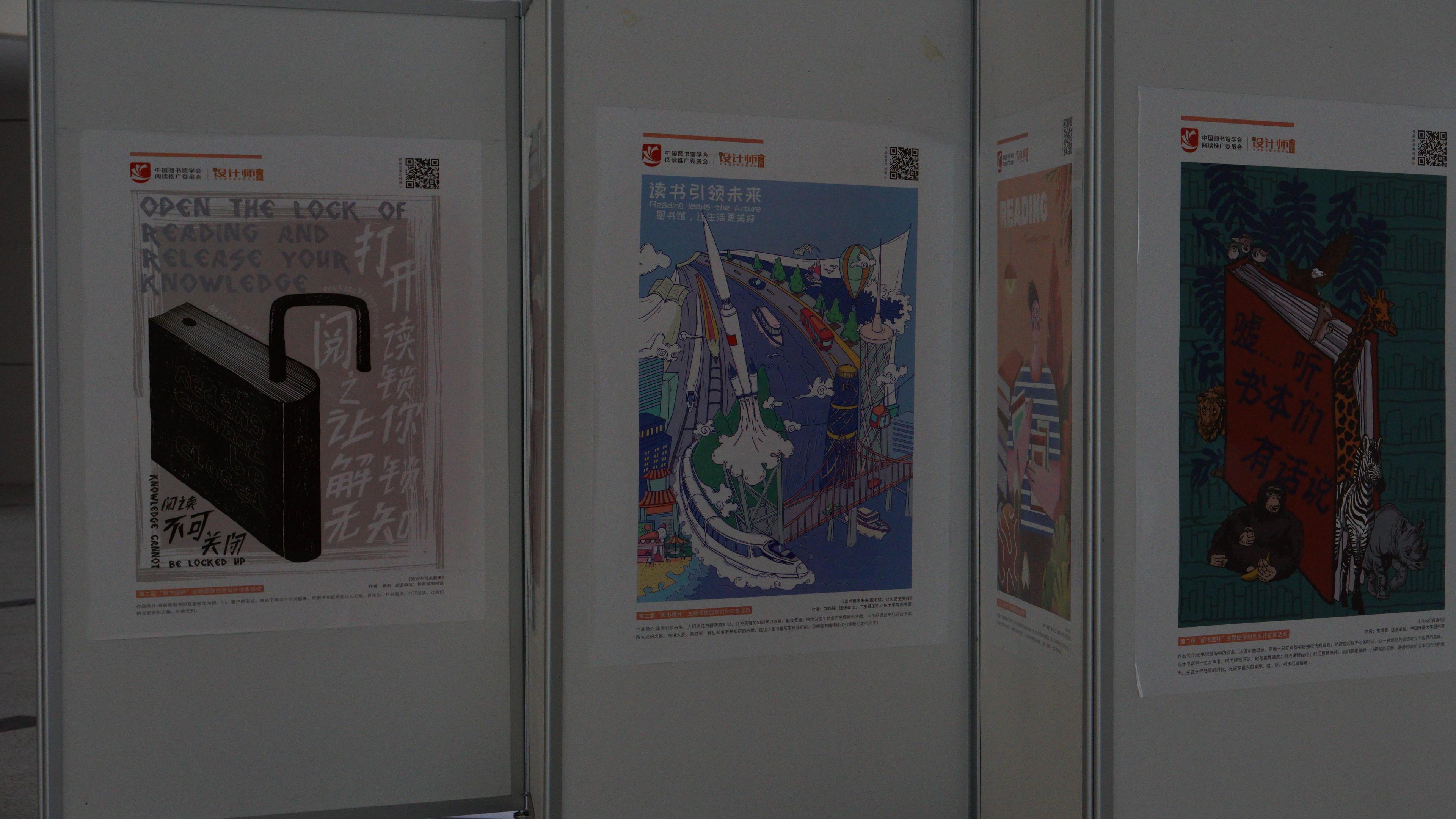}
        \caption{Input\\PSNR}
    \end{subfigure}
    \begin{subfigure}[b]{0.17\textwidth}
        \includegraphics[width=\textwidth]{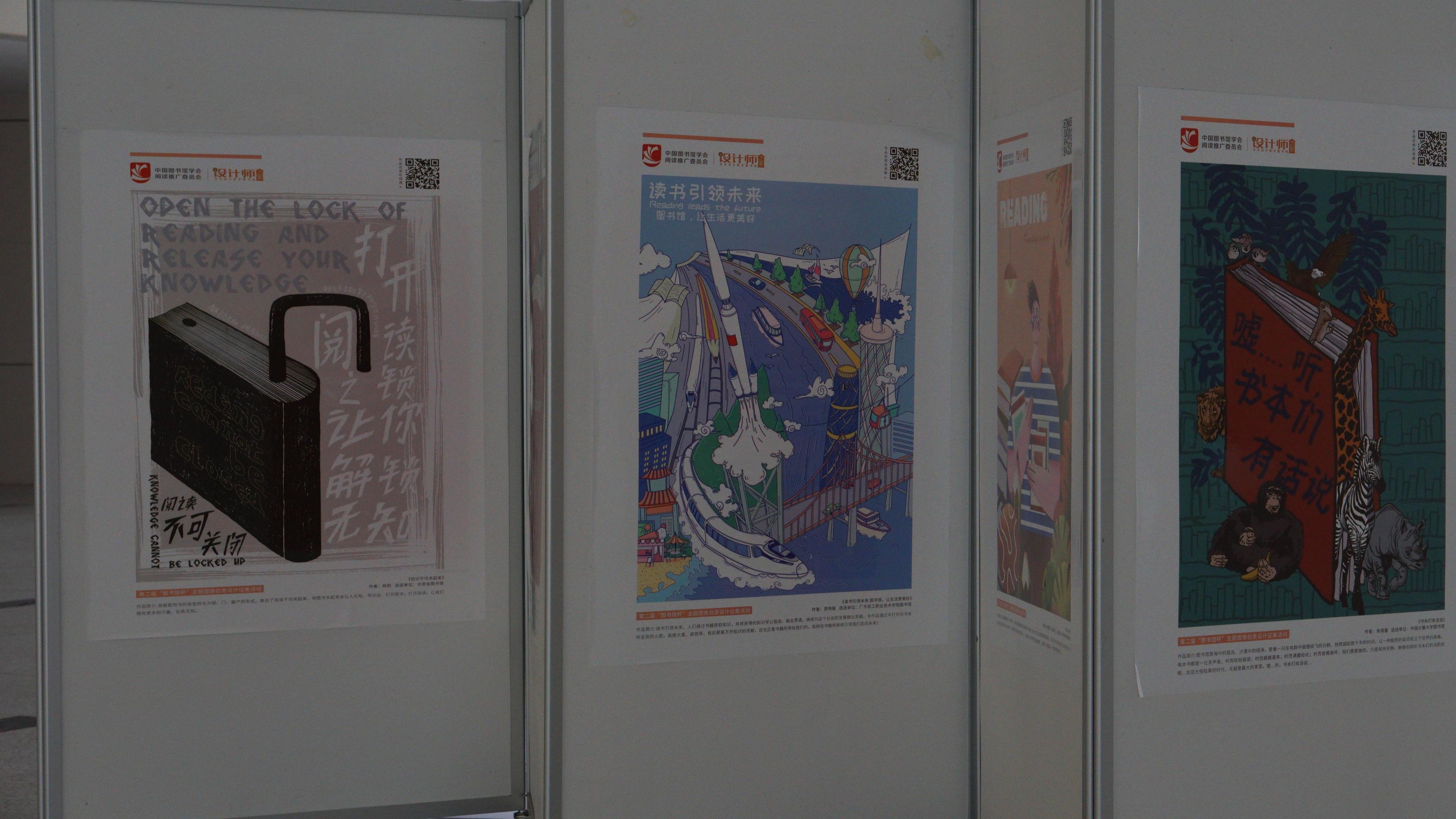}
        \caption{Z.DCE++\\19.94}
    \end{subfigure}
    \begin{subfigure}[b]{0.17\textwidth}
        \includegraphics[width=\textwidth]{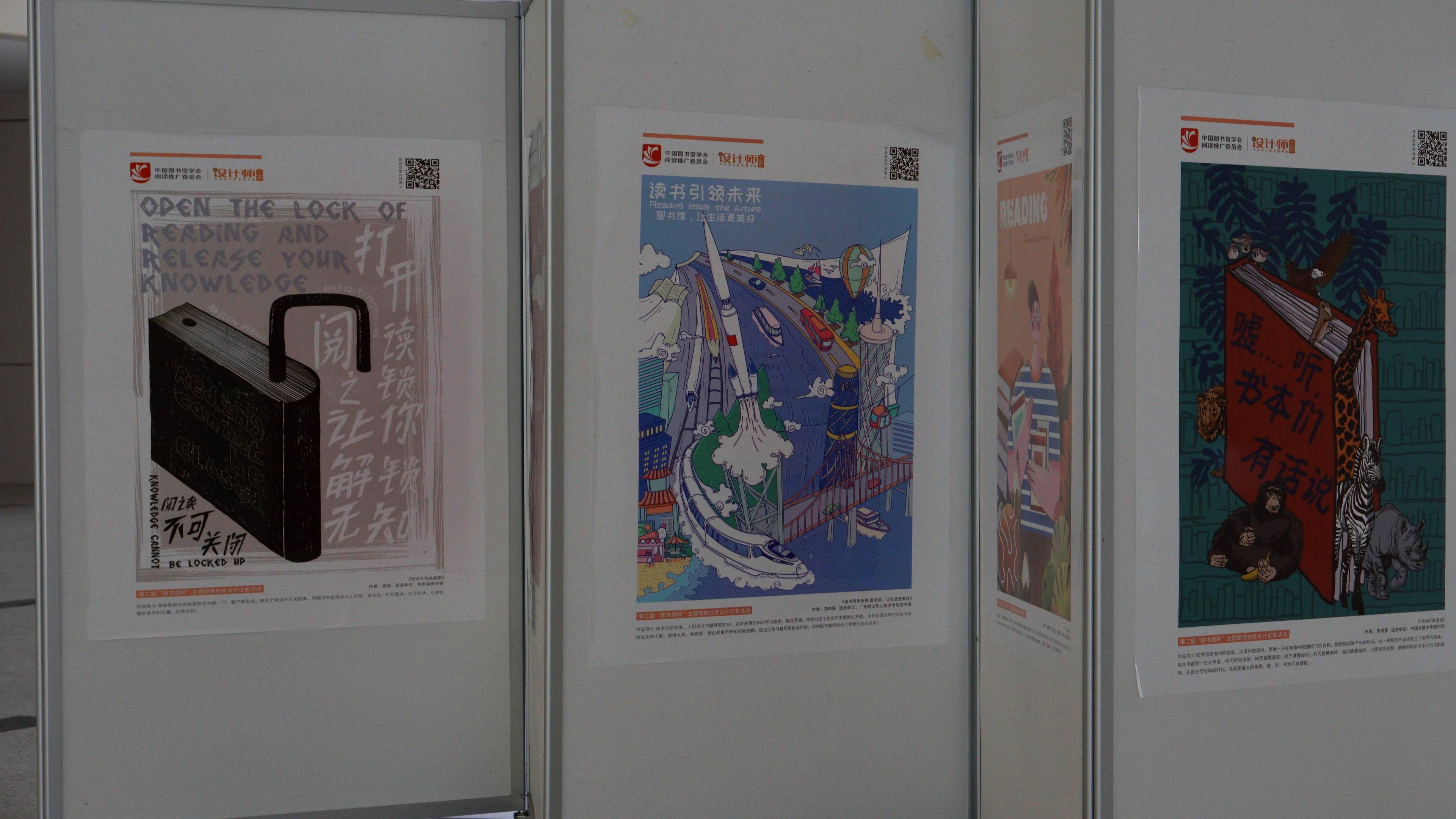}
        \caption{Uformer\\20.55}
    \end{subfigure}
    \begin{subfigure}[b]{0.17\textwidth}
        \includegraphics[width=\textwidth]{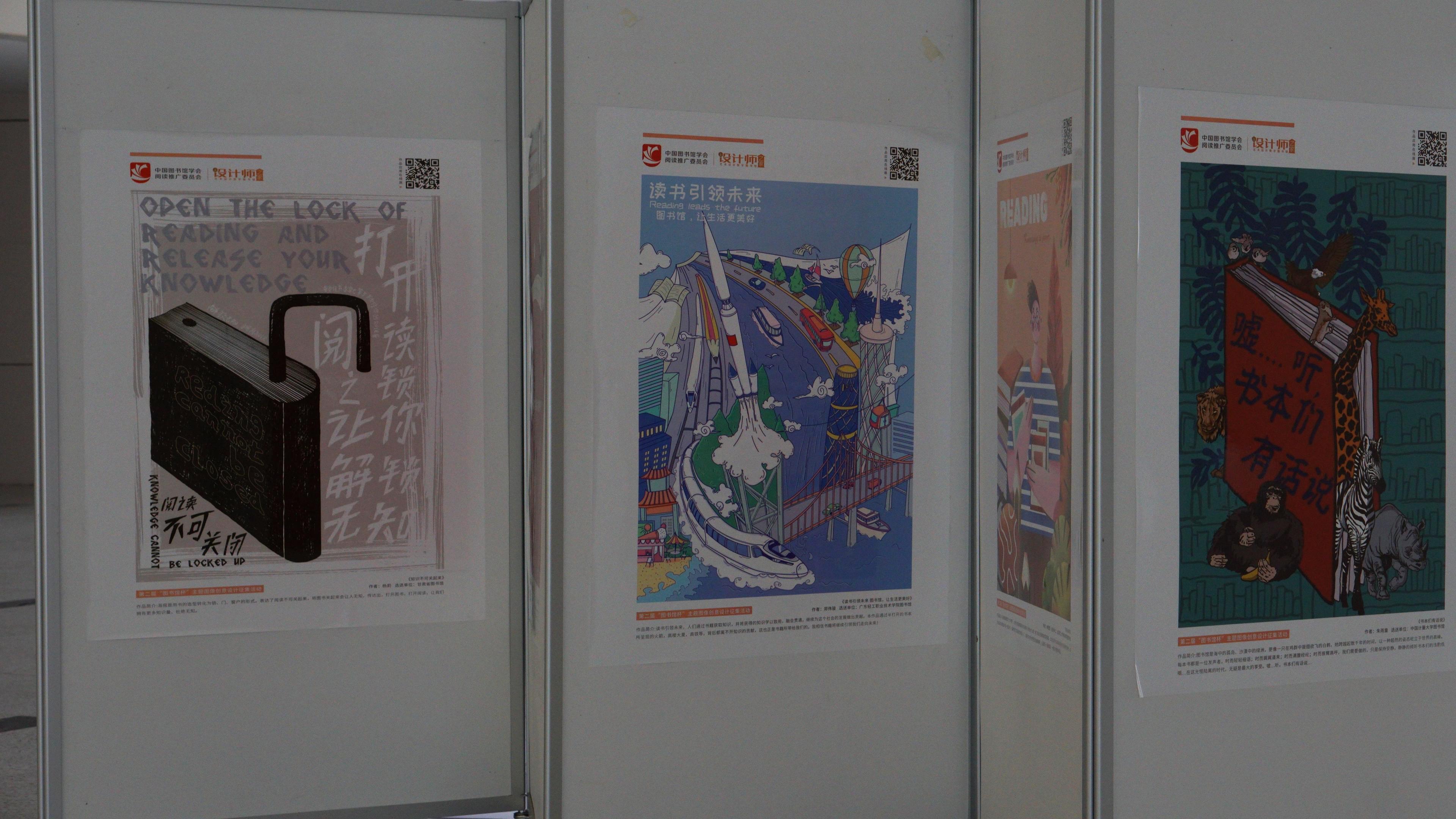}
        \caption{SCI\\21.51}
    \end{subfigure}
    \begin{subfigure}[b]{0.17\textwidth}
        \includegraphics[width=\textwidth]{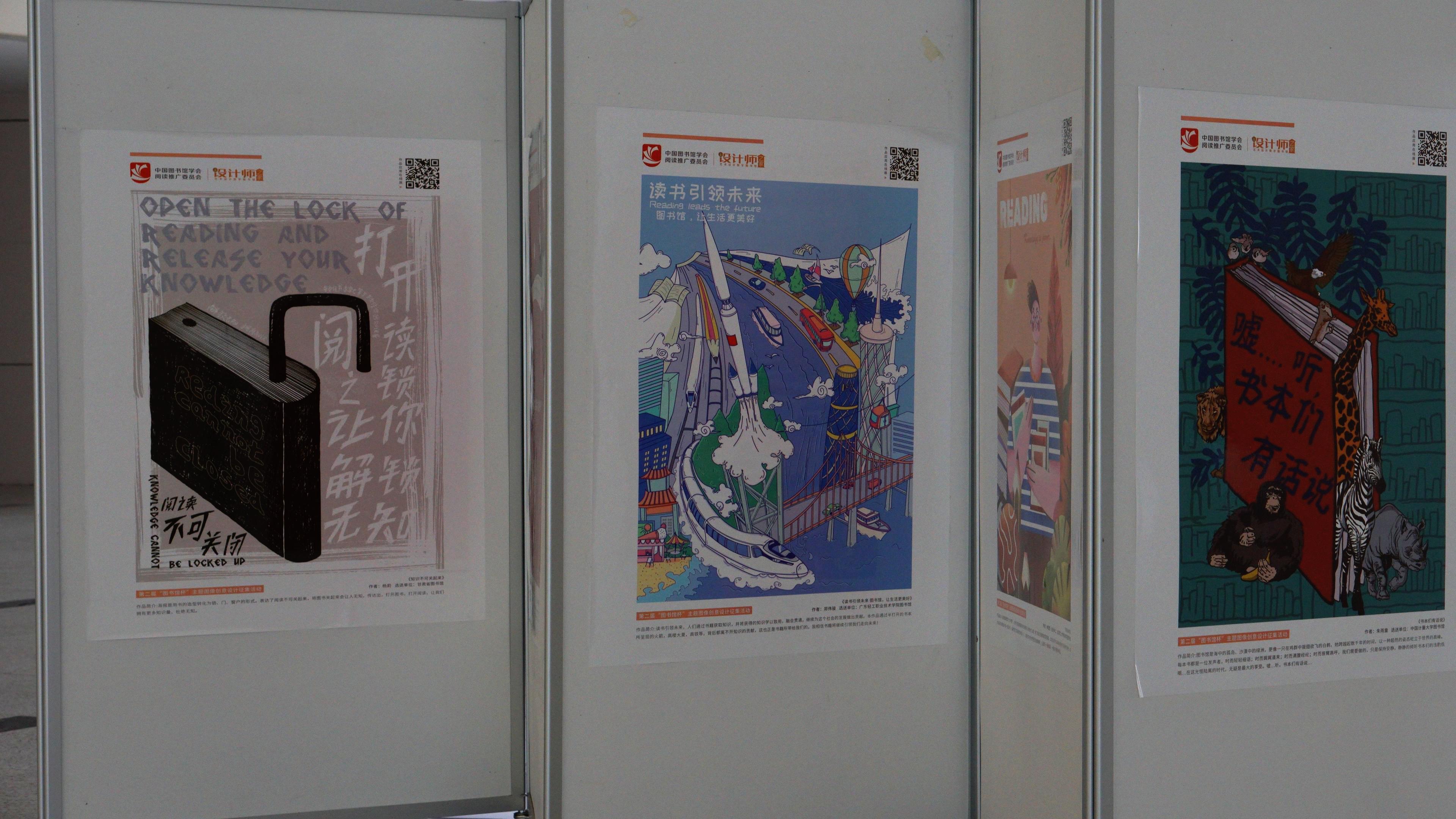}
        \caption{SNR-Aware\\21.99}
    \end{subfigure}
    \begin{subfigure}[b]{0.17\textwidth}
        \includegraphics[width=\textwidth]{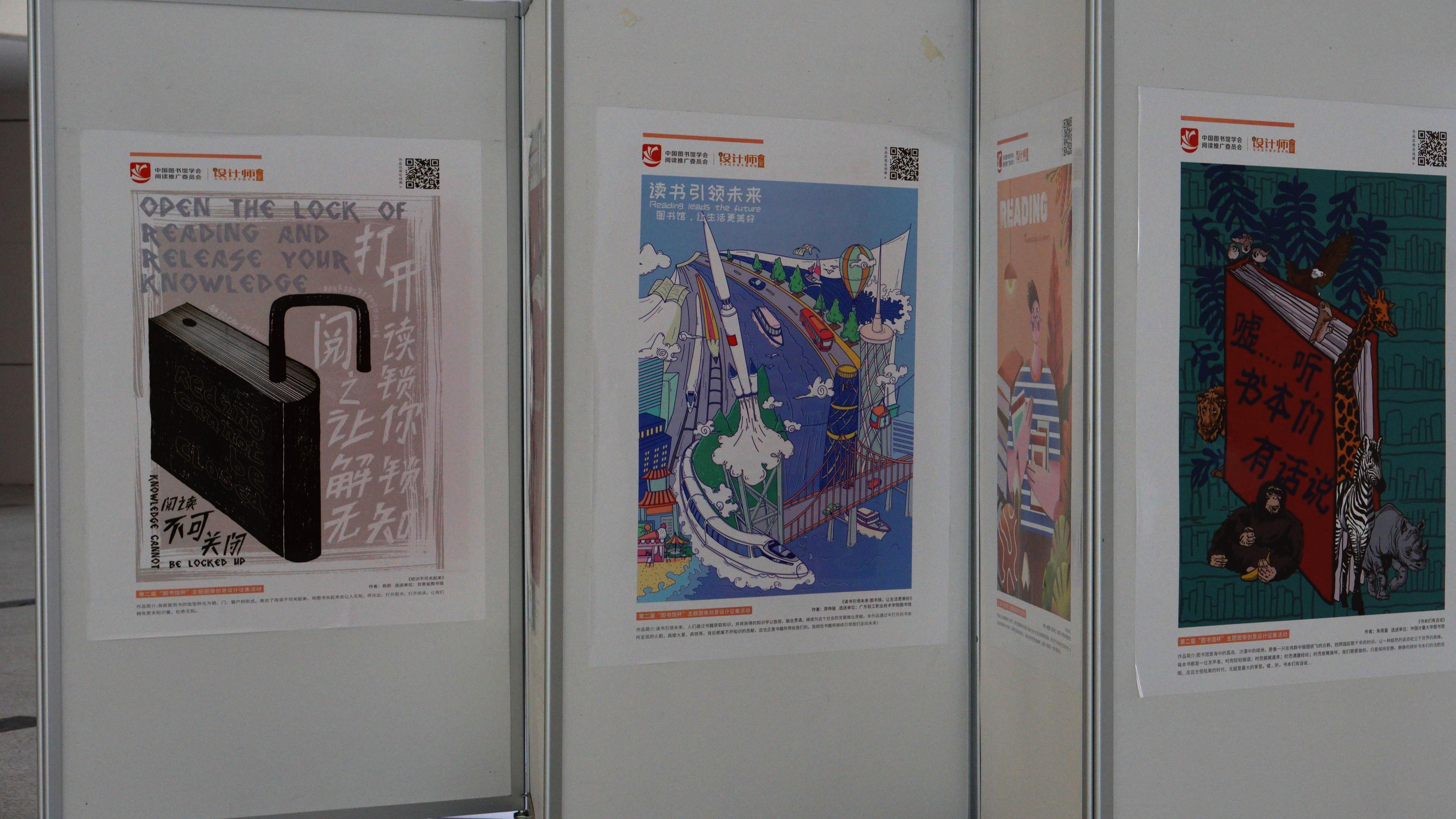}
        \caption{Restormer\\22.44}
    \end{subfigure}
    \begin{subfigure}[b]{0.17\textwidth}
        \includegraphics[width=\textwidth]{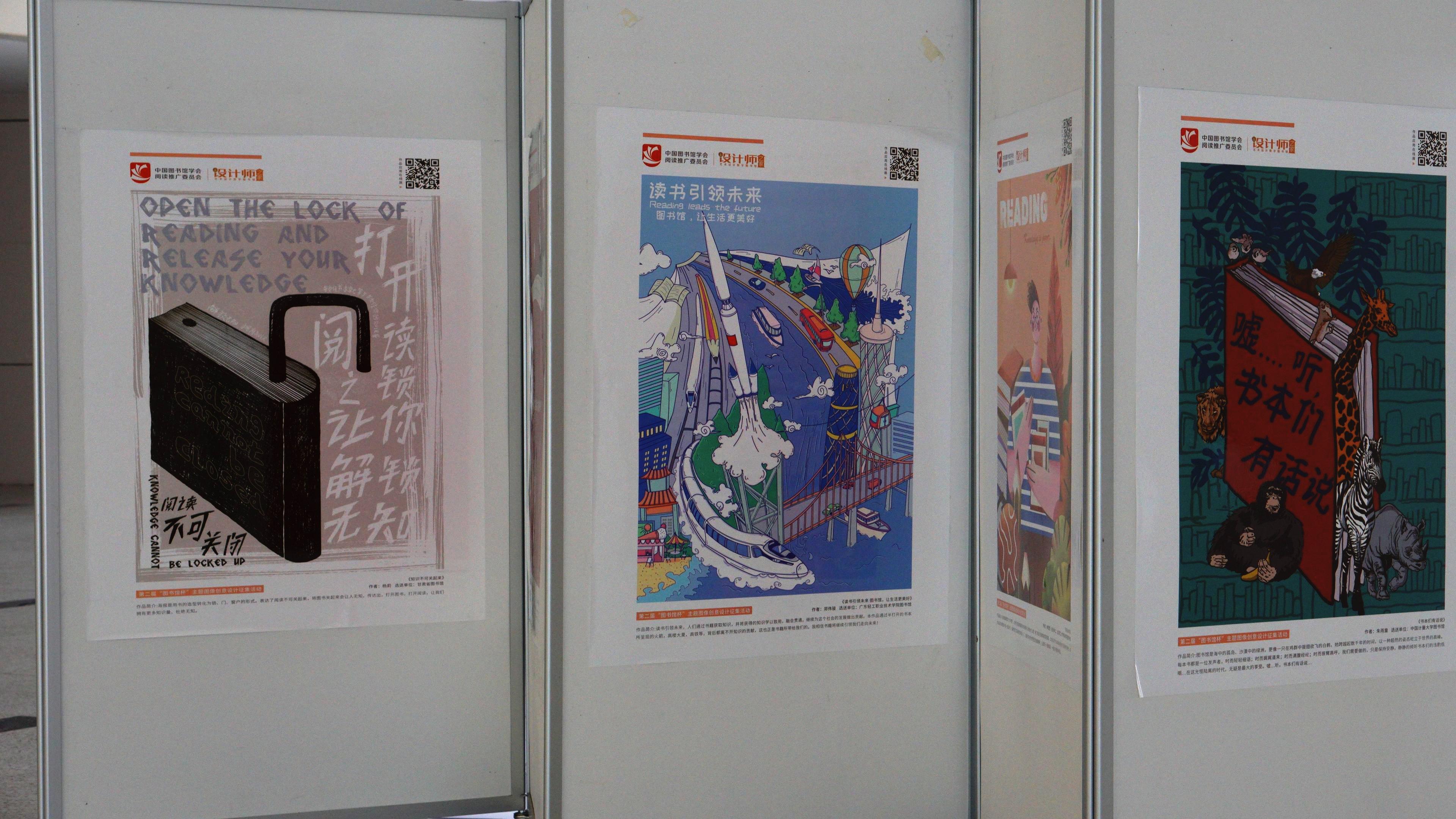}
        \caption{LLFormer\\23.57}
    \end{subfigure}
    \begin{subfigure}[b]{0.17\textwidth}
        \includegraphics[width=\textwidth]{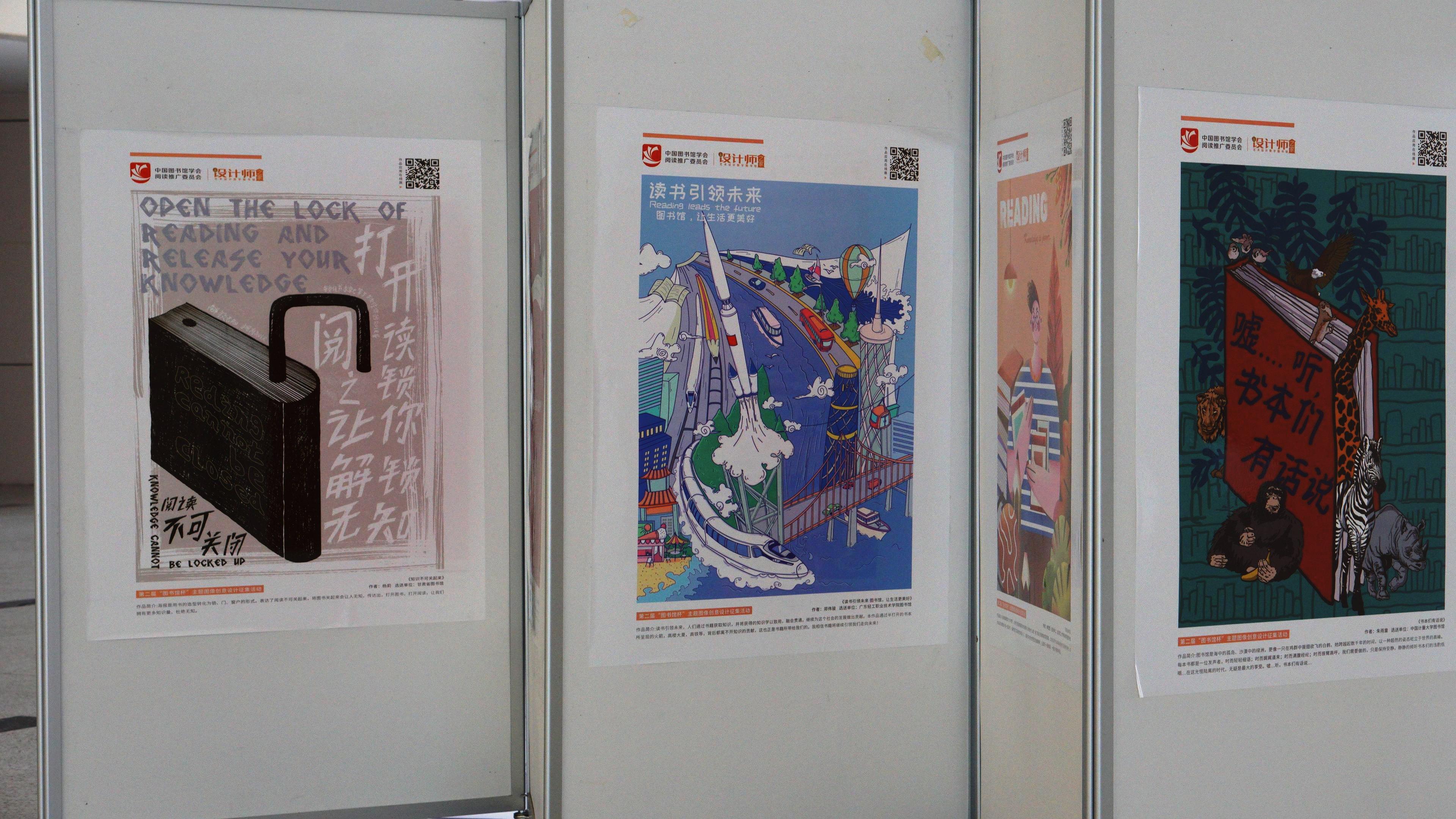}
        \caption{UHDFormer\\25.11}
    \end{subfigure}
    \begin{subfigure}[b]{0.17\textwidth}
        \includegraphics[width=\textwidth]{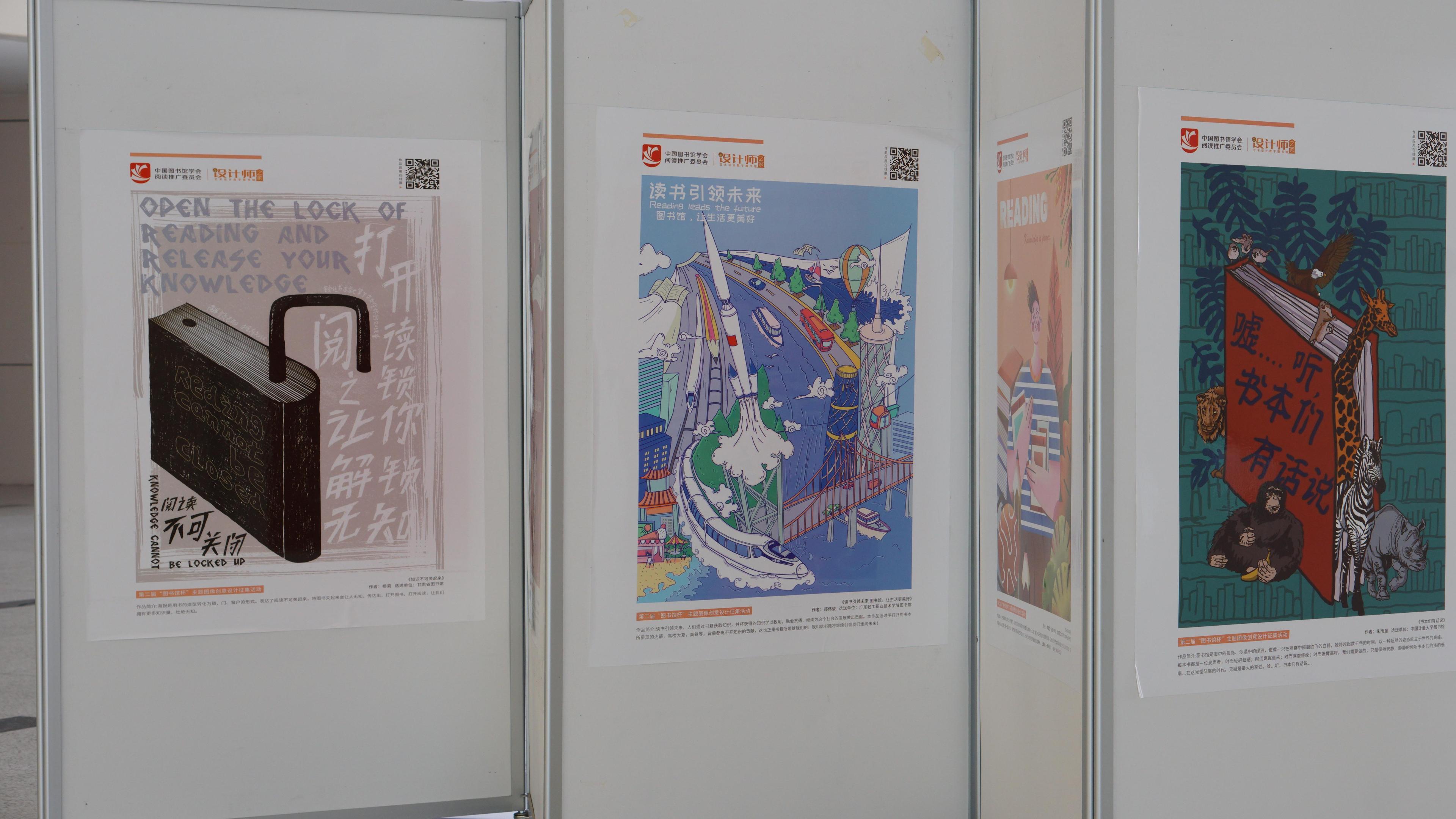}
        \caption{UHDFormer + TripleD\\25.27}
    \end{subfigure}
    \begin{subfigure}[b]{0.17\textwidth}
        \includegraphics[width=\textwidth]{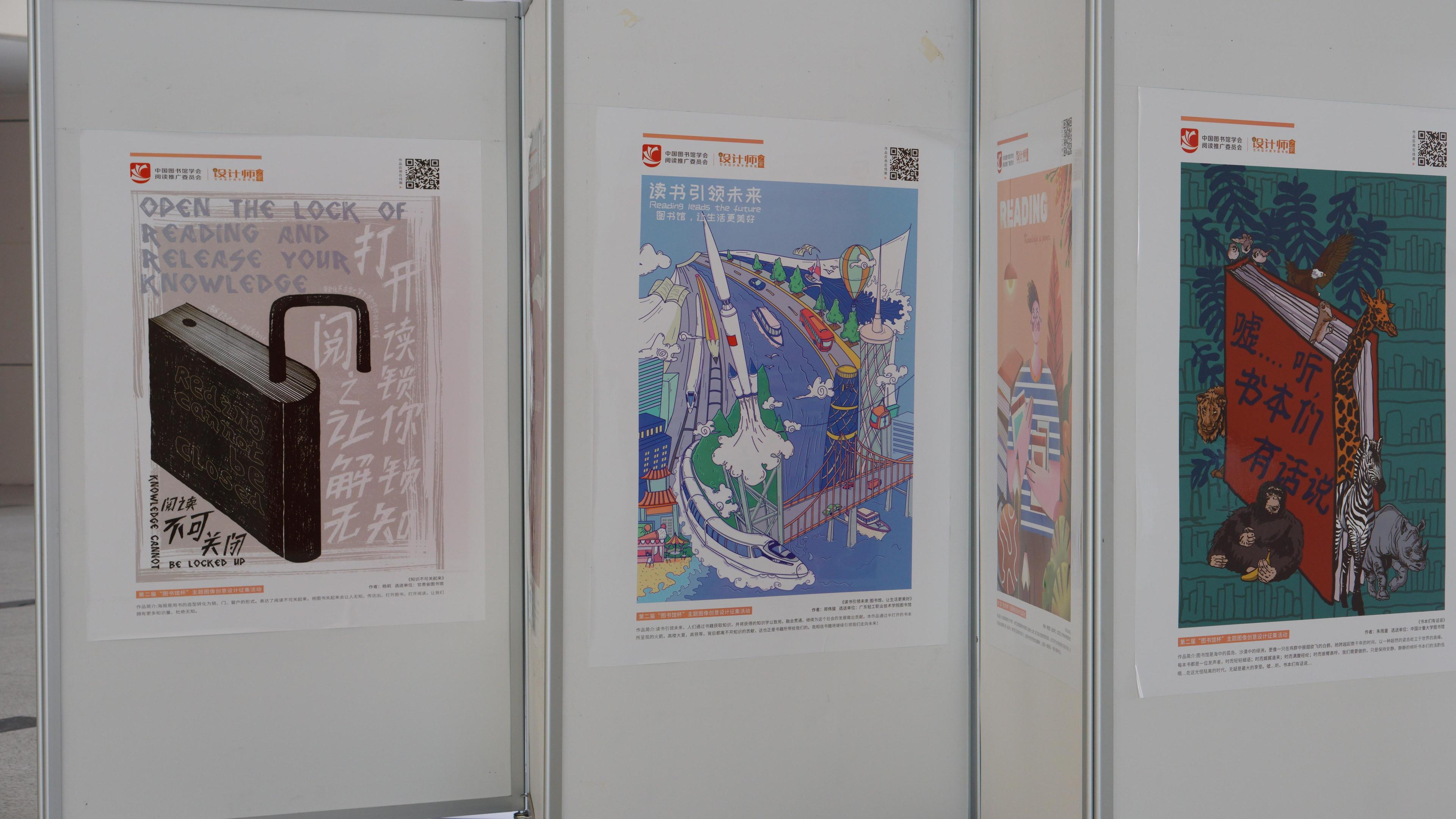}
        \caption{GT\\$+\infty$}
    \end{subfigure}}
    \vspace{-2mm}
    \caption{Comparison of methods on UHD-LL dataset. All comparison methods are retrained on the UHDFormer synthesised dataset.}
    \vspace{-4mm}
    \label{ll}
\end{figure*}

\begin{table}[!t]
\caption{Effect of Gradient Accumulation on PSNR, SSIM, and memory usage for deraining task. The use of gradient accumulation enables efficient training with reduced memory usage while maintaining competitive performance.}
\vspace{-4mm}
\label{tab:grad_accum_comparison}
\begin{center}
\scalebox{0.782}{
\begin{tabular}{c|c|c|c}
\toprule[0.15em]
\textbf{Accumulation Steps} & \textbf{Memory Usage (GB)} & \textbf{PSNR (dB)} & \textbf{SSIM} \\ 
\midrule[0.15em]
1 (No Accumulation) & 24 & 32.15 & 0.922 \\ 
4 & 17  & 32.08 & 0.910 \\ 
8 & 12  & 31.91 & 0.900 \\ 
\bottomrule[0.15em]
\end{tabular}
}\vspace{-4mm}
\end{center}
\end{table}

\subsection{Evaluation of the Feature Extractor}

In the Rain100L deraining task, we experimented with different feature extractors to evaluate their effectiveness in computing sample complexity and uncertainty. Specifically, we compared ResNet-50~\cite{he2016deep}, Vision Transformer (ViT), MLP-Mixer and Mamba to understand their impact on dynamic sample selection. 
Table~\ref{tab:feature_comparison_rain100l} summarizes the results, where we report PSNR and SSIM for each feature extractor when applied in our TripleD framework.

\begin{table*}[!t]
    \centering
    \caption{Performance comparison of PromptIR trained with TripleD against the pre-trained PromptIR model without TripleD.}
    \vspace{-2mm}
    \label{table:promptir_results}
    \begin{center}
\scalebox{0.9}{
    \begin{tabular}{l|cc|cc|ccc}
        \toprule[0.15em]
        \multirow{2}{*}{Method} & \multicolumn{2}{c|}{Dehazing on SOTS} & \multicolumn{2}{c|}{Deraining on Rain100L} & \multicolumn{3}{c}{Denoising on BSD68 (PSNR/SSIM)}  \\
        & PSNR & SSIM & PSNR & SSIM & $\sigma=15$ & $\sigma=25$ & $\sigma=50$  \\
        \midrule[0.15em]
        AirNet &34.90 &0.968& 27.94&0.962 & 33.92/0.933 & 31.26/0.888 &28.00/0.797 \\
        AirNet with TripleD &32.11 &0.923& 26.86 &0.912 & 32.12/0.913 & 30.57/0.826 &27.90/0.779 \\
        \midrule[0.15em]
        PromptIR  & 30.58 & 0.974 & 36.37 & 0.972 & 33.98/0.933 & 31.31/0.888 & 28.06/0.799  \\
        PromptIR with TripleD & 28.50 & 0.960 & 35.78 & 0.964 & 32.88/0.915 & 31.01/0.870 & 28.20/0.819  \\
        \midrule[0.15em]
        AdaIR  & 30.90 & 0.979 & 38.02 & 0.980 & 34.01/0.934 & 31.34/0.889 & 28.06/0.798  \\
        AdaIR with TripleD & 30.49 & 0.971 & 35.08 & 0.968 & 33.88/0.909 & 31.00/0.877 & 28.15/0.809  \\
        \bottomrule[0.15em]
    \end{tabular}}
     \end{center}\vspace{-4mm}
\end{table*}

\begin{figure}[!t]
\begin{center}
\scalebox{0.5}{
    \begin{tabular}{cccccc}
        % 第一行图像
        \includegraphics[width=4.9cm]{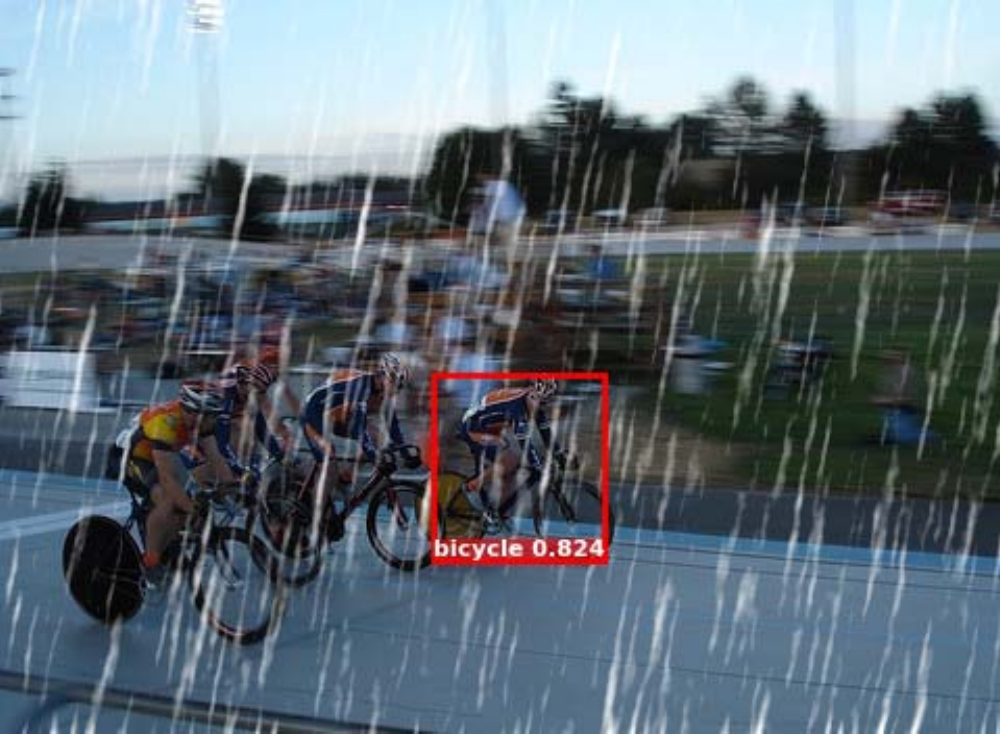} &
        \includegraphics[width=4.9cm]{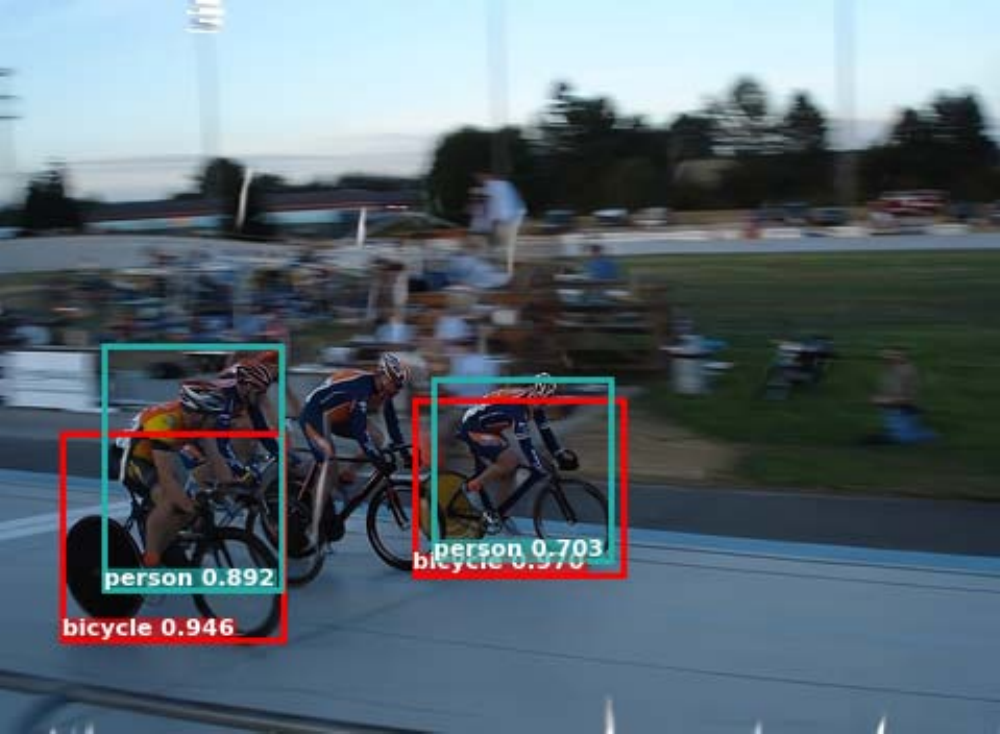} &
        \includegraphics[width=4.9cm]{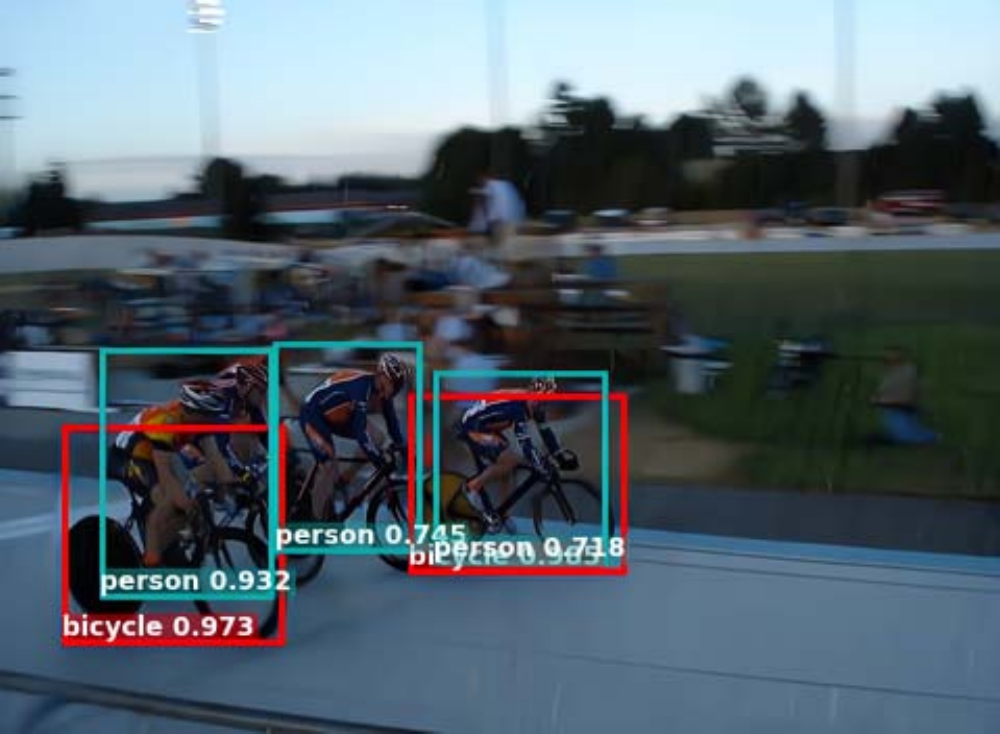}&
        \\
        (a) Input & (b) PromptIR & (c) PromptIR + TripleD \\
    \end{tabular}}
    \end{center}
    \vspace{-4mm}
    \caption{The qualitative results of object detection experiments on low-light transportation surveillance data, which select YOLOv5 \cite{ge2021yolox} as the basic detection method.}
    \vspace{-4mm}
    \label{fig:rain_removal_comparison}
\end{figure}

\begin{figure}
    \centering
    \includegraphics[width=1\linewidth]{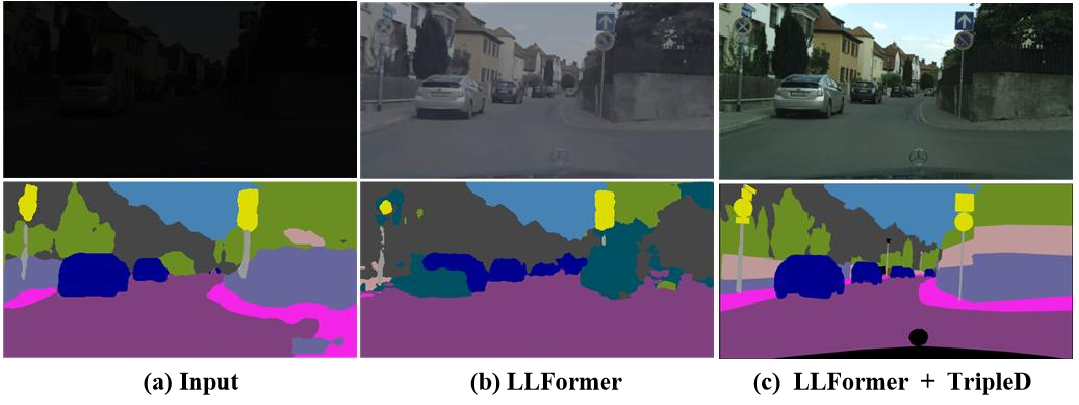}
    \vspace{-4mm}
    \caption{The results of segmentation experiments on the real world, which selects DAFormer \cite{hoyer2022daformer} as the segmentation method.}
    \vspace{-4mm}
    \label{fig:realworld}
\end{figure}

\begin{figure}[t]
    \centering
    % KDE图：分布对比
    \begin{subfigure}[b]{0.225\textwidth}
        \includegraphics[width=\textwidth]{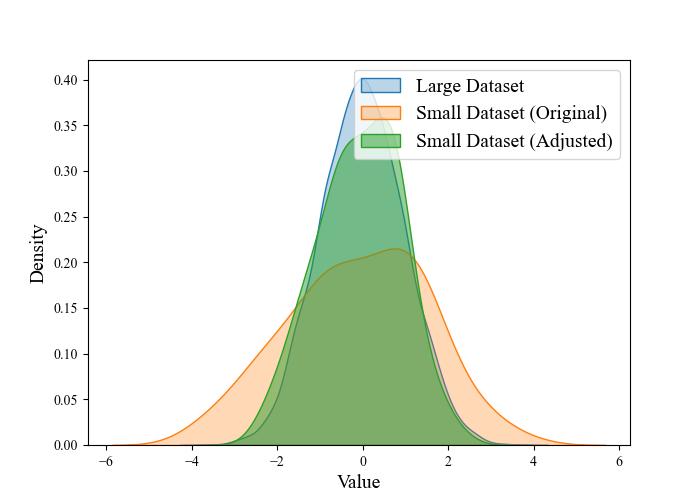}
        \caption{Kernel Density Estimate Plot}
        \label{fig:kde_plot}
    \end{subfigure}
    \hfill
    % QQ图：小数据集分布对比
    \begin{subfigure}[b]{0.225\textwidth}
        \includegraphics[width=\textwidth]{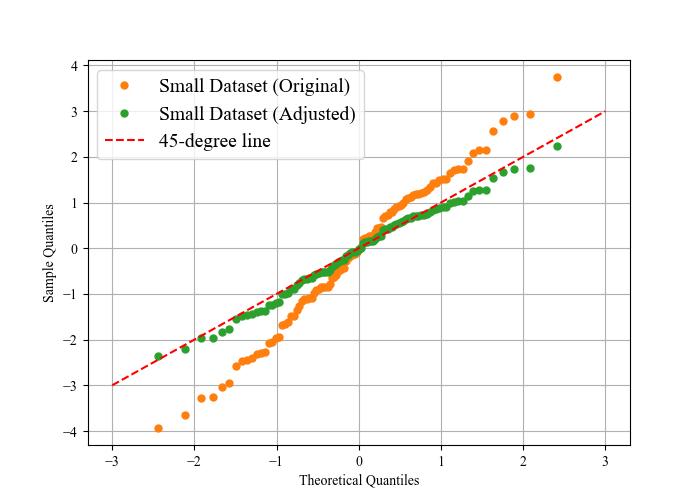}
        \caption{Quantile-Quantile (QQ) Plot}
        \label{fig:qq_plot}
    \end{subfigure}
    \vspace{-2mm}
    \caption{Distribution alignment visualization. (a) KDE plot shows that the adjusted small dataset distribution (green) aligns more closely with the large dataset distribution (blue), compared to the original small dataset (orange). (b) QQ plot illustrates the quantile alignment of the original (orange) and adjusted (green) small datasets with the theoretical quantiles of the large dataset.}
    \vspace{-4mm}
    \label{fig:distribution_alignment}
\end{figure}

%The results in Table~\ref{tab:feature_comparison_rain100l} demonstrate the varying effectiveness of different feature extractors when applied to the dynamic sample selection process in the Rain100L deraining task. ResNet-50, being a CNN-based model, shows reasonable performance with a PSNR of 30.50 dB and SSIM of 0.900. However, as ResNet-50 primarily captures local features, its performance lags behind transformer-based models like ViT and CLIP, which are better at capturing global dependencies in the images.

%Interestingly, ViT and CLIP, when used individually, perform comparably, with PSNR values of 31.08 dB and 31.12 dB, respectively. This suggests that while both models are capable of capturing global contextual information, they may have limitations in fully optimizing sample selection when used independently.

To visually demonstrate the effectiveness of different feature extractors in the Rain100L deraining task, we present qualitative comparisons of the results generated by ResNet-50, ViT, Mamba, and the MLP-Mixer combination. Figure~\ref{fig:clip-rain_removal_comparison} shows the input image and the corresponding derained outputs.
MLP-Mixer and Mamba capture image complexity in a relatively complementary relationship and are robust to degraded images.
ViT shows a better restoration effect for TripleD. In the supplementary material, we have visualised some feature maps to illustrate their effectiveness.
\subsection{Other Image Restoration Tasks}
\noindent \textbf{All-in-One Image Restoration.}
In this section, we evaluate the effectiveness of the TripleD method in training the well-known all-in-one model, PromptIR \cite{potlapalli2023promptir}, across multiple degradation types. To ensure that each degradation type, such as rain, haze, and noise, is adequately represented during training, TripleD dynamically selects samples from each degradation category, balancing the number of samples chosen for each type. 
%This approach guarantees that the model receives sufficient exposure to each degradation, avoiding the potential over-representation of any single type.
%The TripleD-enhanced PromptIR model is compared against the pre-trained PromptIR model without DMBDD, highlighting the ability of DMBDD to provide comparable or improved performance while maintaining efficiency.
Table \ref{table:promptir_results} shows the quantitative results in terms of PSNR and SSIM for different image restoration tasks, including dehazing on SOTS, deraining on Rain100L, and denoising on BSD68 with different noise levels ($\sigma=15, 25, 50$). The PromptIR with TripleD achieved performance close to that of the re-trained PromptIR model, with some tasks showing slightly better performance.
The hardware and parameter settings are the same as for multi-task image restoration.
All comparison methods were retrained on a small-scale dataset distilled from the PromptIR. 
%fig:all-in-one
As shown in Figure~\ref{fig:all-in-one}, we show the effectiveness of our method.
As shown in Figure~\ref{fig:rain_removal_comparison}, we demonstrate an object detection task to show the effectiveness of our method.

\noindent \textbf{UHD Image Restoration.}
In this section, we evaluate the effectiveness of the TripleD method in training the well-known UHD model, UHDFormer~\cite{li2023embedding}, across multiple low-light degradations. 
All comparison methods were retrained on a small-scale dataset distilled from the UHDFormer. 
Table \ref{tab:UHDLOL} and \ref{tab:UHDLL} show the quantitative results in terms of PSNR and SSIM for low-light image restoration tasks. The UHDFormer with TripleD achieved performance close to that of the re-trained UHDFormer model, with some tasks showing slightly better performance.
As shown in Figure~\ref{fig:realworld}, we demonstrate the effectiveness of our method by showing a semantic segmentation task.
As shown in Figure~\ref{ll} and~\ref{4k} , we show the effectiveness of our method.
In the supplementary material, we show more results of UHD image enhancement, including defog, derain, and desnow.
\begin{figure}[!t]
\begin{center}
\scalebox{0.5}{
    \begin{tabular}{cccccc}
        % 第一行图像
        \includegraphics[width=4.6cm]{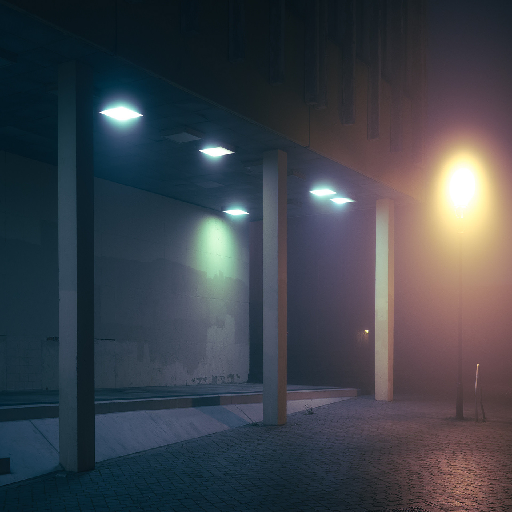} &
        \includegraphics[width=4.6cm]{figs/Real-low-light/0734_UHDFormer.png} &
        \includegraphics[width=4.6cm]{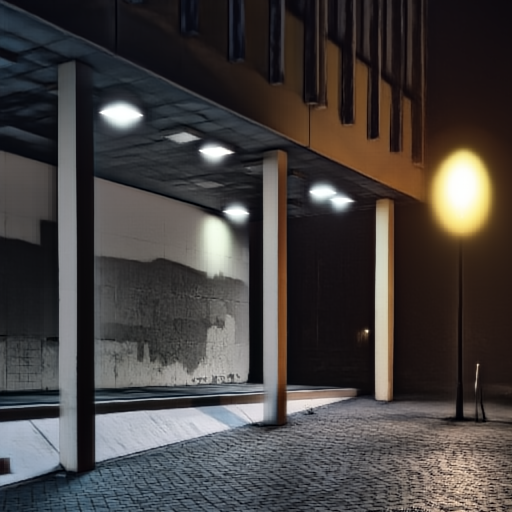}
        \\
        (a) Input & (b) UHDformer (original)  & (d) UHDformer (with TripleD)\\
    \end{tabular}}
    \end{center}
    \vspace{-4mm}
    \caption{The qualitative results in the real world.}
    \vspace{-4mm}
    \label{fig:real-world2}
\end{figure}
\begin{table}[t!]
\caption{Comparison of quantitative results on UHD-LOL4K.}
\vspace{-2mm}
\resizebox{\linewidth}{!}{
\begin{tabular}{l|c|c|cc}
\toprule[0.5mm]
Methods   & Type                                          & Venue    & PSNR $\uparrow$  & SSIM $\uparrow$  \\ \midrule
Z\_DCE++ (TripleD)  & \multicolumn{1}{c|}{\multirow{3}{*}{non-UHD}} & TPAMI'21 & 15.58 & 0.934 \\
%RUAS      & \multicolumn{1}{c|}{}                         & CVPR'21  & 14.68 & 0.757 \\
Uformer (TripleD)  & \multicolumn{1}{c|}{}                         & CVPR'22  & 29.98 & 0.980 \\
Restormer (TripleD) & \multicolumn{1}{c|}{}                         & CVPR'22  & 36.90 & 0.988 \\ \midrule
NSEN (TripleD)     & \multicolumn{1}{c|}{\multirow{5}{*}{UHD}}     & MM'23    & 29.49 & 0.980 \\
UHDFour (TripleD)   & \multicolumn{1}{c|}{}                         & ICLR'23  & 36.12 & \underline{0.990} \\
LLFormer (TripleD)  & \multicolumn{1}{c|}{}                         & AAAI'23  & 37.33 & 0.988  \\
UHDFormer & \multicolumn{1}{c|}{}                         & AAAI'24  & 36.28 & 0.989  \\
UHDFormer (TripleD)    & \multicolumn{1}{c|}{}                         & -        & 35.01 & 0.978 \\
\bottomrule[0.5mm]
\end{tabular}}
\vspace{-2mm}
\label{tab:UHDLOL}
\end{table}

\begin{table}[!]
\caption{Comparison of quantitative results on UHD-LL dataset.}
\vspace{-2mm}
\label{tab:UHDLL}
\resizebox{\linewidth}{!}{
\begin{tabular}{l|c|c|cc}
\toprule[0.5mm]
Methods   & Type                                          & Venue    & PSNR   & SSIM  \\ \midrule
Z\_DCE++ (TripleD) & \multicolumn{1}{c|}{\multirow{5}{*}{non-UHD}} & TPAMI'21 & 16.41  & 0.630 \\
%RUAS      & \multicolumn{1}{c|}{}                         & CVPR'21  & 13.56  & 0.749 \\
%Zhao      & \multicolumn{1}{c|}{}                         & TIP'21   & 21.96  & 0.870 \\
Uformer (TripleD)  & \multicolumn{1}{c|}{}                         & CVPR'22  & 19.28  & 0.849 \\
SCI  (TripleD)     & \multicolumn{1}{c|}{}                         & CVPR'22  & 16.05  & 0.625 \\
SNR-Aware (TripleD) & \multicolumn{1}{c|}{}                         & CVPR'22  & 22.17  & 0.866 \\
Restormer (TripleD) & \multicolumn{1}{c|}{}                         & CVPR'22  & 22.25  & 0.871 \\ \midrule
LLFormer (TripleD)  & \multicolumn{1}{c|}{\multirow{3}{*}{UHD}}     & AAAI'23  & 22.79  & 0.853 \\
UHDformer   & \multicolumn{1}{c|}{}                         & AAAI'24  & 26.22  & 0.900 \\
UHDFormer (TripleD)      & \multicolumn{1}{c|}{}                         & -        & 24.95  & 0.869 \\
\bottomrule[0.5mm]
\end{tabular}}
\vspace{-2mm}
\end{table}

%\section{Discussion}
%
%We observe that the CNN fine-tuned data set has a distribution that is closer to the distilled data set, as shown in Figure~\ref{fig:distribution_alignment}. In addition, as shown in Figure~\ref{fig:real-world2}, we demonstrate that our method has good generalisation in a number of real-world scenarios.
%

\section{Discussion and Conclusion}
We observe that the CNN fine-tuned data set has a distribution that is closer to the distilled data set, as shown in Figure~\ref{fig:distribution_alignment}. In addition, as shown in Figure~\ref{fig:real-world2}, we demonstrate that our method has good generalisation in the real world.
Additionally, there is an unsolved problem: why not directly use clear images (GT) for entropy evaluation? We found that using the dataset after image restoration is more robust for evaluation.

We address an important problem in image restoration, the training cost of the model. Our method uses only 2\% of the image set to maintain 90\% of the ability of the model trained on large-scale datasets.
Especially for UHD image restoration tasks, we increase efficiency by more than 100 $\times$ is trained on synthetic datasets.
Extensive experimental results demonstrate the effectiveness of our methods.

\newpage

{
    \small
    \bibliographystyle{ieeenat_fullname}
    \bibliography{main}
}


\end{document}